\newif\ifprivate\privatetrue %

\documentclass[11pt, a4paper, logo, onecolumn]{googledeepmind}

\pdftrailerid{redacted}

\makeatletter
\renewcommand\bibentry[1]{\nocite{#1}{\frenchspacing\@nameuse{BR@r@#1\@extra@b@citeb}}}
\makeatother

\ifprivate
\newcommand{\jg}[1]{\textcolor{blue}{(\emph{JG: #1})}}
\else
\newcommand{\jg}[1]{}
\fi

\usepackage{kantlipsum, lipsum}
\usepackage{dsfont}
\usepackage{gdm-colors}

\usepackage[authoryear, sort&compress, round]{natbib}
\usepackage{microtype}
\usepackage{graphicx}
\usepackage{subfigure}
\usepackage{booktabs} %
\usepackage{hyperref}       %
\definecolor{linkcol}{RGB}{134,22,87} %
\definecolor{citecol}{RGB}{30, 30, 180} %
\definecolor{urlcol}{RGB}{20,88,21}

\hypersetup{
  colorlinks=true,
  citecolor=magenta,
}
\usepackage{url}
\usepackage{amsmath}
\usepackage{amssymb}
\usepackage{mathtools}
\usepackage{amsthm}
\usepackage[labelfont=bf]{caption}

\usepackage{natbib}
\usepackage{tcolorbox}
\usepackage{color}
\usepackage{xspace}
\usepackage{graphicx}
\usepackage{wrapfig}
\usepackage{multicol}
\usepackage{tcolorbox}
\usepackage{soul}
\usepackage{multirow}
\usepackage{color}
\usepackage{listings}
\usepackage{pifont}
\usepackage{graphics}

\usepackage{inconsolata}
\definecolor{antiquewhite}{rgb}{0.98, 0.92, 0.84}
\definecolor{azure(web)(azuremist)}{rgb}{0.84,0.98,0.92}
\definecolor{lavender}{rgb}{0.92,0.84,0.98}
\definecolor{mistyrose}{rgb}{1,0.89,0.88}
\definecolor{lightsalmon}{rgb}{1,0.63,0.48}
\definecolor{rosybrown}{rgb}{0.74, 0.56, 0.56}
\definecolor{lightcoral}{rgb}{0.94, 0.5, 0.5}
\definecolor{verylightrosybrown}{rgb}{0.9, 0.75, 0.75}
\definecolor{lightrosybrown}{rgb}{0.84, 0.66, 0.66} 
\definecolor{softlilac}{rgb}{0.75, 0.65, 0.8}
\definecolor{paleblush}{rgb}{0.98, 0.8, 0.85}
\definecolor{lightgray}{rgb}{0.9, 0.9, 0.9}
\usepackage[capitalize,noabbrev]{cleveref}

\newcommand{\stoptocwriting}{%
  \addtocontents{toc}{\protect\setcounter{tocdepth}{-5}}}
\newcommand{\resumetocwriting}{%
  \addtocontents{toc}{\protect\setcounter{tocdepth}{\arabic{tocdepth}}}}

\graphicspath{{figures/}}

\title{LLMs are Greedy Agents: Effects of RL Fine-tuning on Decision-Making Abilities}

\correspondingauthor{schmied@ml.jku.at, razp@google.com}

\author[1,+]{Thomas Schmied}
\author[2]{Jörg Bornschein}
\author[2]{Jordi Grau-Moya}
\author[2]{Markus Wulfmeier}
\author[2]{Razvan Pascanu}

\affil[1]{ELLIS Unit, LIT AI Lab, Institute for Machine Learning, JKU Linz, Austria}
\affil[2]{Google DeepMind}
\affil[+]{Work done at Google DeepMind}

\begin{abstract}
The success of Large Language Models (LLMs) has sparked interest in various agentic applications.
A key hypothesis is that LLMs, leveraging common sense and Chain-of-Thought (CoT) reasoning, can effectively explore and efficiently solve complex domains. 
However, LLM agents have been found to suffer from sub-optimal exploration and the knowing-doing gap, the inability to effectively act on knowledge present in the model.
In this work, we systematically study \emph{why} LLMs perform sub-optimally in decision-making scenarios.
In particular, we closely examine three prevalent failure modes: greediness, frequency bias, and the knowing-doing gap.
We propose mitigation of these shortcomings by fine-tuning via Reinforcement Learning (RL) on self-generated CoT rationales.
Our experiments across multi-armed bandits, contextual bandits, and Tic-tac-toe, demonstrate that RL fine-tuning enhances the decision-making abilities of LLMs by increasing exploration and narrowing the knowing-doing gap.
Finally, we study both classic exploration mechanisms, such as $\epsilon$-greedy, and LLM-specific approaches, such as self-correction and self-consistency, to enable more effective fine-tuning of LLMs for decision-making.
\end{abstract}

\begin{document}

\stoptocwriting

\maketitle

\section{Introduction}
\noindent Large Language Models (LLMs) pre-trained on massive internet-scale datasets have demonstrated success across diverse domains, including text generation and language understanding \citep{radford2019language, brown2020language, team2023gemini, team2024gemini, dubey2024llama}.
Their broad pre-training distribution, enables generalization to a wide range of scenarios including coding assistance \citep{li2022competition}, education \citep{team2024learnlm}, and medicine \citep{saab2024capabilities}.
Therefore, their success has sparked interest in using LLMs for decision-making problems \citep{chen2023introspective, krishnamurthy2024can, nie2024evolve} at the core of agentic AI systems \citep{durante2024agent}.

\noindent One key hypothesis is that LLMs can generate informed action predictions without extensive environment interaction \citep{lu2024intelligent} due to “world knowledge” present in the model. %
Moreover, Chain-of-Thought (CoT) \citep{wei2022chain} equips models with the ability to reason about the observed history and their actions, which facilitates environment interaction.
However, these advantages do not seem to materialize into strong performance when LLMs are faced with decision-making scenarios. 
Notably, \citet{krishnamurthy2024can} and \citet{nie2024evolve} found that LLMs do not robustly engage in \emph{exploration} resulting in sub-optimal behavior.
Similar shortcomings of LLMs have been observed by \citet{paglieri2024balrog} and \citet{ruoss2024lmact} on stateful environments commonly used in RL (e.g., grid-worlds, Atari).
Both works attribute the shortcomings to the \emph{knowing-doing gap}, which states that models can posses knowledge about a task or can describe the consequences of their behavior (i.e., they know what to do), but cannot materialize this knowledge when acting (i.e., incapable of doing).
Consequently, sub-optimal exploration and the knowing-doing gap are considerable obstacles towards more powerful agentic LLMs.

\noindent In this work, we aim to understand \emph{why} LLMs often perform sub-optimally in simple decision-making scenarios. 
In particular, we systematically study three prevalent failure modes in small-to-medium-scale LLMs: greediness, frequency bias, and the knowing-doing gap (see Section \ref{sec:failure-modes}).
Our analysis shows that final performance often remains sub-optimal, because LLMs prematurely commit to greedy action selection strategies leading to stagnating action coverage that leave a large part of the action space unexplored (up to 55\%).
Moreover, we observe that small-scale LLMs (2B) tend to copy the most frequent actions in the context regardless of their respective reward, which we refer to as frequency bias.
In contrast, larger LLMs (27B) mostly diminish the frequency bias, yet they remain prone to greedy behavior at the cost of exploration.
Similarly, we quantify the knowing-doing gap and find that LLMs often know how to solve a task (87\% correct rationales) but fail at acting on this knowledge as they prioritize greedy actions (64\% of actions when rationale is correct). 

\noindent To overcome these shortcomings, we propose Reinforcement Learning Fine-Tuning (RLFT) on self-generated CoT rationales.
RL is the pre-dominant learning paradigm in decision-making scenarios and has been successful in game-playing \citep{silver2016mastering, vinyals2019grandmaster}, robotics \citep{pmlr-v270-tirumala25a}, plasma-control \citep{degrave2022magnetic}, or navigating stratospheric balloons \citep{bellemare2020autonomous}.
We study the effects of RLFT on pre-trained Gemma2 models \citep{team2024gemma,team2024gemma2} in three sizes (2B, 9B, and 27B) in multi-arm bandit (MAB) and contextual bandit (CB) settings proposed by \citet{nie2024evolve}, and the textual Tic-tac-toe environment released by \citet{ruoss2024lmact}.
Across environments, we find that RLFT enhances the decision-making abilities of LLMs by increasing exploration and narrowing the knowing-doing gap.
While RLFT positively affects exploration of LLM agents, their exploration strategies remain sub-optimal.
Therefore, we empirically evaluate both “classic” exploration mechanisms commonly employed in RL, such as $\epsilon$-greedy, and LLM-specific approaches, such as self-correction and self-consistency, to enable more effective fine-tuning for decision-making scenarios.
Finally, in our ablations we investigate the importance of CoT reasoning for decision-making, highlight the effectiveness of leveraging expert data, and show the benefits of giving the agent more reasoning tokens to solve the decision-making problem. 

\noindent In summary, we make the following \textbf{contributions}: 
\begin{itemize}
    \item We systematically examine three failure modes of small-to-medium scale LLMs in decision-making scenarios: greediness, frequency bias, and the knowing-doing gap.
    \item We study how fine-tuning LLMs via RL on self-generated CoT rationales affects these shortcomings, highlighting positive effects of RLFT on exploration and decision-making abilities.
    \item We evaluate a variety of exploration mechanisms (e.g., $\epsilon$-greedy) and LLM-specific approaches (e.g., self-consistency), to enable more effective RLFT for LLMs.
\end{itemize}
\begin{figure*}[t]
    \centering
        \includegraphics[width=0.85\linewidth]{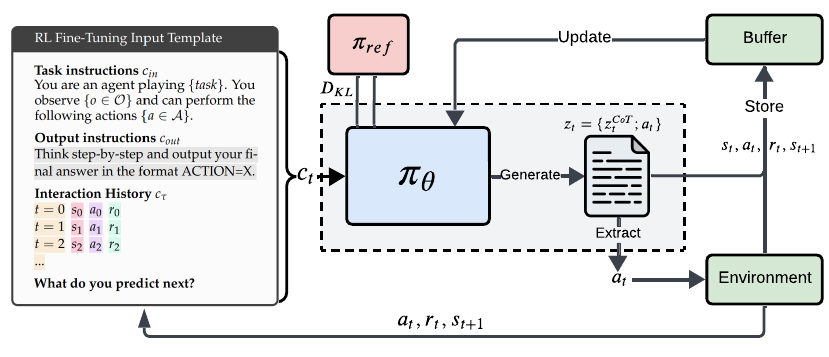}     
    \caption{Illustration of our \textbf{Reinforcement Learning Fine Tuning (RLFT)} pipeline. We fine-tune a pre-trained LLM $\pi_{\theta}$ via self-generated Chain-of-Thought (CoT) rationales on environment rewards.}
    \label{fig:sca}
\end{figure*}

\section{Related Work}
\noindent\textbf{Exploration in RL and LLMs.}
The trade-off between \emph{exploration} and \emph{exploitation} is a long-standing challenge in the field of RL \citep{schmidhuber1991curious,schmidhuber1991possibility,still2012information,oudeyer2007intrinsic}.
Widely used RL agents have often relied on random schemes \citep{mnih2015human}, heuristics such as state-visitation counts \citep{ecoffet2019go,raileanu2020ride}, intrinsic curiosity \citep{pathak2017curiosity, burda2018exploration,groth2021curiosity}, behavior priors \citep{raolearning}, or maximum entropy regularization \citep{schulman2017proximal,haarnoja2018soft}.
Naturally, a number of works looked into leveraging LLMs for improving exploration of RL agents either as a source of rewards \citep{klissarov2023motif, lu2024intelligent} or to orchestrate exploration strategies \citep{klissarov2024maestromotif}.
\cite{krishnamurthy2024can} investigate the in-context exploration abilities of LLMs when acting directly as a policy.
Similarly, \citet{nie2024evolve} study the exploration abilities of LLMs when fine-tuned on expert trajectories. 
In contrast, our work investigates the effects of RLFT on the exploration abilities of LLMs and focuses on \emph{why} models fail. 

\noindent\textbf{In-context Learning for Decision-Making.}
ICL is a form of Meta-learning, also referred to as learning-to-learn \citep{Schmidhuber:87}.
While meta-learning is targeted via a meta-training phase \citep{Santoro:16,Mishra:18,Finn:17,Wang:16,duan2016rl,Kirsch:19,Flennerhag:19,Team2023HumanTimescaleAI}, ICL emerges as a result of the pre-training data distribution \citep{Chan:22,Kirsch:22}. ICL has been rediscovered in LLMs \citep{Brown:20} after initial observations by \citet{Hochreiter:01} in LSTMs \citep{Hochreiter:97}.
\citet{mirchandani2023large} leverage the ICL abilities of LLMs to operate as general pattern machines. 
A number of works leverage the CoT abilities \citep{wei2022chain} of LLMs in simple text-based scenarios \citep{shinn2023reflexion,yao2022react}. 
Similar in-context abilities have been observed in decision-making with models trained from scratch, albeit in restricted environments \citep{Laskin:22,Lee:22,Kirsch:23,Raparthy:23,schmied2024retrieval,schmied2024large}.

\noindent\textbf{Self-Correction in LLMs.}
A critical component for LLM agents is the ability to self-correct over previously explored attempts.
Existing works focus primarily on math benchmarks \citep{cobbe2021training,hendrycks2021measuring,welleck2022generating}.
\citet{zelikman2022star} leverage hints to iteratively generate correct answers and fine-tune on the respective CoT rationales.  
\citet{kumar2024training} employ RLFT over multiple trials to induce self-correction.
Similarly, \citet{zelikman2024quiet} make use of RL fine-tuning, but instead generate rationales at every token position.
Instead of imitation, \citet{wang2025critique} rely on critique fine-tuning to induce self-correction.
\citet{wulfmeier2024irl} make use of inverse RL to avoid compounding errors. 
Other works rely on ICL abilities to learn from previous mistakes \citep{zhang2024context,monea2024llms}.
While conceptual corrections are possible, exact token-level correction is usually difficult for autoregressive generation \citep{cundy2023sequencematch}.

\section{Methodology}
\subsection{Background}
\textbf{Reinforcement Learning.} 
We assume the standard RL formulation via a Markov Decision Process (MDP) represented by a tuple of $(\mathcal{S},\mathcal{A},\mathcal{P},\mathcal{R})$, where $\mathcal{S}$ and $\mathcal{A}$ denote state and action spaces, respectively.
At every timestep $t$ the agent observes state $s_t \in \mathcal{S}$, predicts action $a_t \in \mathcal{A}$, and receives a reward $r_t$ given by the reward function $\mathcal{R}(s_t, a_t)$.
$\mathcal{P}(s_{t+1}\mid s_t, a_t)$ defines the transition dynamics constituting a probability distribution over next states $s_{t+1}$.
The goal of RL is to learn a policy $\pi_{\theta}(a_t\mid s_t)$ with parameters $\theta$ that predicts an action $a_t$ in state $s_t$ that maximizes cumulative reward.

\noindent\textbf{Reinforcement Learning from Human Feedback.}
RLHF aims to fine-tune pre-trained models towards human preferences \citep{christiano2017deep}. 
Preferences are typically encoded via a reward model $r_{\phi}$ with parameters $\phi$ learned from a human annotated dataset $\mathcal{D}$ consisting of query-response pairs $x$ and $y$, respectively. 
RLHF optimizes a constrained REINFORCE estimator \citep{williams1992simple}: 
\begin{equation}
    \max_{\theta} \mathbb{E}_{x \sim \mathcal{D}, y \sim \pi_{\theta}(\cdot \mid x)} \big[ (r_{\phi}(x, y) - b) \nabla_{\theta} \log \pi_{\theta}(y \mid x)  - \beta D_{KL}( \pi_{\theta}(\cdot \mid x) \mid\mid \pi_{ref}(\cdot \mid x)\big]
\label{eqn:rlhf}
\end{equation}
Here $\pi_{ref}$ is a reference policy, which is typically the frozen pre-trained model, and $\beta$ is as weighting term. 
The baseline $b$ represents a baseline to reduce variance and is commonly instantiated by a value function \citep{schulman2017proximal,ouyang2022training} or a Monte-Carlo (MC) estimate of the returns \citep{ahmadian2024back,ramesh2024group,shao2024deepseekmath}.

\subsection{Reinforcement Learning Fine-Tuning (RLFT)}\label{sec:method}
Our RLFT approach relies on fine-tuning on self-generated CoT rationales on rewards obtained from environment interaction.
During RLFT the model learns to iteratively refine its reasoning process, favoring CoT patterns and actions that lead to higher rewards (see Figure \ref{fig:sca}).
Our approach is similar to \citet{guo2025deepseek} and \citet{zhai2025fine}, but specialized for decision-making scenarios.

\noindent\textbf{Context Representation.}
The input tokens to our model at step $t$ consists of input instructions $c_{t}^{in}$, output instructions $c_{t}^{out}$, and the most recent interaction history $c_t^{\tau_{t-C:t}}$ (see Figure \ref{fig:sca}).
The history representation contains the trajectory $\tau_{t-C:t}=(s_{t-C}, a_{t-C}, r_{t-C}, \dots, s_t, a_t, r_t)$ of the $C$ most recent states, actions, and rewards. 
We opt for task-specific instructions for $c_t^{in}$ rather than a generic instruction template, providing the agent with information about the observations, the possible actions, and its objective. 
Consequently, $c_t$ is represented by the concatenation of the instruction and history tokens $c_t = [c_{t}^{in}; c_{t}^{out}; c_t^{\tau_{t-C:t}}]$.

\noindent\textbf{Factorization of Action Tokens.}
At every interaction step $t$, the agent generates action tokens $z_t = [z_t^{CoT}; a_t]$ containing both the CoT reasoning tokens $z_t^{CoT}$ and the action to be executed in the environment $a_t$.
To extract $a_t$ from $z_t$, we make use of an extraction function $a_t = g(z_t)$.
In practice, $g$ consists of regular expressions to match the output pattern given by $c_{t}^{out}$.
If no valid action is found a random action is executed.
To allow for flexibility in refining the reasoning process, we opt for a permissive output template (i.e., \texttt{ACTION=X}), rather than enforcing a structured output template (e.g., \texttt{<thought>} and \texttt{<action>} blocks). 
We employ a token generation budget of $G$ tokens ($G=256$ by default), therefore $|z_t| \leq G$.

\noindent\textbf{Reward Shaping for Valid Actions.}
In addition to the environment reward $r_t^{env}$, we employ a reward shaping term $r_t^{valid}$ to encourage the model to adhere to the output template, $r_t= r_t^{env} + r_t^{valid}$.
More specifically, we make use of a reward penalty of $-5$ if $g$ cannot extract a valid action,  $r_t^{valid} = -5 \cdot \mathds{1}(g(a_t^{act}) \notin \mathcal{A})$.
To ensure that the reward penalty does not overly bias optimization, we employ reward normalization to the environment rewards. 

\noindent\textbf{Fine-tuning objective.} 
We fine-tune using the clipping objective introduced by \citet{schulman2017proximal} with and additional KL constraint to the reference policy $\pi_{ref}$: 
\begin{equation}
\max_{\theta} \mathbb{E}_{(c,z) \sim \mathcal{D}} \left[ \min\left(\frac{\pi_\theta(z|c)}{\pi_{\theta_{old}}(z|c)}A_{adv}, \text{clip}_{\epsilon}\left(\frac{\pi_\theta(z|c)}{\pi_{\theta_{old}}(z|c)}\right)A_{adv}\right) - \beta D_{KL}(\pi_\theta(\cdot|c)||\pi_{ref}(\cdot|c)) \right]
\label{eqn:rlft}
\end{equation}
Here $\pi_{\theta_{old}}$ refers to the rollout generating policy, $D$ is the rollout buffer, and $\epsilon$ is a hyperparameter.
To allow for memory efficient fine-tuning in environments with fixed episode lengths (bandits), we make use of a Monte Carlo baseline to estimate $A_{adv}$.
Instead of exploiting multiple rollouts, as used by \citet{ahmadian2024back} and \citet{ramesh2024group}, we compute rewards-to-go.
For environments with variable episode lengths (Tic-tac-toe), we learn a separate state-value head on top of the last layer LLM representations and make use of generalized advantage estimation \citep{schulman2015high}.

\noindent We provide additional implementation and training details in Appendix \ref{appendix:exp-details}.

\section{Experiments}
We study the effect of fine-tuning Gemma2 \citep{team2024gemma,team2024gemma2} models in MAB and CB settings proposed by \citet{nie2024evolve}, and on a text-based version of Tic-tac-toe released by \citet{paglieri2024balrog}. 
We describe our environments and baselines in Section \ref{sec:environments-baselines}.
For our experiments, we compare Gemma2 \citep{team2024gemma2} models at three model scales: 2B, 9B and 27B.
\noindent In Section \ref{sec:failure-modes}, we first analyze three common failure modes of LLM agents in MAB scenarios: \textbf{(1)} greediness, \textbf{(2)} frequency bias, and \textbf{(3)} the knowing-doing gap. 
Then we investigate the effects of fine-tuning on self-generated CoT rationales or expert rationales in MABs and CBs (see Section \ref{sec:multi-armed-bandits}), and in Tic-tac-toe (see Section \ref{sec:ablations}).
In Section \ref{sec:exps-exp-interventions}, we study the effects of a exploration mechanisms on the fine-tuning performance. 
Finally, in Section \ref{sec:ablations} we empirically examine important components of our approach.

\subsection{Environments \& Baselines} \label{sec:environments-baselines}

\textbf{Multi-armed and Contextual Bandits.}
MABs \citep{slivkins2019introduction, lattimore2020bandit} are a classic problem setting in RL that isolates the \emph{exploration-exploitation} trade-off.
For our MAB experiments, we leverage the text-based bandit scenarios released by \citet{nie2024evolve}.
We focus on the \emph{continuous} and \emph{button} variants, as illustrated in Figure \ref{fig:bandits-with-cot}.

\begin{wrapfigure}{r}{0.5\textwidth}
\vspace{-0.3cm}
\noindent
\small
\centering
  \footnotesize
    \begin{tcolorbox}[colback=gray!3!white, colframe=gray!50!black, title=Button Multi-armed Bandit (Gaussian)]

        You are a bandit algorithm in a room with \sethlcolor{paleblush}~\hl{5 buttons} labeled \sethlcolor{paleblush}~\hl{red, green, blue, yellow, orange}. [...].
        Your goal is to maximize the total reward. [More instructions]\newline

        \sethlcolor{lightgray}~\hl{Think step-by-step and output your final answer in the format ACTION=X where X is one of the arms listed above. 
        IMPORTANT: Provide your (SHORT!) thinking process and your answer ACTION=X}\newline
        
        So far you have tried/seen: \newline
        \sethlcolor{antiquewhite}~\hl{Step=0} \sethlcolor{lavender}~\hl{Action=green} \sethlcolor{azure(web)(azuremist)}~\hl{Reward=0.3} \newline
        \sethlcolor{antiquewhite}~\hl{Step=1} \sethlcolor{lavender}~\hl{Action=blue} \sethlcolor{azure(web)(azuremist)}~\hl{Reward=0.1} \newline
        \sethlcolor{antiquewhite}~\hl{Step=2} \sethlcolor{lavender}~\hl{Action=orange} \sethlcolor{azure(web)(azuremist)}~\hl{Reward=-0.5} \newline
        \sethlcolor{antiquewhite}~\hl{Step=3} \sethlcolor{lavender}~\hl{Action=red} \sethlcolor{azure(web)(azuremist)}~\hl{Reward=0.5} \newline
        \sethlcolor{antiquewhite}~\hl{Step=4} \sethlcolor{lavender}~\hl{Action=green} \sethlcolor{azure(web)(azuremist)}~\hl{Reward=0.24} \newline
        \sethlcolor{antiquewhite}~\hl{...} \newline 
        What do you predict next? 
    \end{tcolorbox}
\caption{Illustration of a Gaussian MAB for the \emph{button} scenario from \citep{nie2024evolve} using our context representation and instructions.}
\label{fig:bandits-with-cot}
\end{wrapfigure}
\noindent We report results for MAB with $k \in \{5,10,20\}$ arms ($|\mathcal{A}| = k$) and payoffs of the arms being either Gaussian or Bernoulli distributed.
In addition, we consider three levels of stochasticity (low/medium/high) that determine the standard deviation or delta gap in Gaussian or Bernoulli bandits, respectively.
For all MAB settings, we limit the horizon $T$ to 50 interaction steps.
\noindent We compare against two commonly used baselines for MABs: Upper-confidence Bound (UCB) \citep{auer2002using} and a random agent that selects actions uniformly at random. 
UCB is considered optimal and represents the upper-bound for agent performance, whereas the random baseline represents the lower bound. 
We provide more details on our MAB and CB setups in Appendices \ref{appendix:mabs} and \ref{appendix:cbs}, respectively.

\noindent\textbf{Tic-tac-toe.}
In addition, we use the text-based Tic-tac-toe environment released by \citet{ruoss2024lmact}, which exhibits proper state transitions.
\citet{ruoss2024lmact} demonstrated that frontier models struggle to achieve strong performance in this environment and barely beat a random opponent.
Consequently, it is a good target to investigate the efficacy of RLFT.
In Appendix \ref{appendix:ttt}, we provide addition details on our environment and training setup.

\subsection{Why do LLMs perform suboptimally in decision-making?} \label{sec:failure-modes}
Prior works found that LLM agents perform suboptimally and fail to explore sufficiently in interactive settings \citep{paglieri2024balrog,ruoss2024lmact}. 
Therefore, we first examine \emph{why} models perform suboptimally and identify three prevalent failure modes:
\textbf{(1)} greediness, \textbf{(2)} frequency bias, and \textbf{(3)} the knowing-doing gap.
In this section, we present analyses of Gemma2 models when given input contexts that elucidate the failure modes.
We conduct our analyses on the \emph{button} instance of our MAB experiments at three model scales, and find that the failure modes persist across model scales (see Appendix \ref{appendix:failures-modes} for \emph{continuous} instance). 

\begin{figure*}[h]
\noindent
\small
    \centering
        
\subfigure[Action Coverage: 10 arms]{
    \includegraphics[width=0.31\linewidth]{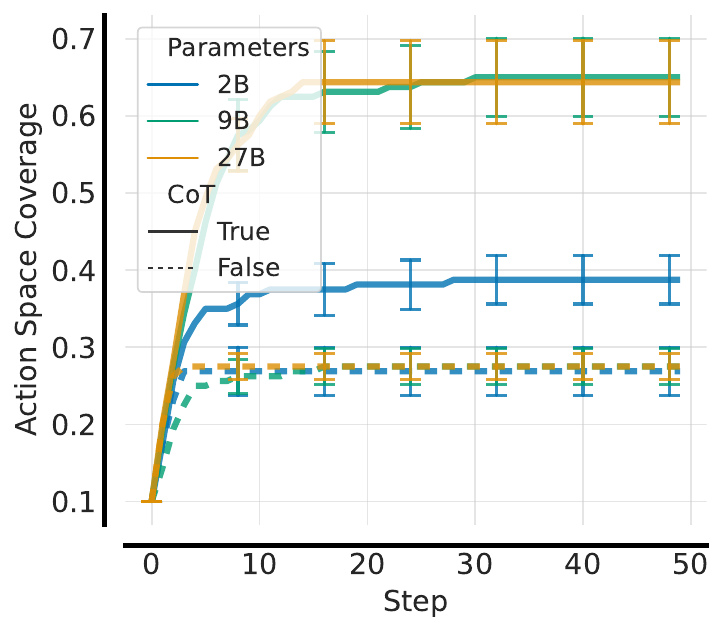}
}
~
\subfigure[Action Coverage: 20 arms]{
    \includegraphics[width=0.31\linewidth]{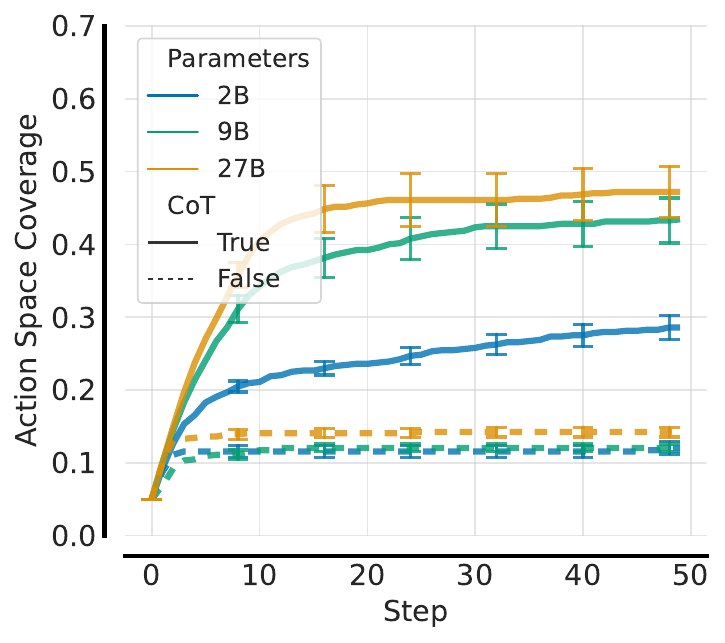}
}
~
\subfigure[Coverage vs. Regret: 10 arms]{
    \includegraphics[width=0.31\linewidth]{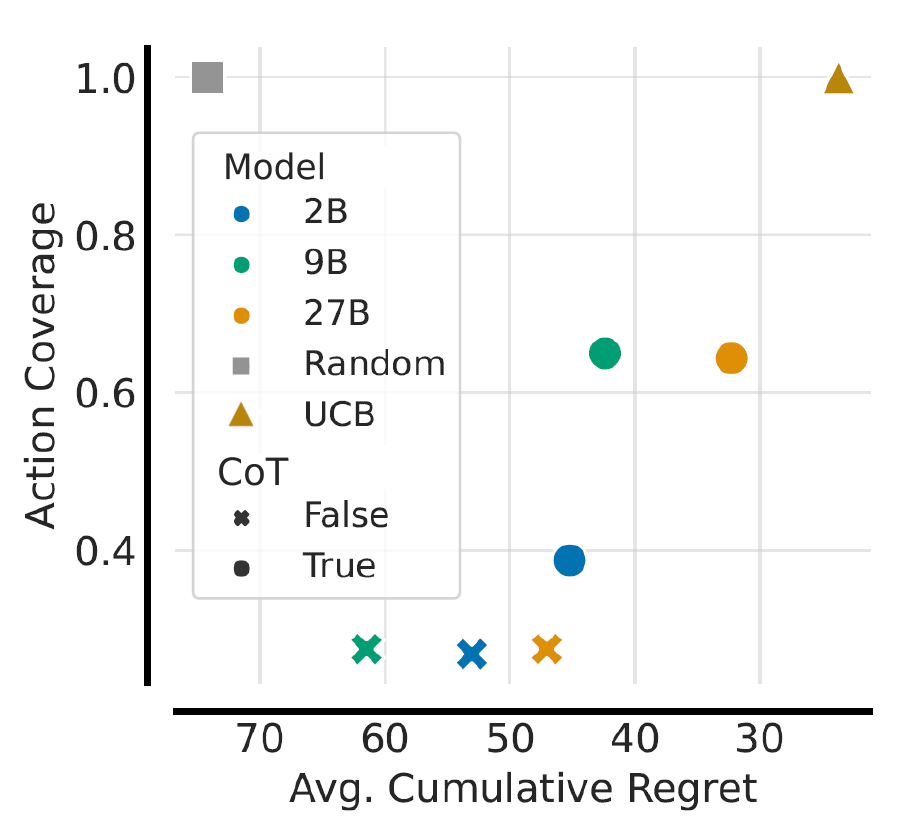} 
}
\caption{Illustration of \textbf{Greediness}. We show action coverage for Gemma2 2B/9B/27B w/ and w/o CoT for \textbf{(a)} 10 and \textbf{(b)} 20 arms over 50 interaction steps. Agents favor the best performing action among the set of selected actions, leading to stagnating action coverage, despite benefits of larger models and CoT. 
In \textbf{(c)}, we plot cumulative regret against action coverage. The agents exhibit suboptimal regret, because of greedy action selection strategies.
}
\label{fig:failure-greedy}
\end{figure*}
\noindent\textbf{Greediness.}
The first and most pervasive failure mode is \emph{greediness}, which is characterized by the LLM overly favoring the best performing action among a small set of actions seen so far. 
To illustrate this failure mode, we show the average action coverage achieved by Gemma2 2B/9B/27B with and without CoT across 64 MABs with 10 and 20 arms over 50 interaction steps (see Figure \ref{fig:failure-greedy} a and b).
We define action coverage $C_t$ at step $t$ as the fraction of available actions that have been selected at least once, $C_t=\frac{\{ a \in \mathcal{A} :\, N_t(a) > 0 \}}{\mid \mathcal{A} \mid}$ with $N_t(a)$ representing the number of times action $a \in \mathcal{A}$ has been selected until $t$.
For 10 arms and averaged over 64 parallel environments, we find that Gemma2 2B covers $40\%$ of all actions, while 9B/27B cover $65\%$ (i.e., 6.5 actions), leaving a significant part of the action space unexplored. 
Note that without CoT all models explore merely $25\%$ of all actions in the 10 arms setting. 
The suboptimal coverage is caused by the model overly favoring high-reward actions (see Figure \ref{fig:failure-greedy-pmass} in Appendix \ref{appendix:failures-modes-greediness}). 
Consequently, the model prematurely commits to a greedy strategy leading to a stagnating action coverage beyond 10 steps.
Increasing the number of arms makes the greediness even more apparent, with the largest models only covering $45\%$ of all actions.
Due to this, the regret remains high compared to UCB, even though the models improve significantly over a random agent (see Figure \ref{fig:failure-greedy}c).

\noindent\textbf{Frequency Bias.}
The next prevalent failure mode is \emph{frequency bias}, which is characterized by repeatedly selecting the most frequently occurring action in the context, even when that action gives low reward.
To understand how the model's behavior is influenced by the frequency of actions, we construct prefix histories using a random policy, vary the number of repetitions of the last action in the context history (0 to 100) and record the entropy over all actions (see Figure \ref{fig:failure-copycat}a and c).
We provide details on the context generation in Appendix \ref{appendix:failures-modes-freq}.
To quantify frequency bias, we categorize an action as \emph{frequent} action $a_f = \arg\max_{a \in A} N_T(a)$, \emph{greedy} $a_g = \arg\max_{a \in \{a \in A : N_T(a) > 0\}} R_T(a)$, or \emph{other} if they are neither frequent nor greedy. Note that action is optimal with 10\% probability.
Subsequently, we compute the frequent $F_f$, greedy $F_g$ and other $F_o$ fractions as reported in Figure \ref{fig:failure-copycat} (see Appendix \ref{fig:failure-copycat} for definitions).
\begin{figure*}[h]
\noindent
\small
    \centering
\includegraphics[width=0.55\linewidth]{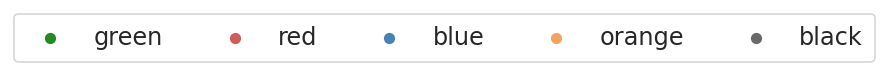}

\subfigure[2B: Correlations]{
    \includegraphics[width=0.31\linewidth]{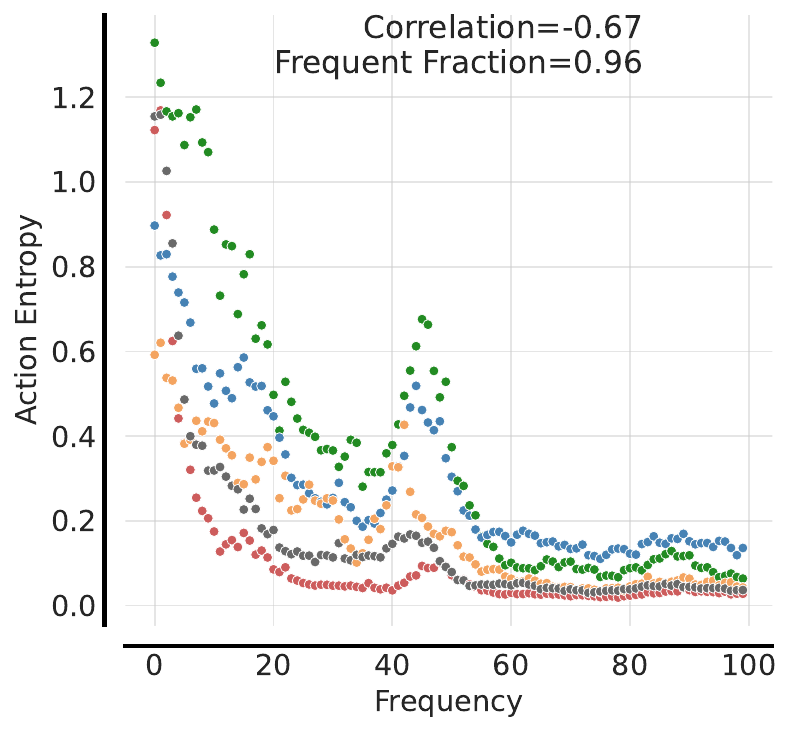}
}
~
\subfigure[Strategy per Window]{
    \includegraphics[width=0.27\linewidth]{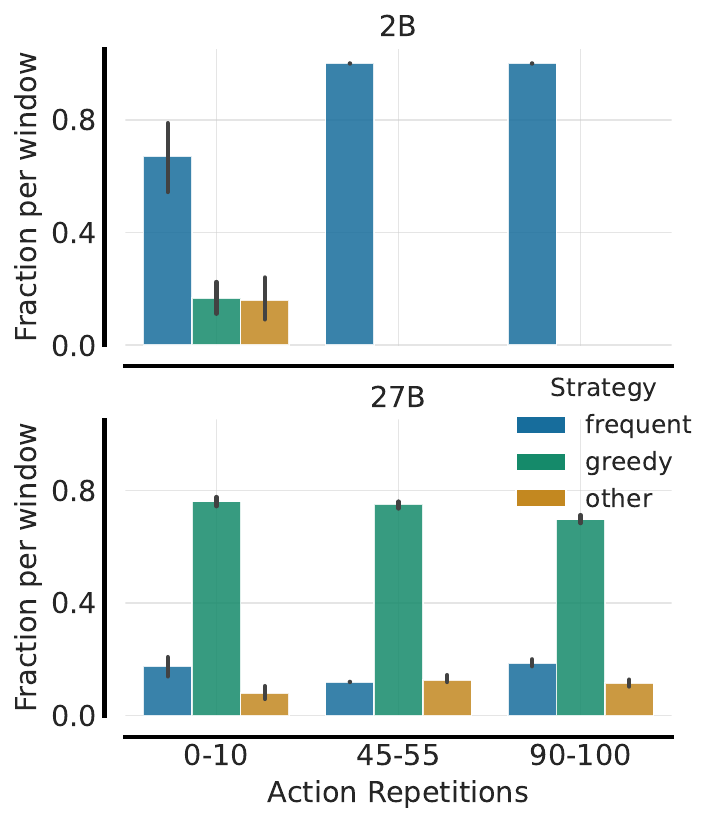}
}
~
\subfigure[27B: Correlations]{
    \includegraphics[width=0.31\linewidth]{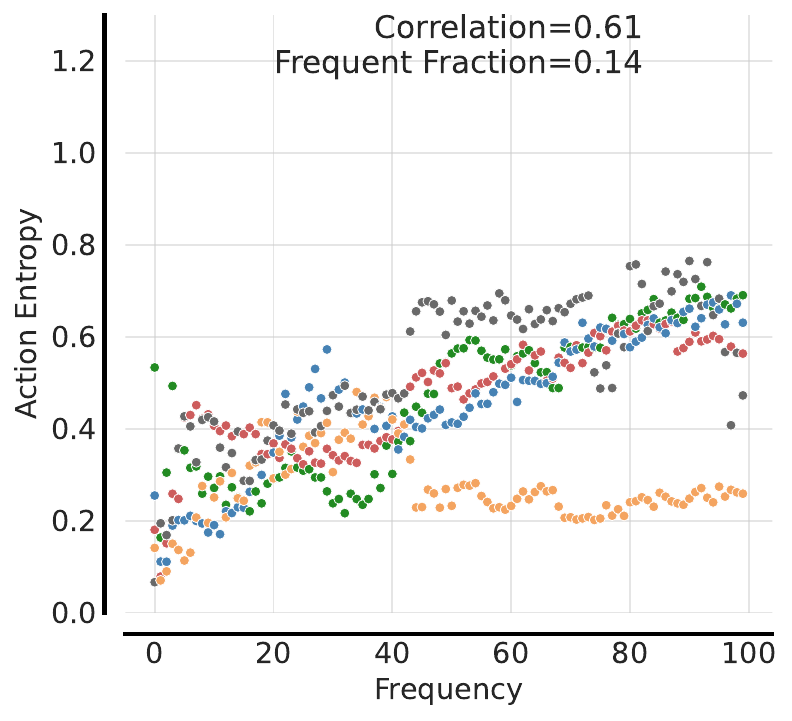}
}
\caption{Illustration of \textbf{Frequency Bias}. We plot the frequency of the repeated action in the context against the action entropy across all actions for 10 armed MABs. \textbf{(a)} Gemma2 2B heavily suffers from frequency bias, becoming more certain of the most frequent action, the more often it occurs in the context.
\textbf{(c)} Gemma2 27B overcomes the frequency bias, but instead behaves greedily. In \textbf{(b)} we show the action strategies for three repetition windows.}
\label{fig:failure-copycat}
\end{figure*}

\noindent Gemma2 2B heavily suffers from repeated actions, exhibiting a decreasing entropy with increasing repetitions (96\% $F_f$, see Figure \ref{fig:failure-copycat}a).
In contrast, 27B escapes the frequency bias ($14\%$, see Figure \ref{fig:failure-copycat}c) and interestingly becomes less certain of its action prediction with increasing repetitions. 
To examine this further, we show the bucketized fractions with 0-10, 45-55 and 90-100 repetitions for 2B and 27B in Figure \ref{fig:failure-copycat}b.
Indeed, for 2B $F_f$ keeps increasing with increasing repetitions. 
While 27B escapes the frequency bias it suffers heavily from greediness.
Similar biases have been identified in Behavior Cloning (BC) settings and termed \emph{copycat} bias \citep{wen2020fighting, schmied2024retrieval}.
This suggests that frequency bias is an artifact of supervised pre-training, and motivates the use of RL as a counter-measurement. 

\begin{wrapfigure}{r}{0.45\textwidth}
\vspace{-0.85cm}
\noindent
\small
    \centering
    \subfigure{
        \includegraphics[width=1\linewidth]{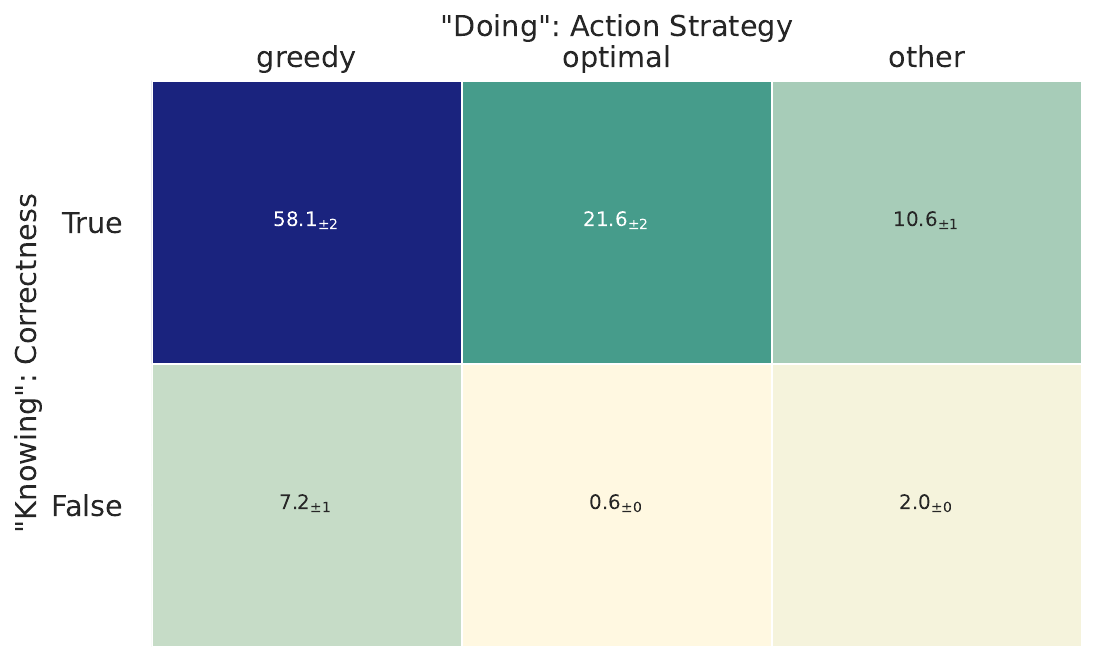}

    }
\caption{Confusion matrix for the \textbf{Knowing-Doing Gap} of Gemma2 27B. The agent “knows” how to solve the task (87\% correct rationales, sum of top row), but fails at "doing" (58\% greedy actions among correct rationales). See Figure \ref{fig:ucb-agent-knowing-doing}, for instructions and an agent response.}
\label{fig:failure-knowing-doing}
\end{wrapfigure}
\noindent\textbf{Knowing-Doing Gap.}
The \emph{knowing-doing gap} has been observed by \citet{paglieri2024balrog} and \citet{ruoss2024lmact}. %
To illustrate the gap in our setting, we first task Gemma2 27B to produce the UCB algorithm, to compute the relevant quantities accordingly ("knowing"), and finally to act according to the computed quantities ("doing", see Figure \ref{fig:ucb-agent-knowing-doing} for the instructions and an agent response).
We let Gemma2 27B interact with the environment (64 instances) for 50 timesteps with $G=2048$ per step, and extract the UCB quantities from the rationales.

\noindent To quantify "knowing", we compare the UCB values computed by the model against the real UCB values, and consider the rationale $z_{CoT}$ as correct if the arm with the highest UCB values match (see Appendix \ref{appendix:failures-modes-kd} for details).
\noindent To quantify "doing", we categorize the generated actions as \emph{optimal} action if the model selects the action with the highest UCB value, as \emph{greedy} if it selects the action with the highest UCB value among the set of actions tried so far, and as \emph{other} if the action is neither optimal nor greedy. 
Subsequently, we compute the percentages of greedy/optimal/other actions. 
The agent clearly knows how to solve the task, with 87\% of all rationales being correct (see Figure \ref{fig:failure-knowing-doing}). 
However, even for correctly computed rationales, the model often selects the greedy action (58\%) over the optimal action (21\%).
This discrepancy highlights the shortcomings of the LLM when it comes to "acting" even when "knowing" the algorithm.

\subsection{Effectiveness of RL Fine-Tuning} \label{sec:multi-armed-bandits}
\begin{figure}[htbp]
\noindent
\small
    \centering
\includegraphics[width=0.9\linewidth]{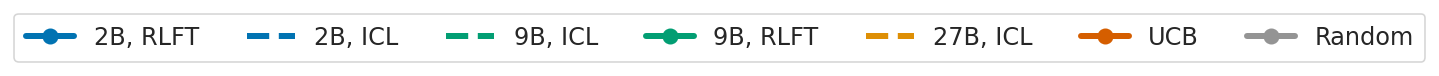} 
\vspace{-0.25cm}

\subfigure{
    \includegraphics[width=0.95\linewidth]{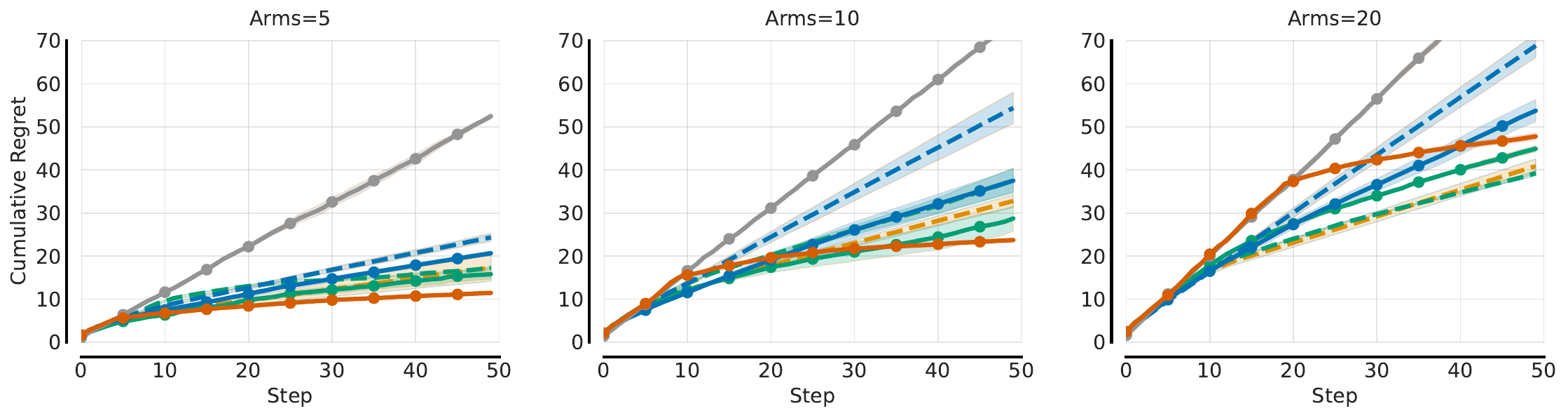} 
}
\caption{Main Comparison on \textbf{Gaussian MABs} button scenario in the medium noise ($\sigma=1$) setting. We compare cumulative regrets (lower is better) of classic baselines against ICL and RLFT performances for 5, 10, and 20 arms.
See Figure \ref{fig:mab-gaussian-karms-low-mednoise} for $\sigma=0.1$ and $\sigma=3$.}
\label{fig:mab-gaussian-karms}
\end{figure}
\noindent Next, we study the effects of RLFT on cumulative regret (w.r.t. optimal policy) and whether it alleviates the highlighted failure modes.
We fine-tune Gemma2 2B and 9B on self-generated CoT rationales for 30K updates with an (accumulated) batch size of 128. 
To avoid memorization of reward distributions, we maintain a pool of 512 MABs and randomly select a subset of 16 MABs per rollout.
We refer to Appendix \ref{appendix:exp-details} for training details and hyperparameters. 

\begin{wrapfigure}{r}{0.3\textwidth}
\vspace{-0.25cm}
\noindent
\small
    \centering
    \includegraphics[width=1\linewidth]{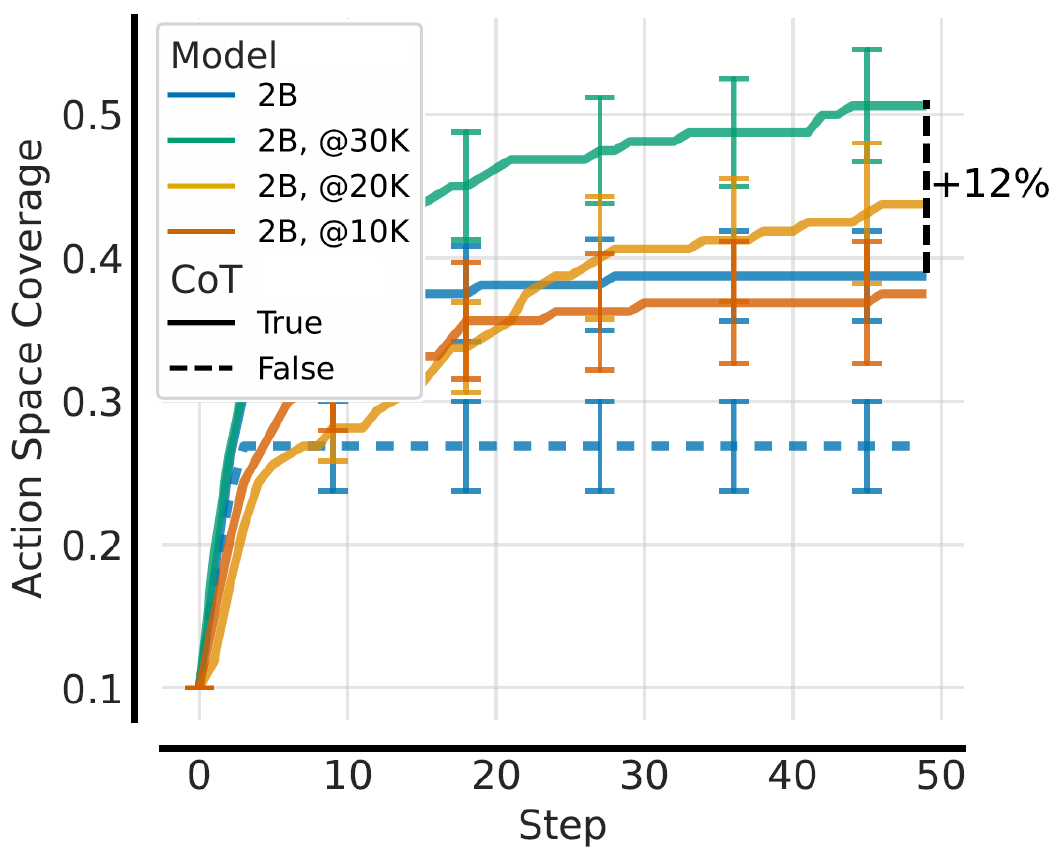}
\caption{Effect of RLFT on greediness for Gemma2 2B.}
\label{fig:failure-greedy-rlft}
\end{wrapfigure}
\noindent \textbf{RLFT lowers regret.}
In Figure \ref{fig:mab-gaussian-karms}, we report the cumulative regrets across model sizes and arms for a medium noise $\sigma=1.0$ scenario (see Appendix \ref{appendix:results-mabs} for low/high noise).
Across environments, the LLMs clearly outperform the random baseline and RLFT lowers regret for both 2B and 9B. 
For 2B, RLFT narrows the gap to it's larger counterparts and UCB. 
Similarly, RLFT lowers regret for Gemma2 9B.
Note, that the lower cumulative regret of Gemma2 9/27B compared to UCB after 50 environment steps in the 20 arms scenario is an artifact of the limited interaction steps, but the trends remain clear.
We repeat RLFT for CBs, and observe similar performance improvements for Gemma2 2B (see Appendix \ref{appendix:results-cbs}).
Consequently, reinforcing self-generated CoT rationales towards environment rewards improves performance on simple decision-making scenarios.

\noindent \textbf{RLFT mitigates greediness.} In Figure \ref{fig:failure-greedy-rlft}, we report the action coverage for 2B after RLFT at different numbers of gradient steps (10K, 20K, 30K).
Indeed, we observe that RLFT results in increased action coverage (+12\%) after 30K updates.
Interestingly, we first observe a decrease (at 10K) followed by an increase in action coverage (20K, 30K). 
We observe similar effects for the 20 arms scenario (see Figure \ref{fig:failure-greedy-rlft-20arms}).
Via RLFT the agent learns to explore and consequently mitigates greediness.

\noindent \textbf{RLFT counteracts frequency bias.}
We find that RLFT counteracts frequency bias (Figure \ref{fig:failure-copycat-postrlft-both}).
In particular, for 0-10 repetitions we observe a strong decrease in the fraction of frequent actions ($70\%\rightarrow35\%$) and increase in "other" actions ($8\%\rightarrow35\%$).
However, $F_f$ remains elevated for high repetitions.
Consequently, RLFT counteracts frequency bias, but does not fully alleviate it.

\subsection{Effect of Exploration Mechanisms }\label{sec:exps-exp-interventions}
\noindent While RLFT decreases regret and encourages exploration, we observe that the action coverage remains sub-optimal (see Section \ref{sec:multi-armed-bandits}).
For RLFT, we relied solely on the exploration properties for CoT reasoning.
However, in RL it is common practice to employ additional exploration strategies  \citep{mnih2015human, schulman2017proximal, haarnoja2018soft}.
Therefore, we study the effects of classic exploration mechanisms and LLM-specific strategies to encourage exploration.
We compare: \textbf{(1)} try-all actions initially similar to UCB, \textbf{(2)} $\epsilon$-greedy, \textbf{(3)} context randomization, \textbf{(4)} context summary similar to \citet{krishnamurthy2024can} and \citet{nie2024evolve}, \textbf{(5)} self-correction similar to \citet{kumar2024training}, \textbf{(6)} self-consistency \citep{wang2022self}, and \textbf{(7)} exploration bonus. 
We provide details on mechanisms in Appendix \ref{appendix:exp-mechanisms}.
\begin{figure*}[h]
\noindent
\small
    \centering
\includegraphics[width=1.0\linewidth]{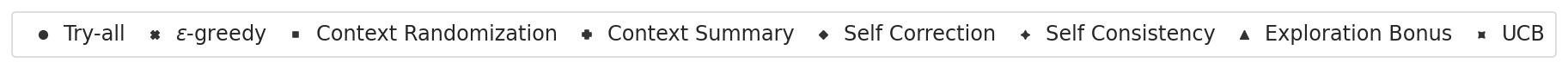} 

\includegraphics[width=1\linewidth]{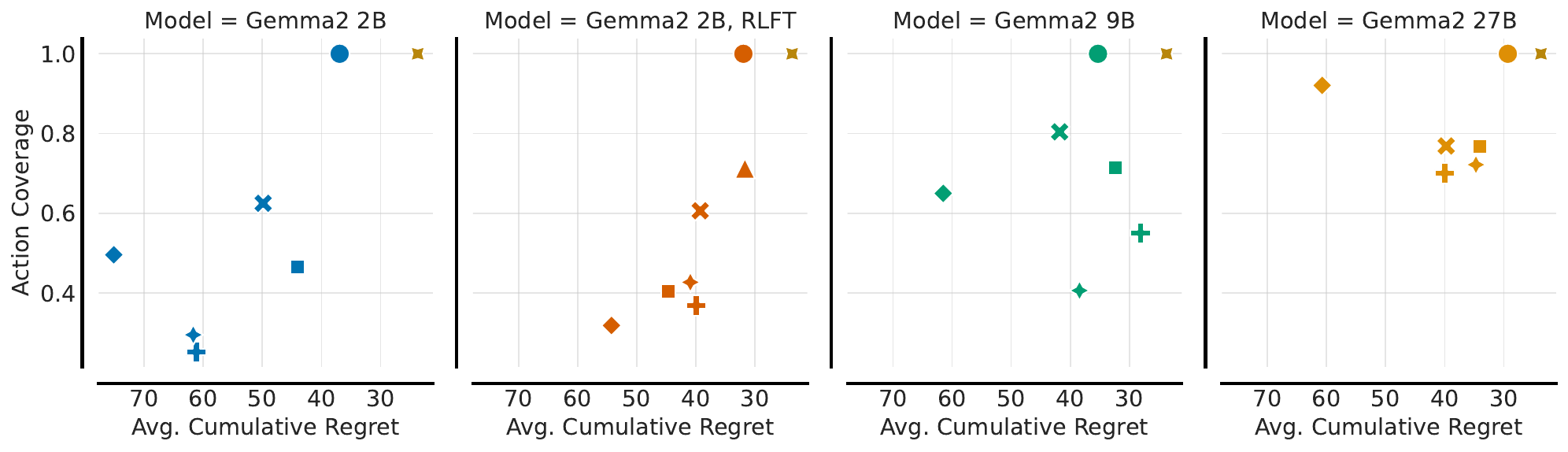}
\caption{Effect of \textbf{exploration mechanisms} on action coverage and cumulative regret. }
\label{fig:exp-mechanisms}
\end{figure*}

\noindent Across model scales, we observe that the mechanisms result in varied effects on action coverage (see Figure \ref{fig:exp-mechanisms}).
First, we find that the simple \textit{try-all} strategy, which reduces the need for additional exploration by trying all actions, results in the biggest performance improvements. 
Gemma2 27B almost closes the gap to the optimal UCB agent. 
This suggests that only given sufficient information about the (sub-)optimality of actions, LLMs are able to select actions accordingly, underscoring their exploration shortcomings. 
Second, we observe that RLFT lowers regret and improves exploration across different exploration mechanisms. 
Most importantly, a simple \emph{exploration bonus} (+1 reward for untried actions during RLFT), significantly increases exploration ($50\% \rightarrow 70\%$) and lowers regret towards the expert compared to regular RLFT.
This highlights the importance of reward shaping for fine-tuning LLMs in decision-making scenarios to elucidate a desired behavior.

\subsection{Ablations}\label{sec:ablations}

\begin{figure*}[h]
\noindent
\small
    \centering
\subfigure[Tic-tac-toe: Opponents]{
    \includegraphics[width=0.29\linewidth]{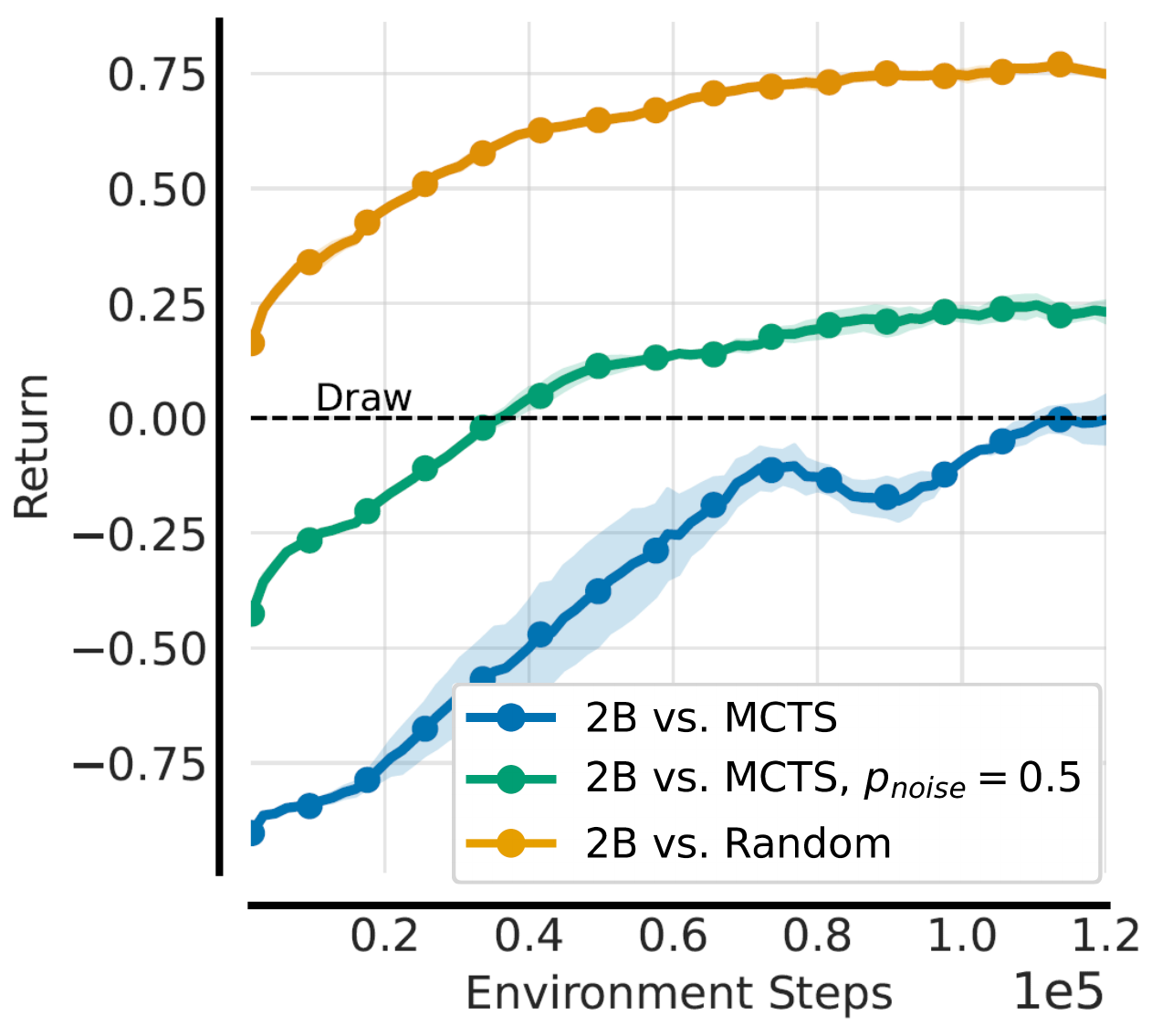} 
}
~
\subfigure[MAB: Effect of CoT]{
    \includegraphics[width=0.28\linewidth]{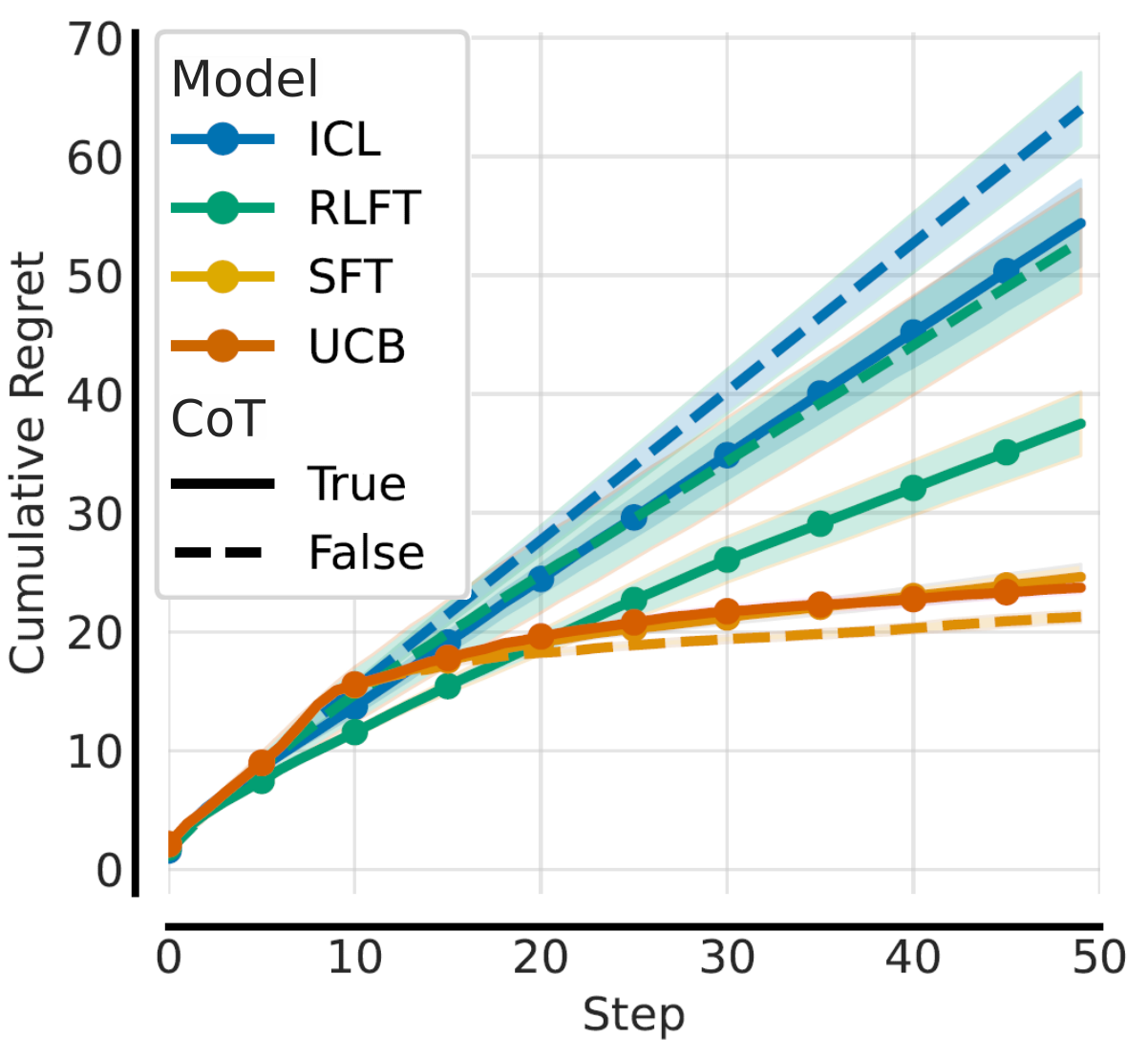}
}
~
\subfigure[MAB: "Thinking" Time]{
    \includegraphics[width=0.285\linewidth]{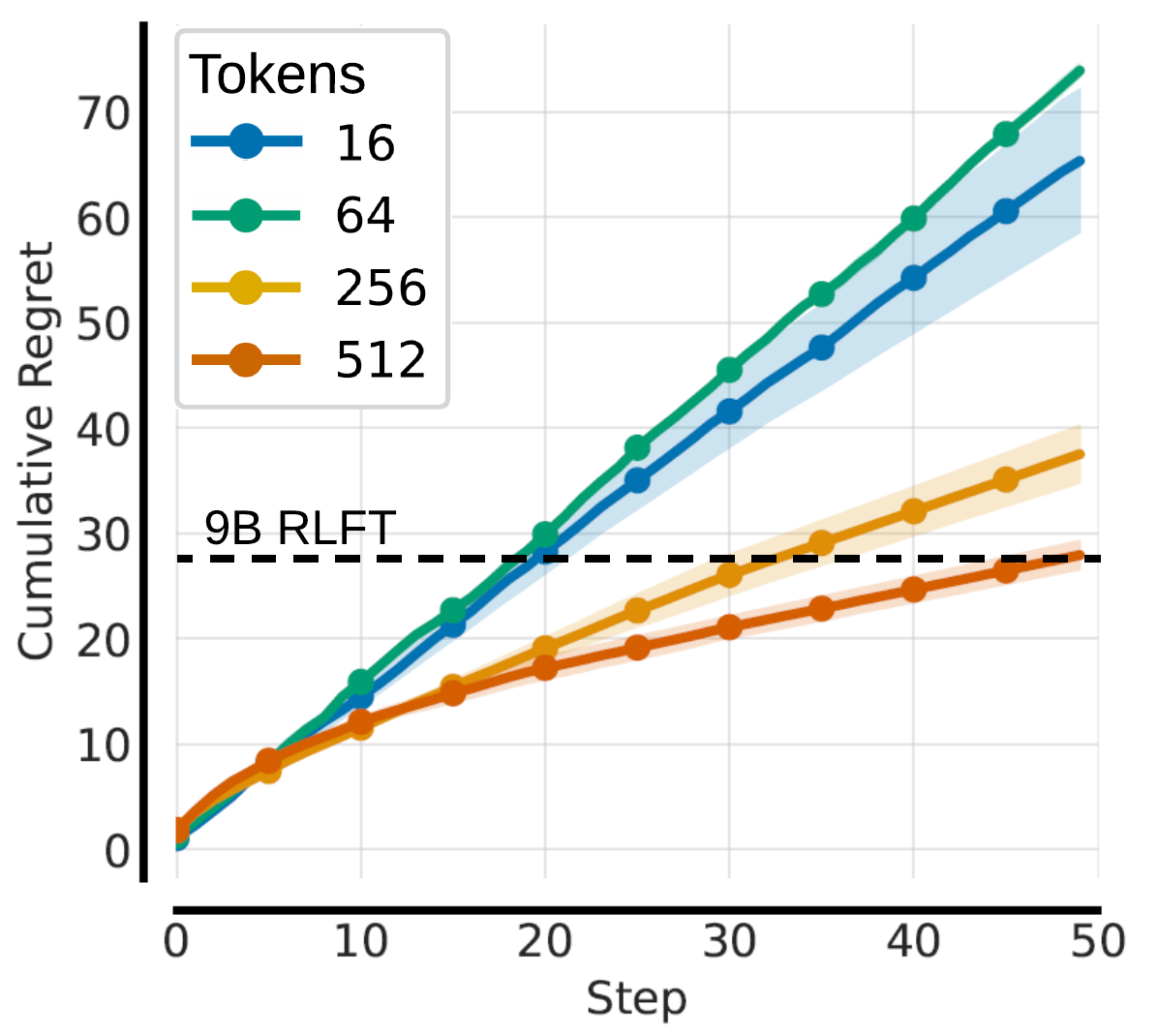}
}
\caption{Ablations. \textbf{(a)} Effect of RLFT in Tic-tac-toe from \citet{ruoss2024lmact}. \textbf{(b)} Effect of CoT on ICL, RLFT and SFT (expert data) performance on MABs. \textbf{(c)} Effect of increasing the number of "thinking" tokens to generate during RLFT.}
\label{fig:tictactoe}
\end{figure*}

\noindent\textbf{RLFT in Tic-tac-toe.}
To investigate the efficacy of RLFT in stateful environments, we evaluate on Tic-tac-toe from \citet{ruoss2024lmact}, in which frontier models struggle to achieve strong performance (see Appendix \ref{appendix:exp-details} for training details).
We fine-tune against three opponents: a random agent, Monte Carlo Tree Search (MCTS) \citep{coulom2006efficient}, and noisy MCTS (50\% of actions selected at random).
We find that RLFT significantly enhances the win-rate of Gemma2 2B against all opponent compared to ICL (see Figure \ref{fig:tictactoe}a).
Against the random agent, RLFT elevates the average return from 0.15 (i.e., winning 15\% of games) to 0.75.
Notably, the agent even manages to draw against the optimal MCTS baseline ($-0.95\rightarrow0.0$), underscoring the effectiveness of RLFT for decision-making. 
However, for high performance it is essential to provide the legal actions in the context (see Figure \ref{fig:tictactoe-legal-actions}).

\noindent \textbf{Importance of CoT for RLFT.}
CoT reasoning is critical for ICL performance (see Figure \ref{fig:failure-greedy}), but the question remains how CoT influences RLFT. 
Therefore, we run RLFT on Gemma2 2B on the 10 arms Gaussian MAB both w/ and w/o CoT (see Figure \ref{fig:tictactoe}b, RLFT). 
Indeed, without CoT, RLFT barely attains the performance of ICL w/ CoT. 
This highlights the function of CoT as a vital exploration and rationalization mechanism for decision-making. 

\noindent \textbf{Expert Behavior Cloning vs. Thought Cloning.}
BC is a prevalent approach in sequence models for decision-making \citep{pomerleau1988alvinn,brohan2022rt,brohan2023rt} and relies on expert datasets. 
Consequently, we construct two UCB expert dataset comprising 32K rollouts (1.6M transitions) across different MABs either w/ or w/o CoT (see Figure \ref{fig:sft-data-instances}) and perform SFT on them.
Notably, both SFT variants successfully mimic the expert achieving comparable regret to UCB (see Figure \ref{fig:tictactoe}b, SFT).
This result underscores the effectiveness of expert data in decision-making, echoing recent findings in reasoning tasks \citep{muennighoff2025s1}. 

\noindent \textbf{Effect of "Thinking" Time.}
Finally, we investigate the effect of giving the agent more/less time to "think" during RLFT by varying the generation budget $G$ (see Figure \ref{fig:tictactoe}c and Appendix \ref{appendix:ablations}).
Decreasing $G$ results in poor performance, as the agent is unable to rationalize its decisions. 
Increasing $G$ to 512 improves performance to the level of 9B w/ RLFT.
The agent effectively leverages the additional tokens, which reflects recent observations in mathematical reasoning \citep{guo2025deepseek}. 
However, when increasing $G$, rollout generation can make up the majority of the training time due to the multi-step nature of decision-making tasks (e.g., for $H=50$, $G=500$ the agent generates 25K tokens).

\section{Conclusion}
In this work, we study \textit{why} LLMs perform sub-optimally in decision-making scenarios.
We examine three prevalent failure modes of LLMs: greediness, frequency bias, and the knowing-doing gap.
We show that RLFT on CoT rationales mitigates greediness, counteracts frequency bias, and improves final performance. 
While RLFT improves the exploration, it remains sub-optimal compared to bandit algorithms. 
Therefore, we investigate a variety of mechanisms, such as $\epsilon$-greedy or self-consistency, to improve exploration.
Models act near-optimally if provided with sufficient information underscoring their shortcomings in exploration.
Finally, we highlight the importance of reward shaping for RLFT. 

\noindent\textbf{Limitations \& Future Work.}
We focused our evaluation on the Gemma2 series and small-to-medium scale models.
While we expect that our findings transfer to larger models, we deem research into frontier models important. 
Moreover, our MAB experiments were conducted with a limited horizon of 50 environment steps, which is sufficient for 5 and 10 arms, but insufficient for 20 arms. 
For future work, we believe that evaluating the exploration abilities of LLM agents is particularly interesting in environments that require targeted exploration towards an end-goal. 
First, this includes other stateful environments from \citet{paglieri2024balrog} and \citet{ruoss2024lmact}, such as Crafter \citep{hafner2021benchmarking}.
Second, we deem a systematic investigation into exploration abilities of LLMs in existing agentic benchmarks \citep{mialon2023gaia,he2024webvoyager,zhou2023webarena} interesting.
In our ablation studies, we found that LLMs benefit from additional "thinking" time and believe that allowing for a larger generation budget will become increasingly important for agentic scenarios, especially for scenarios with high-stakes decisions (e.g., economics or ethics).
We deem investigations into such high-stakes scenarios fruitful for future work. 
While increasing "thinking" time improves performance, it comes with excessive computational cost at training time due to the rollout generation and the multi-step nature of decision-making. 
Therefore, modern recurrent architectures \citep{de2024griffin, beck2025xlstm} that allow for faster inference may be promising alternatives for decision-making.

\bibliographystyle{abbrvnat}
\nobibliography*
\bibliography{refs}

\begin{thebibliography}{109}
\providecommand{\natexlab}[1]{#1}
\providecommand{\url}[1]{\texttt{#1}}
\expandafter\ifx\csname urlstyle\endcsname\relax
  \providecommand{\doi}[1]{doi: #1}\else
  \providecommand{\doi}{doi: \begingroup \urlstyle{rm}\Url}\fi

\bibitem[Agarwal et~al.(2021)Agarwal, Schwarzer, Castro, Courville, and
  Bellemare]{agarwal2021deep}
R.~Agarwal, M.~Schwarzer, P.~S. Castro, A.~C. Courville, and M.~Bellemare.
\newblock Deep reinforcement learning at the edge of the statistical precipice.
\newblock \emph{Advances in neural information processing systems},
  34:\penalty0 29304--29320, 2021.

\bibitem[Ahmadian et~al.(2024)Ahmadian, Cremer, Gall{\'e}, Fadaee, Kreutzer,
  {\"U}st{\"u}n, and Hooker]{ahmadian2024back}
A.~Ahmadian, C.~Cremer, M.~Gall{\'e}, M.~Fadaee, J.~Kreutzer, A.~{\"U}st{\"u}n,
  and S.~Hooker.
\newblock Back to basics: Revisiting reinforce style optimization for learning
  from human feedback in llms.
\newblock \emph{arXiv preprint arXiv:2402.14740}, 2024.

\bibitem[Arjona-Medina et~al.(2019)Arjona-Medina, Gillhofer, Widrich,
  Unterthiner, Brandstetter, and Hochreiter]{arjona2019rudder}
J.~A. Arjona-Medina, M.~Gillhofer, M.~Widrich, T.~Unterthiner, J.~Brandstetter,
  and S.~Hochreiter.
\newblock Rudder: Return decomposition for delayed rewards.
\newblock \emph{Advances in Neural Information Processing Systems}, 32, 2019.

\bibitem[Auer(2002)]{auer2002using}
P.~Auer.
\newblock Using confidence bounds for exploitation-exploration trade-offs.
\newblock \emph{Journal of Machine Learning Research}, 3\penalty0
  (Nov):\penalty0 397--422, 2002.

\bibitem[Beck et~al.(2025)Beck, P{\"o}ppel, Spanring, Auer, Prudnikova, Kopp,
  Klambauer, Brandstetter, and Hochreiter]{beck2025xlstm}
M.~Beck, K.~P{\"o}ppel, M.~Spanring, A.~Auer, O.~Prudnikova, M.~Kopp,
  G.~Klambauer, J.~Brandstetter, and S.~Hochreiter.
\newblock xlstm: Extended long short-term memory.
\newblock \emph{Advances in Neural Information Processing Systems},
  37:\penalty0 107547--107603, 2025.

\bibitem[Bellemare et~al.(2013)Bellemare, Naddaf, Veness, and
  Bowling]{bellemare2013arcade}
M.~G. Bellemare, Y.~Naddaf, J.~Veness, and M.~Bowling.
\newblock The arcade learning environment: An evaluation platform for general
  agents.
\newblock \emph{Journal of Artificial Intelligence Research}, 47:\penalty0
  253--279, 2013.

\bibitem[Bellemare et~al.(2020)Bellemare, Candido, Castro, Gong, Machado,
  Moitra, Ponda, and Wang]{bellemare2020autonomous}
M.~G. Bellemare, S.~Candido, P.~S. Castro, J.~Gong, M.~C. Machado, S.~Moitra,
  S.~S. Ponda, and Z.~Wang.
\newblock Autonomous navigation of stratospheric balloons using reinforcement
  learning.
\newblock \emph{Nature}, 588\penalty0 (7836):\penalty0 77--82, 2020.

\bibitem[Brohan et~al.(2022)Brohan, Brown, Carbajal, Chebotar, Dabis, Finn,
  Gopalakrishnan, Hausman, Herzog, Hsu, et~al.]{brohan2022rt}
A.~Brohan, N.~Brown, J.~Carbajal, Y.~Chebotar, J.~Dabis, C.~Finn,
  K.~Gopalakrishnan, K.~Hausman, A.~Herzog, J.~Hsu, et~al.
\newblock Rt-1: Robotics transformer for real-world control at scale.
\newblock \emph{arXiv preprint arXiv:2212.06817}, 2022.

\bibitem[Brohan et~al.(2023)Brohan, Brown, Carbajal, Chebotar, Chen,
  Choromanski, Ding, Driess, Dubey, Finn, et~al.]{brohan2023rt}
A.~Brohan, N.~Brown, J.~Carbajal, Y.~Chebotar, X.~Chen, K.~Choromanski,
  T.~Ding, D.~Driess, A.~Dubey, C.~Finn, et~al.
\newblock Rt-2: Vision-language-action models transfer web knowledge to robotic
  control.
\newblock \emph{arXiv preprint arXiv:2307.15818}, 2023.

\bibitem[Brown et~al.(2020{\natexlab{a}})Brown, Mann, Ryder, Subbiah, Kaplan,
  Dhariwal, Neelakantan, Shyam, Sastry, Askell, et~al.]{Brown:20}
T.~Brown, B.~Mann, N.~Ryder, M.~Subbiah, J.~D. Kaplan, P.~Dhariwal,
  A.~Neelakantan, P.~Shyam, G.~Sastry, A.~Askell, et~al.
\newblock Language models are few-shot learners.
\newblock \emph{Advances in neural information processing systems},
  33:\penalty0 1877--1901, 2020{\natexlab{a}}.

\bibitem[Brown et~al.(2020{\natexlab{b}})Brown, Mann, Ryder, Subbiah, Kaplan,
  Dhariwal, Neelakantan, Shyam, Sastry, Askell, et~al.]{brown2020language}
T.~Brown, B.~Mann, N.~Ryder, M.~Subbiah, J.~D. Kaplan, P.~Dhariwal,
  A.~Neelakantan, P.~Shyam, G.~Sastry, A.~Askell, et~al.
\newblock Language models are few-shot learners.
\newblock \emph{Advances in neural information processing systems},
  33:\penalty0 1877--1901, 2020{\natexlab{b}}.

\bibitem[Burda et~al.(2018)Burda, Edwards, Storkey, and
  Klimov]{burda2018exploration}
Y.~Burda, H.~Edwards, A.~Storkey, and O.~Klimov.
\newblock Exploration by random network distillation.
\newblock \emph{arXiv preprint arXiv:1810.12894}, 2018.

\bibitem[Chan et~al.(2022)Chan, Santoro, Lampinen, Wang, Singh, Richemond,
  McClelland, and Hill]{Chan:22}
S.~Chan, A.~Santoro, A.~K. Lampinen, J.~Wang, A.~Singh, P.~H. Richemond, J.~L.
  McClelland, and F.~Hill.
\newblock Data distributional properties drive emergent in-context learning in
  transformers.
\newblock In S.~Koyejo, S.~Mohamed, A.~Agarwal, D.~Belgrave, K.~Cho, and A.~Oh,
  editors, \emph{Advances in Neural Information Processing Systems 35: Annual
  Conference on Neural Information Processing Systems 2022, NeurIPS 2022, New
  Orleans, LA, USA, November 28 - December 9, 2022}, 2022.

\bibitem[Chen et~al.(2023)Chen, Wang, Dong, Du, Yan, Yang, Li, Zhao, Qin,
  Rajmohan, et~al.]{chen2023introspective}
L.~Chen, L.~Wang, H.~Dong, Y.~Du, J.~Yan, F.~Yang, S.~Li, P.~Zhao, S.~Qin,
  S.~Rajmohan, et~al.
\newblock Introspective tips: Large language model for in-context decision
  making.
\newblock \emph{arXiv preprint arXiv:2305.11598}, 2023.

\bibitem[Christiano et~al.(2017)Christiano, Leike, Brown, Martic, Legg, and
  Amodei]{christiano2017deep}
P.~F. Christiano, J.~Leike, T.~Brown, M.~Martic, S.~Legg, and D.~Amodei.
\newblock Deep reinforcement learning from human preferences.
\newblock \emph{Advances in neural information processing systems}, 30, 2017.

\bibitem[Chu et~al.(2011)Chu, Li, Reyzin, and Schapire]{chu2011contextual}
W.~Chu, L.~Li, L.~Reyzin, and R.~Schapire.
\newblock Contextual bandits with linear payoff functions.
\newblock In \emph{Proceedings of the fourteenth international conference on
  artificial intelligence and statistics}, pages 208--214. JMLR Workshop and
  Conference Proceedings, 2011.

\bibitem[Cobbe et~al.(2021)Cobbe, Kosaraju, Bavarian, Chen, Jun, Kaiser,
  Plappert, Tworek, Hilton, Nakano, et~al.]{cobbe2021training}
K.~Cobbe, V.~Kosaraju, M.~Bavarian, M.~Chen, H.~Jun, L.~Kaiser, M.~Plappert,
  J.~Tworek, J.~Hilton, R.~Nakano, et~al.
\newblock Training verifiers to solve math word problems.
\newblock \emph{arXiv preprint arXiv:2110.14168}, 2021.

\bibitem[Coulom(2006)]{coulom2006efficient}
R.~Coulom.
\newblock Efficient selectivity and backup operators in monte-carlo tree
  search.
\newblock In \emph{International conference on computers and games}, pages
  72--83. Springer, 2006.

\bibitem[Cundy and Ermon(2023)]{cundy2023sequencematch}
C.~Cundy and S.~Ermon.
\newblock Sequencematch: Imitation learning for autoregressive sequence
  modelling with backtracking.
\newblock \emph{arXiv preprint arXiv:2306.05426}, 2023.

\bibitem[De et~al.(2024)De, Smith, Fernando, Botev, Cristian-Muraru, Gu,
  Haroun, Berrada, Chen, Srinivasan, et~al.]{de2024griffin}
S.~De, S.~L. Smith, A.~Fernando, A.~Botev, G.~Cristian-Muraru, A.~Gu,
  R.~Haroun, L.~Berrada, Y.~Chen, S.~Srinivasan, et~al.
\newblock Griffin: Mixing gated linear recurrences with local attention for
  efficient language models.
\newblock \emph{arXiv preprint arXiv:2402.19427}, 2024.

\bibitem[Degrave et~al.(2022)Degrave, Felici, Buchli, Neunert, Tracey,
  Carpanese, Ewalds, Hafner, Abdolmaleki, de~Las~Casas,
  et~al.]{degrave2022magnetic}
J.~Degrave, F.~Felici, J.~Buchli, M.~Neunert, B.~Tracey, F.~Carpanese,
  T.~Ewalds, R.~Hafner, A.~Abdolmaleki, D.~de~Las~Casas, et~al.
\newblock Magnetic control of tokamak plasmas through deep reinforcement
  learning.
\newblock \emph{Nature}, 602\penalty0 (7897):\penalty0 414--419, 2022.

\bibitem[Duan et~al.(2016)Duan, Schulman, Chen, Bartlett, Sutskever, and
  Abbeel]{duan2016rl}
Y.~Duan, J.~Schulman, X.~Chen, P.~L. Bartlett, I.~Sutskever, and P.~Abbeel.
\newblock Rl2: Fast reinforcement learning via slow reinforcement learning.
\newblock \emph{arXiv preprint arXiv:1611.02779}, 2016.

\bibitem[Dubey et~al.(2024)Dubey, Jauhri, Pandey, Kadian, Al-Dahle, Letman,
  Mathur, Schelten, Yang, Fan, et~al.]{dubey2024llama}
A.~Dubey, A.~Jauhri, A.~Pandey, A.~Kadian, A.~Al-Dahle, A.~Letman, A.~Mathur,
  A.~Schelten, A.~Yang, A.~Fan, et~al.
\newblock The llama 3 herd of models.
\newblock \emph{arXiv preprint arXiv:2407.21783}, 2024.

\bibitem[Durante et~al.(2024)Durante, Huang, Wake, Gong, Park, Sarkar, Taori,
  Noda, Terzopoulos, Choi, et~al.]{durante2024agent}
Z.~Durante, Q.~Huang, N.~Wake, R.~Gong, J.~S. Park, B.~Sarkar, R.~Taori,
  Y.~Noda, D.~Terzopoulos, Y.~Choi, et~al.
\newblock Agent ai: Surveying the horizons of multimodal interaction.
\newblock \emph{arXiv preprint arXiv:2401.03568}, 2024.

\bibitem[Ecoffet et~al.(2019)Ecoffet, Huizinga, Lehman, Stanley, and
  Clune]{ecoffet2019go}
A.~Ecoffet, J.~Huizinga, J.~Lehman, K.~O. Stanley, and J.~Clune.
\newblock Go-explore: a new approach for hard-exploration problems.
\newblock \emph{arXiv preprint arXiv:1901.10995}, 2019.

\bibitem[Finn et~al.(2017)Finn, Abbeel, and Levine]{Finn:17}
C.~Finn, P.~Abbeel, and S.~Levine.
\newblock Model-agnostic meta-learning for fast adaptation of deep networks.
\newblock In \emph{International conference on machine learning}, pages
  1126--1135. PMLR, 2017.

\bibitem[Flennerhag et~al.(2019)Flennerhag, Rusu, Pascanu, Visin, Yin, and
  Hadsell]{Flennerhag:19}
S.~Flennerhag, A.~A. Rusu, R.~Pascanu, F.~Visin, H.~Yin, and R.~Hadsell.
\newblock Meta-learning with warped gradient descent.
\newblock \emph{arXiv preprint arXiv:1909.00025}, 2019.

\bibitem[Groth et~al.(2021)Groth, Wulfmeier, Vezzani, Dasagi, Hertweck, Hafner,
  Heess, and Riedmiller]{groth2021curiosity}
O.~Groth, M.~Wulfmeier, G.~Vezzani, V.~Dasagi, T.~Hertweck, R.~Hafner,
  N.~Heess, and M.~Riedmiller.
\newblock Is curiosity all you need? on the utility of emergent behaviours from
  curious exploration.
\newblock \emph{arXiv e-prints}, pages arXiv--2109, 2021.

\bibitem[Guo et~al.(2025)Guo, Yang, Zhang, Song, Zhang, Xu, Zhu, Ma, Wang, Bi,
  et~al.]{guo2025deepseek}
D.~Guo, D.~Yang, H.~Zhang, J.~Song, R.~Zhang, R.~Xu, Q.~Zhu, S.~Ma, P.~Wang,
  X.~Bi, et~al.
\newblock Deepseek-r1: Incentivizing reasoning capability in llms via
  reinforcement learning.
\newblock \emph{arXiv preprint arXiv:2501.12948}, 2025.

\bibitem[Haarnoja et~al.(2018)Haarnoja, Zhou, Hartikainen, Tucker, Ha, Tan,
  Kumar, Zhu, Gupta, Abbeel, et~al.]{haarnoja2018soft}
T.~Haarnoja, A.~Zhou, K.~Hartikainen, G.~Tucker, S.~Ha, J.~Tan, V.~Kumar,
  H.~Zhu, A.~Gupta, P.~Abbeel, et~al.
\newblock Soft actor-critic algorithms and applications.
\newblock \emph{arXiv preprint arXiv:1812.05905}, 2018.

\bibitem[Hafner(2021)]{hafner2021benchmarking}
D.~Hafner.
\newblock Benchmarking the spectrum of agent capabilities.
\newblock \emph{arXiv preprint arXiv:2109.06780}, 2021.

\bibitem[Harper and Konstan(2015)]{harper2015movielens}
F.~M. Harper and J.~A. Konstan.
\newblock The movielens datasets: History and context.
\newblock \emph{Acm transactions on interactive intelligent systems (tiis)},
  5\penalty0 (4):\penalty0 1--19, 2015.

\bibitem[He et~al.(2024)He, Yao, Ma, Yu, Dai, Zhang, Lan, and
  Yu]{he2024webvoyager}
H.~He, W.~Yao, K.~Ma, W.~Yu, Y.~Dai, H.~Zhang, Z.~Lan, and D.~Yu.
\newblock Webvoyager: Building an end-to-end web agent with large multimodal
  models.
\newblock \emph{arXiv preprint arXiv:2401.13919}, 2024.

\bibitem[Hendrycks et~al.(2021)Hendrycks, Burns, Kadavath, Arora, Basart, Tang,
  Song, and Steinhardt]{hendrycks2021measuring}
D.~Hendrycks, C.~Burns, S.~Kadavath, A.~Arora, S.~Basart, E.~Tang, D.~Song, and
  J.~Steinhardt.
\newblock Measuring mathematical problem solving with the math dataset.
\newblock \emph{arXiv preprint arXiv:2103.03874}, 2021.

\bibitem[Hessel et~al.(2018)Hessel, Modayil, Van~Hasselt, Schaul, Ostrovski,
  Dabney, Horgan, Piot, Azar, and Silver]{hessel2018rainbow}
M.~Hessel, J.~Modayil, H.~Van~Hasselt, T.~Schaul, G.~Ostrovski, W.~Dabney,
  D.~Horgan, B.~Piot, M.~Azar, and D.~Silver.
\newblock Rainbow: Combining improvements in deep reinforcement learning.
\newblock In \emph{Proceedings of the AAAI conference on artificial
  intelligence}, volume~32, 2018.

\bibitem[Hochreiter and Schmidhuber(1997)]{Hochreiter:97}
S.~Hochreiter and J.~Schmidhuber.
\newblock Long short-term memory.
\newblock \emph{Neural computation}, 9\penalty0 (8):\penalty0 1735--1780, 1997.

\bibitem[Hochreiter et~al.(2001)Hochreiter, Younger, and
  Conwell]{Hochreiter:01}
S.~Hochreiter, A.~S. Younger, and P.~R. Conwell.
\newblock Learning to learn using gradient descent.
\newblock In \emph{Artificial Neural Networks—ICANN 2001: International
  Conference Vienna, Austria, August 21--25, 2001 Proceedings 11}, pages
  87--94. Springer, 2001.

\bibitem[Hu et~al.(2022)Hu, Shen, Wallis, Allen-Zhu, Li, Wang, Wang, Chen,
  et~al.]{hu2022lora}
E.~J. Hu, Y.~Shen, P.~Wallis, Z.~Allen-Zhu, Y.~Li, S.~Wang, L.~Wang, W.~Chen,
  et~al.
\newblock Lora: Low-rank adaptation of large language models.
\newblock \emph{ICLR}, 1\penalty0 (2):\penalty0 3, 2022.

\bibitem[Hu and Clune(2023)]{hu2023thought}
S.~Hu and J.~Clune.
\newblock Thought cloning: Learning to think while acting by imitating human
  thinking.
\newblock \emph{Advances in Neural Information Processing Systems},
  36:\penalty0 44451--44469, 2023.

\bibitem[Kirsch et~al.(2019)Kirsch, van Steenkiste, and Schmidhuber]{Kirsch:19}
L.~Kirsch, S.~van Steenkiste, and J.~Schmidhuber.
\newblock Improving generalization in meta reinforcement learning using learned
  objectives.
\newblock \emph{arXiv preprint arXiv:1910.04098}, 2019.

\bibitem[Kirsch et~al.(2022)Kirsch, Harrison, Sohl-Dickstein, and
  Metz]{Kirsch:22}
L.~Kirsch, J.~Harrison, J.~Sohl-Dickstein, and L.~Metz.
\newblock General-purpose in-context learning by meta-learning transformers.
\newblock \emph{arXiv preprint arXiv:2212.04458}, 2022.

\bibitem[Kirsch et~al.(2023)Kirsch, Harrison, Freeman, Sohl-Dickstein, and
  Schmidhuber]{Kirsch:23}
L.~Kirsch, J.~Harrison, C.~Freeman, J.~Sohl-Dickstein, and J.~Schmidhuber.
\newblock Towards general-purpose in-context learning agents.
\newblock In \emph{NeurIPS 2023 Workshop on Generalization in Planning}, 2023.

\bibitem[Klissarov et~al.(2023)Klissarov, D'Oro, Sodhani, Raileanu, Bacon,
  Vincent, Zhang, and Henaff]{klissarov2023motif}
M.~Klissarov, P.~D'Oro, S.~Sodhani, R.~Raileanu, P.-L. Bacon, P.~Vincent,
  A.~Zhang, and M.~Henaff.
\newblock Motif: Intrinsic motivation from artificial intelligence feedback.
\newblock \emph{arXiv preprint arXiv:2310.00166}, 2023.

\bibitem[Klissarov et~al.(2024)Klissarov, Henaff, Raileanu, Sodhani, Vincent,
  Zhang, Bacon, Precup, Machado, and D'Oro]{klissarov2024maestromotif}
M.~Klissarov, M.~Henaff, R.~Raileanu, S.~Sodhani, P.~Vincent, A.~Zhang, P.-L.
  Bacon, D.~Precup, M.~C. Machado, and P.~D'Oro.
\newblock Maestromotif: Skill design from artificial intelligence feedback.
\newblock \emph{arXiv preprint arXiv:2412.08542}, 2024.

\bibitem[Krishnamurthy et~al.(2024)Krishnamurthy, Harris, Foster, Zhang, and
  Slivkins]{krishnamurthy2024can}
A.~Krishnamurthy, K.~Harris, D.~J. Foster, C.~Zhang, and A.~Slivkins.
\newblock Can large language models explore in-context?
\newblock \emph{arXiv preprint arXiv:2403.15371}, 2024.

\bibitem[Kumar et~al.(2024)Kumar, Zhuang, Agarwal, Su, Co-Reyes, Singh, Baumli,
  Iqbal, Bishop, Roelofs, et~al.]{kumar2024training}
A.~Kumar, V.~Zhuang, R.~Agarwal, Y.~Su, J.~D. Co-Reyes, A.~Singh, K.~Baumli,
  S.~Iqbal, C.~Bishop, R.~Roelofs, et~al.
\newblock Training language models to self-correct via reinforcement learning.
\newblock \emph{arXiv preprint arXiv:2409.12917}, 2024.

\bibitem[Laskin et~al.(2022)Laskin, Wang, Oh, Parisotto, Spencer, Steigerwald,
  Strouse, Hansen, Filos, Brooks, et~al.]{Laskin:22}
M.~Laskin, L.~Wang, J.~Oh, E.~Parisotto, S.~Spencer, R.~Steigerwald,
  D.~Strouse, S.~Hansen, A.~Filos, E.~Brooks, et~al.
\newblock In-context reinforcement learning with algorithm distillation.
\newblock \emph{arXiv preprint arXiv:2210.14215}, 2022.

\bibitem[Lattimore and Szepesv{\'a}ri(2020)]{lattimore2020bandit}
T.~Lattimore and C.~Szepesv{\'a}ri.
\newblock \emph{Bandit algorithms}.
\newblock Cambridge University Press, 2020.

\bibitem[Lee et~al.(2022)Lee, Nachum, Yang, Lee, Freeman, Xu, Guadarrama,
  Fischer, Jang, Michalewski, et~al.]{Lee:22}
K.-H. Lee, O.~Nachum, M.~Yang, L.~Lee, D.~Freeman, W.~Xu, S.~Guadarrama,
  I.~Fischer, E.~Jang, H.~Michalewski, et~al.
\newblock Multi-game decision transformers.
\newblock \emph{arXiv preprint arXiv:2205.15241}, 2022.

\bibitem[Li et~al.(2022)Li, Choi, Chung, Kushman, Schrittwieser, Leblond,
  Eccles, Keeling, Gimeno, Dal~Lago, et~al.]{li2022competition}
Y.~Li, D.~Choi, J.~Chung, N.~Kushman, J.~Schrittwieser, R.~Leblond, T.~Eccles,
  J.~Keeling, F.~Gimeno, A.~Dal~Lago, et~al.
\newblock Competition-level code generation with alphacode.
\newblock \emph{Science}, 378\penalty0 (6624):\penalty0 1092--1097, 2022.

\bibitem[Lu et~al.(2024)Lu, Hu, and Clune]{lu2024intelligent}
C.~Lu, S.~Hu, and J.~Clune.
\newblock Intelligent go-explore: Standing on the shoulders of giant foundation
  models.
\newblock \emph{arXiv preprint arXiv:2405.15143}, 2024.

\bibitem[Mialon et~al.(2023)Mialon, Fourrier, Wolf, LeCun, and
  Scialom]{mialon2023gaia}
G.~Mialon, C.~Fourrier, T.~Wolf, Y.~LeCun, and T.~Scialom.
\newblock Gaia: a benchmark for general ai assistants.
\newblock In \emph{The Twelfth International Conference on Learning
  Representations}, 2023.

\bibitem[Mirchandani et~al.(2023)Mirchandani, Xia, Florence, Ichter, Driess,
  Arenas, Rao, Sadigh, and Zeng]{mirchandani2023large}
S.~Mirchandani, F.~Xia, P.~Florence, B.~Ichter, D.~Driess, M.~G. Arenas,
  K.~Rao, D.~Sadigh, and A.~Zeng.
\newblock Large language models as general pattern machines.
\newblock \emph{arXiv preprint arXiv:2307.04721}, 2023.

\bibitem[Mishra et~al.(2018)Mishra, Rohaninejad, Chen, and Abbeel]{Mishra:18}
N.~Mishra, M.~Rohaninejad, X.~Chen, and P.~Abbeel.
\newblock A simple neural attentive meta-learner.
\newblock In \emph{6th International Conference on Learning Representations,
  {ICLR} 2018, Vancouver, BC, Canada, April 30 - May 3, 2018, Conference Track
  Proceedings}. OpenReview.net, 2018.
\newblock URL \url{https://openreview.net/forum?id=B1DmUzWAW}.

\bibitem[Mnih et~al.(2015)Mnih, Kavukcuoglu, Silver, Rusu, Veness, Bellemare,
  Graves, Riedmiller, Fidjeland, Ostrovski, et~al.]{mnih2015human}
V.~Mnih, K.~Kavukcuoglu, D.~Silver, A.~A. Rusu, J.~Veness, M.~G. Bellemare,
  A.~Graves, M.~Riedmiller, A.~K. Fidjeland, G.~Ostrovski, et~al.
\newblock Human-level control through deep reinforcement learning.
\newblock \emph{nature}, 518\penalty0 (7540):\penalty0 529--533, 2015.

\bibitem[Monea et~al.(2024)Monea, Bosselut, Brantley, and Artzi]{monea2024llms}
G.~Monea, A.~Bosselut, K.~Brantley, and Y.~Artzi.
\newblock Llms are in-context reinforcement learners.
\newblock \emph{arXiv preprint arXiv:2410.05362}, 2024.

\bibitem[Muennighoff et~al.(2025)Muennighoff, Yang, Shi, Li, Fei-Fei,
  Hajishirzi, Zettlemoyer, Liang, Cand{\`e}s, and Hashimoto]{muennighoff2025s1}
N.~Muennighoff, Z.~Yang, W.~Shi, X.~L. Li, L.~Fei-Fei, H.~Hajishirzi,
  L.~Zettlemoyer, P.~Liang, E.~Cand{\`e}s, and T.~Hashimoto.
\newblock s1: Simple test-time scaling.
\newblock \emph{arXiv preprint arXiv:2501.19393}, 2025.

\bibitem[Nie et~al.(2024)Nie, Su, Chang, Lee, Chi, Le, and Chen]{nie2024evolve}
A.~Nie, Y.~Su, B.~Chang, J.~N. Lee, E.~H. Chi, Q.~V. Le, and M.~Chen.
\newblock Evolve: Evaluating and optimizing llms for exploration.
\newblock \emph{arXiv preprint arXiv:2410.06238}, 2024.

\bibitem[Oudeyer et~al.(2007)Oudeyer, Kaplan, and Hafner]{oudeyer2007intrinsic}
P.-Y. Oudeyer, F.~Kaplan, and V.~V. Hafner.
\newblock Intrinsic motivation systems for autonomous mental development.
\newblock \emph{IEEE transactions on evolutionary computation}, 11\penalty0
  (2):\penalty0 265--286, 2007.

\bibitem[Ouyang et~al.(2022)Ouyang, Wu, Jiang, Almeida, Wainwright, Mishkin,
  Zhang, Agarwal, Slama, Ray, et~al.]{ouyang2022training}
L.~Ouyang, J.~Wu, X.~Jiang, D.~Almeida, C.~Wainwright, P.~Mishkin, C.~Zhang,
  S.~Agarwal, K.~Slama, A.~Ray, et~al.
\newblock Training language models to follow instructions with human feedback.
\newblock \emph{Advances in neural information processing systems},
  35:\penalty0 27730--27744, 2022.

\bibitem[Paglieri et~al.(2024)Paglieri, Cupia{\l}, Coward, Piterbarg, Wolczyk,
  Khan, Pignatelli, Kuci{\'n}ski, Pinto, Fergus, et~al.]{paglieri2024balrog}
D.~Paglieri, B.~Cupia{\l}, S.~Coward, U.~Piterbarg, M.~Wolczyk, A.~Khan,
  E.~Pignatelli, {\L}.~Kuci{\'n}ski, L.~Pinto, R.~Fergus, et~al.
\newblock Balrog: Benchmarking agentic llm and vlm reasoning on games.
\newblock \emph{arXiv preprint arXiv:2411.13543}, 2024.

\bibitem[Pathak et~al.(2017)Pathak, Agrawal, Efros, and
  Darrell]{pathak2017curiosity}
D.~Pathak, P.~Agrawal, A.~A. Efros, and T.~Darrell.
\newblock Curiosity-driven exploration by self-supervised prediction.
\newblock In \emph{International conference on machine learning}, pages
  2778--2787. PMLR, 2017.

\bibitem[Pomerleau(1988)]{pomerleau1988alvinn}
D.~A. Pomerleau.
\newblock Alvinn: An autonomous land vehicle in a neural network.
\newblock \emph{Advances in neural information processing systems}, 1, 1988.

\bibitem[Radford et~al.(2019)Radford, Wu, Child, Luan, Amodei, Sutskever,
  et~al.]{radford2019language}
A.~Radford, J.~Wu, R.~Child, D.~Luan, D.~Amodei, I.~Sutskever, et~al.
\newblock Language models are unsupervised multitask learners.
\newblock \emph{OpenAI blog}, 1\penalty0 (8):\penalty0 9, 2019.

\bibitem[Raileanu and Rockt{\"a}schel(2020)]{raileanu2020ride}
R.~Raileanu and T.~Rockt{\"a}schel.
\newblock Ride: Rewarding impact-driven exploration for procedurally-generated
  environments.
\newblock \emph{arXiv preprint arXiv:2002.12292}, 2020.

\bibitem[Ramesh et~al.(2024)Ramesh, Hu, Chaimalas, Mehta, Sessa, Ammar, and
  Bogunovic]{ramesh2024group}
S.~S. Ramesh, Y.~Hu, I.~Chaimalas, V.~Mehta, P.~G. Sessa, H.~B. Ammar, and
  I.~Bogunovic.
\newblock Group robust preference optimization in reward-free rlhf.
\newblock \emph{arXiv preprint arXiv:2405.20304}, 2024.

\bibitem[Rao et~al.(2021)Rao, Sadeghi, Hasenclever, Wulfmeier, Zambelli,
  Vezzani, Tirumala, Aytar, Merel, Heess, et~al.]{raolearning}
D.~Rao, F.~Sadeghi, L.~Hasenclever, M.~Wulfmeier, M.~Zambelli, G.~Vezzani,
  D.~Tirumala, Y.~Aytar, J.~Merel, N.~Heess, et~al.
\newblock Learning transferable motor skills with hierarchical latent mixture
  policies.
\newblock In \emph{International Conference on Learning Representations}, 2021.

\bibitem[Raparthy et~al.(2023)Raparthy, Hambro, Kirk, Henaff, and
  Raileanu]{Raparthy:23}
S.~C. Raparthy, E.~Hambro, R.~Kirk, M.~Henaff, and R.~Raileanu.
\newblock Generalization to new sequential decision making tasks with
  in-context learning, 2023.

\bibitem[Reed et~al.(2022)Reed, Zolna, Parisotto, Colmenarejo, Novikov,
  Barth-Maron, Gimenez, Sulsky, Kay, Springenberg, et~al.]{reed2022generalist}
S.~Reed, K.~Zolna, E.~Parisotto, S.~G. Colmenarejo, A.~Novikov, G.~Barth-Maron,
  M.~Gimenez, Y.~Sulsky, J.~Kay, J.~T. Springenberg, et~al.
\newblock A generalist agent.
\newblock \emph{arXiv preprint arXiv:2205.06175}, 2022.

\bibitem[Ruoss et~al.(2024)Ruoss, Pardo, Chan, Li, Mnih, and
  Genewein]{ruoss2024lmact}
A.~Ruoss, F.~Pardo, H.~Chan, B.~Li, V.~Mnih, and T.~Genewein.
\newblock Lmact: A benchmark for in-context imitation learning with long
  multimodal demonstrations.
\newblock \emph{arXiv preprint arXiv:2412.01441}, 2024.

\bibitem[Saab et~al.(2024)Saab, Tu, Weng, Tanno, Stutz, Wulczyn, Zhang,
  Strother, Park, Vedadi, et~al.]{saab2024capabilities}
K.~Saab, T.~Tu, W.-H. Weng, R.~Tanno, D.~Stutz, E.~Wulczyn, F.~Zhang,
  T.~Strother, C.~Park, E.~Vedadi, et~al.
\newblock Capabilities of gemini models in medicine.
\newblock \emph{arXiv preprint arXiv:2404.18416}, 2024.

\bibitem[Santoro et~al.(2016)Santoro, Bartunov, Botvinick, Wierstra, and
  Lillicrap]{Santoro:16}
A.~Santoro, S.~Bartunov, M.~Botvinick, D.~Wierstra, and T.~Lillicrap.
\newblock Meta-learning with memory-augmented neural networks.
\newblock In \emph{International conference on machine learning}, pages
  1842--1850. PMLR, 2016.

\bibitem[Schmidhuber(1987)]{Schmidhuber:87}
J.~Schmidhuber.
\newblock Evolutionary principles in self-referential learning. on learning now
  to learn: The meta-meta-meta...-hook.
\newblock Diploma thesis, Technische Universitat Munchen, Germany, 14 May 1987.

\bibitem[Schmidhuber(1991{\natexlab{a}})]{schmidhuber1991curious}
J.~Schmidhuber.
\newblock Curious model-building control systems.
\newblock In \emph{Proc. international joint conference on neural networks},
  pages 1458--1463, 1991{\natexlab{a}}.

\bibitem[Schmidhuber(1991{\natexlab{b}})]{schmidhuber1991possibility}
J.~Schmidhuber.
\newblock A possibility for implementing curiosity and boredom in
  model-building neural controllers.
\newblock In \emph{Proc. of the international conference on simulation of
  adaptive behavior: From animals to animats}, pages 222--227,
  1991{\natexlab{b}}.

\bibitem[Schmied et~al.(2023)Schmied, Hofmarcher, Paischer, Pascanu, and
  Hochreiter]{schmied2023learning}
T.~Schmied, M.~Hofmarcher, F.~Paischer, R.~Pascanu, and S.~Hochreiter.
\newblock Learning to modulate pre-trained models in rl.
\newblock \emph{Advances in Neural Information Processing Systems},
  36:\penalty0 38231--38265, 2023.

\bibitem[Schmied et~al.(2024{\natexlab{a}})Schmied, Adler, Patil, Beck,
  P{\"o}ppel, Brandstetter, Klambauer, Pascanu, and
  Hochreiter]{schmied2024large}
T.~Schmied, T.~Adler, V.~Patil, M.~Beck, K.~P{\"o}ppel, J.~Brandstetter,
  G.~Klambauer, R.~Pascanu, and S.~Hochreiter.
\newblock A large recurrent action model: xlstm enables fast inference for
  robotics tasks.
\newblock \emph{arXiv preprint arXiv:2410.22391}, 2024{\natexlab{a}}.

\bibitem[Schmied et~al.(2024{\natexlab{b}})Schmied, Paischer, Patil,
  Hofmarcher, Pascanu, and Hochreiter]{schmied2024retrieval}
T.~Schmied, F.~Paischer, V.~Patil, M.~Hofmarcher, R.~Pascanu, and
  S.~Hochreiter.
\newblock Retrieval-augmented decision transformer: External memory for
  in-context rl.
\newblock \emph{arXiv preprint arXiv:2410.07071}, 2024{\natexlab{b}}.

\bibitem[Schulman et~al.(2015)Schulman, Moritz, Levine, Jordan, and
  Abbeel]{schulman2015high}
J.~Schulman, P.~Moritz, S.~Levine, M.~Jordan, and P.~Abbeel.
\newblock High-dimensional continuous control using generalized advantage
  estimation.
\newblock \emph{arXiv preprint arXiv:1506.02438}, 2015.

\bibitem[Schulman et~al.(2017)Schulman, Wolski, Dhariwal, Radford, and
  Klimov]{schulman2017proximal}
J.~Schulman, F.~Wolski, P.~Dhariwal, A.~Radford, and O.~Klimov.
\newblock Proximal policy optimization algorithms.
\newblock \emph{arXiv preprint arXiv:1707.06347}, 2017.

\bibitem[Shao et~al.(2024)Shao, Wang, Zhu, Xu, Song, Bi, Zhang, Zhang, Li, Wu,
  et~al.]{shao2024deepseekmath}
Z.~Shao, P.~Wang, Q.~Zhu, R.~Xu, J.~Song, X.~Bi, H.~Zhang, M.~Zhang, Y.~Li,
  Y.~Wu, et~al.
\newblock Deepseekmath: Pushing the limits of mathematical reasoning in open
  language models.
\newblock \emph{arXiv preprint arXiv:2402.03300}, 2024.

\bibitem[Shazeer and Stern(2018)]{shazeer2018adafactor}
N.~Shazeer and M.~Stern.
\newblock Adafactor: Adaptive learning rates with sublinear memory cost.
\newblock In \emph{International Conference on Machine Learning}, pages
  4596--4604. PMLR, 2018.

\bibitem[Shinn et~al.(2023)Shinn, Cassano, Labash, Gopinath, Narasimhan, and
  Yao]{shinn2023reflexion}
N.~Shinn, F.~Cassano, B.~Labash, A.~Gopinath, K.~Narasimhan, and S.~Yao.
\newblock Reflexion: Language agents with verbal reinforcement learning.(2023).
\newblock \emph{arXiv preprint cs.AI/2303.11366}, 2023.

\bibitem[Silver et~al.(2016)Silver, Huang, Maddison, Guez, Sifre, Van
  Den~Driessche, Schrittwieser, Antonoglou, Panneershelvam, Lanctot,
  et~al.]{silver2016mastering}
D.~Silver, A.~Huang, C.~J. Maddison, A.~Guez, L.~Sifre, G.~Van Den~Driessche,
  J.~Schrittwieser, I.~Antonoglou, V.~Panneershelvam, M.~Lanctot, et~al.
\newblock Mastering the game of go with deep neural networks and tree search.
\newblock \emph{nature}, 529\penalty0 (7587):\penalty0 484--489, 2016.

\bibitem[Slivkins et~al.(2019)]{slivkins2019introduction}
A.~Slivkins et~al.
\newblock Introduction to multi-armed bandits.
\newblock \emph{Foundations and Trends{\textregistered} in Machine Learning},
  12\penalty0 (1-2):\penalty0 1--286, 2019.

\bibitem[Still and Precup(2012)]{still2012information}
S.~Still and D.~Precup.
\newblock An information-theoretic approach to curiosity-driven reinforcement
  learning.
\newblock \emph{Theory in Biosciences}, 131\penalty0 (3):\penalty0 139--148,
  2012.

\bibitem[Tassa et~al.(2018)Tassa, Doron, Muldal, Erez, Li, Casas, Budden,
  Abdolmaleki, Merel, Lefrancq, et~al.]{tassa2018deepmind}
Y.~Tassa, Y.~Doron, A.~Muldal, T.~Erez, Y.~Li, D.~d.~L. Casas, D.~Budden,
  A.~Abdolmaleki, J.~Merel, A.~Lefrancq, et~al.
\newblock Deepmind control suite.
\newblock \emph{arXiv preprint arXiv:1801.00690}, 2018.

\bibitem[Team et~al.(2023{\natexlab{a}})Team, Bauer, Baumli, Baveja, Behbahani,
  Bhoopchand, Bradley-Schmieg, Chang, Clay, Collister, Dasagi, Gonzalez,
  Gregor, Hughes, Kashem, Loks-Thompson, Openshaw, Parker-Holder, Pathak,
  Nieves, Rakicevic, Rockt{\"a}schel, Schroecker, Sygnowski, Tuyls, York,
  Zacherl, and Zhang]{Team2023HumanTimescaleAI}
A.~A. Team, J.~Bauer, K.~Baumli, S.~Baveja, F.~M.~P. Behbahani, A.~Bhoopchand,
  N.~Bradley-Schmieg, M.~Chang, N.~Clay, A.~Collister, V.~Dasagi, L.~Gonzalez,
  K.~Gregor, E.~Hughes, S.~Kashem, M.~Loks-Thompson, H.~Openshaw,
  J.~Parker-Holder, S.~Pathak, N.~P. Nieves, N.~Rakicevic, T.~Rockt{\"a}schel,
  Y.~Schroecker, J.~Sygnowski, K.~Tuyls, S.~York, A.~Zacherl, and L.~M. Zhang.
\newblock Human-timescale adaptation in an open-ended task space.
\newblock In \emph{International Conference on Machine Learning},
  2023{\natexlab{a}}.

\bibitem[Team et~al.(2023{\natexlab{b}})Team, Anil, Borgeaud, Alayrac, Yu,
  Soricut, Schalkwyk, Dai, Hauth, Millican, et~al.]{team2023gemini}
G.~Team, R.~Anil, S.~Borgeaud, J.-B. Alayrac, J.~Yu, R.~Soricut, J.~Schalkwyk,
  A.~M. Dai, A.~Hauth, K.~Millican, et~al.
\newblock Gemini: a family of highly capable multimodal models.
\newblock \emph{arXiv preprint arXiv:2312.11805}, 2023{\natexlab{b}}.

\bibitem[Team et~al.(2024{\natexlab{a}})Team, Georgiev, Lei, Burnell, Bai,
  Gulati, Tanzer, Vincent, Pan, Wang, et~al.]{team2024gemini}
G.~Team, P.~Georgiev, V.~I. Lei, R.~Burnell, L.~Bai, A.~Gulati, G.~Tanzer,
  D.~Vincent, Z.~Pan, S.~Wang, et~al.
\newblock Gemini 1.5: Unlocking multimodal understanding across millions of
  tokens of context.
\newblock \emph{arXiv preprint arXiv:2403.05530}, 2024{\natexlab{a}}.

\bibitem[Team et~al.(2024{\natexlab{b}})Team, Mesnard, Hardin, Dadashi,
  Bhupatiraju, Pathak, Sifre, Rivi{\`e}re, Kale, Love, et~al.]{team2024gemma}
G.~Team, T.~Mesnard, C.~Hardin, R.~Dadashi, S.~Bhupatiraju, S.~Pathak,
  L.~Sifre, M.~Rivi{\`e}re, M.~S. Kale, J.~Love, et~al.
\newblock Gemma: Open models based on gemini research and technology.
\newblock \emph{arXiv preprint arXiv:2403.08295}, 2024{\natexlab{b}}.

\bibitem[Team et~al.(2024{\natexlab{c}})Team, Riviere, Pathak, Sessa, Hardin,
  Bhupatiraju, Hussenot, Mesnard, Shahriari, Ram{\'e}, et~al.]{team2024gemma2}
G.~Team, M.~Riviere, S.~Pathak, P.~G. Sessa, C.~Hardin, S.~Bhupatiraju,
  L.~Hussenot, T.~Mesnard, B.~Shahriari, A.~Ram{\'e}, et~al.
\newblock Gemma 2: Improving open language models at a practical size.
\newblock \emph{arXiv preprint arXiv:2408.00118}, 2024{\natexlab{c}}.

\bibitem[Team et~al.(2024{\natexlab{d}})Team, Modi, Veerubhotla, Rysbek, Huber,
  Wiltshire, Veprek, Gillick, Kasenberg, Ahmed, et~al.]{team2024learnlm}
L.~Team, A.~Modi, A.~S. Veerubhotla, A.~Rysbek, A.~Huber, B.~Wiltshire,
  B.~Veprek, D.~Gillick, D.~Kasenberg, D.~Ahmed, et~al.
\newblock Learnlm: Improving gemini for learning.
\newblock \emph{arXiv preprint arXiv:2412.16429}, 2024{\natexlab{d}}.

\bibitem[Tirumala et~al.(2025)Tirumala, Wulfmeier, Moran, Huang, Humplik,
  Lever, Haarnoja, Hasenclever, Byravan, Batchelor, sreendra, Patel, Gwira,
  Nori, Riedmiller, and Heess]{pmlr-v270-tirumala25a}
D.~Tirumala, M.~Wulfmeier, B.~Moran, S.~Huang, J.~Humplik, G.~Lever,
  T.~Haarnoja, L.~Hasenclever, A.~Byravan, N.~Batchelor, N.~sreendra, K.~Patel,
  M.~Gwira, F.~Nori, M.~Riedmiller, and N.~Heess.
\newblock Learning robot soccer from egocentric vision with deep reinforcement
  learning.
\newblock In P.~Agrawal, O.~Kroemer, and W.~Burgard, editors, \emph{Proceedings
  of The 8th Conference on Robot Learning}, volume 270 of \emph{Proceedings of
  Machine Learning Research}, pages 165--184. PMLR, 06--09 Nov 2025.
\newblock URL \url{https://proceedings.mlr.press/v270/tirumala25a.html}.

\bibitem[Vinyals et~al.(2019)Vinyals, Babuschkin, Czarnecki, Mathieu, Dudzik,
  Chung, Choi, Powell, Ewalds, Georgiev, et~al.]{vinyals2019grandmaster}
O.~Vinyals, I.~Babuschkin, W.~M. Czarnecki, M.~Mathieu, A.~Dudzik, J.~Chung,
  D.~H. Choi, R.~Powell, T.~Ewalds, P.~Georgiev, et~al.
\newblock Grandmaster level in starcraft ii using multi-agent reinforcement
  learning.
\newblock \emph{nature}, 575\penalty0 (7782):\penalty0 350--354, 2019.

\bibitem[Wang et~al.(2016)Wang, Kurth-Nelson, Tirumala, Soyer, Leibo, Munos,
  Blundell, Kumaran, and Botvinick]{Wang:16}
J.~X. Wang, Z.~Kurth-Nelson, D.~Tirumala, H.~Soyer, J.~Z. Leibo, R.~Munos,
  C.~Blundell, D.~Kumaran, and M.~Botvinick.
\newblock Learning to reinforcement learn.
\newblock \emph{arXiv preprint arXiv:1611.05763}, 2016.

\bibitem[Wang et~al.(2022)Wang, Wei, Schuurmans, Le, Chi, Narang, Chowdhery,
  and Zhou]{wang2022self}
X.~Wang, J.~Wei, D.~Schuurmans, Q.~Le, E.~Chi, S.~Narang, A.~Chowdhery, and
  D.~Zhou.
\newblock Self-consistency improves chain of thought reasoning in language
  models.
\newblock \emph{arXiv preprint arXiv:2203.11171}, 2022.

\bibitem[Wang et~al.(2025)Wang, Yue, and Chen]{wang2025critique}
Y.~Wang, X.~Yue, and W.~Chen.
\newblock Critique fine-tuning: Learning to critique is more effective than
  learning to imitate.
\newblock \emph{arXiv preprint arXiv:2501.17703}, 2025.

\bibitem[Wei et~al.(2022)Wei, Wang, Schuurmans, Bosma, Xia, Chi, Le, Zhou,
  et~al.]{wei2022chain}
J.~Wei, X.~Wang, D.~Schuurmans, M.~Bosma, F.~Xia, E.~Chi, Q.~V. Le, D.~Zhou,
  et~al.
\newblock Chain-of-thought prompting elicits reasoning in large language
  models.
\newblock \emph{Advances in neural information processing systems},
  35:\penalty0 24824--24837, 2022.

\bibitem[Welleck et~al.(2022)Welleck, Lu, West, Brahman, Shen, Khashabi, and
  Choi]{welleck2022generating}
S.~Welleck, X.~Lu, P.~West, F.~Brahman, T.~Shen, D.~Khashabi, and Y.~Choi.
\newblock Generating sequences by learning to self-correct.
\newblock \emph{arXiv preprint arXiv:2211.00053}, 2022.

\bibitem[Wen et~al.(2020)Wen, Lin, Darrell, Jayaraman, and
  Gao]{wen2020fighting}
C.~Wen, J.~Lin, T.~Darrell, D.~Jayaraman, and Y.~Gao.
\newblock Fighting copycat agents in behavioral cloning from observation
  histories.
\newblock \emph{Advances in Neural Information Processing Systems},
  33:\penalty0 2564--2575, 2020.

\bibitem[Williams(1992)]{williams1992simple}
R.~J. Williams.
\newblock Simple statistical gradient-following algorithms for connectionist
  reinforcement learning.
\newblock \emph{Machine learning}, 8:\penalty0 229--256, 1992.

\bibitem[Wulfmeier et~al.(2024)Wulfmeier, Bloesch, Vieillard, Ahuja,
  Bornschein, Huang, Sokolov, Barnes, Desjardins, Bewley, Bechtle,
  Springenberg, Momchev, Bachem, Geist, and Riedmiller]{wulfmeier2024irl}
M.~Wulfmeier, M.~Bloesch, N.~Vieillard, A.~Ahuja, J.~Bornschein, S.~Huang,
  A.~Sokolov, M.~Barnes, G.~Desjardins, A.~Bewley, S.~M.~E. Bechtle, J.~T.
  Springenberg, N.~Momchev, O.~Bachem, M.~Geist, and M.~Riedmiller.
\newblock Imitating language via scalable inverse reinforcement learning.
\newblock In A.~Globerson, L.~Mackey, D.~Belgrave, A.~Fan, U.~Paquet,
  J.~Tomczak, and C.~Zhang, editors, \emph{Advances in Neural Information
  Processing Systems}, volume~37, pages 90714--90735. Curran Associates, Inc.,
  2024.
\newblock URL
  \url{https://proceedings.neurips.cc/paper_files/paper/2024/file/a5036c166e44b731f214f41813364d01-Paper-Conference.pdf}.

\bibitem[Yao et~al.(2022)Yao, Zhao, Yu, Du, Shafran, Narasimhan, and
  Cao]{yao2022react}
S.~Yao, J.~Zhao, D.~Yu, N.~Du, I.~Shafran, K.~Narasimhan, and Y.~Cao.
\newblock React: Synergizing reasoning and acting in language models.
\newblock \emph{arXiv preprint arXiv:2210.03629}, 2022.

\bibitem[Zelikman et~al.(2022)Zelikman, Wu, Mu, and Goodman]{zelikman2022star}
E.~Zelikman, Y.~Wu, J.~Mu, and N.~Goodman.
\newblock Star: Bootstrapping reasoning with reasoning.
\newblock \emph{Advances in Neural Information Processing Systems},
  35:\penalty0 15476--15488, 2022.

\bibitem[Zelikman et~al.(2024)Zelikman, Harik, Shao, Jayasiri, Haber, and
  Goodman]{zelikman2024quiet}
E.~Zelikman, G.~Harik, Y.~Shao, V.~Jayasiri, N.~Haber, and N.~D. Goodman.
\newblock Quiet-star: Language models can teach themselves to think before
  speaking.
\newblock \emph{arXiv preprint arXiv:2403.09629}, 2024.

\bibitem[Zhai et~al.(2025)Zhai, Bai, Lin, Pan, Tong, Zhou, Suhr, Xie, LeCun,
  Ma, et~al.]{zhai2025fine}
S.~Zhai, H.~Bai, Z.~Lin, J.~Pan, P.~Tong, Y.~Zhou, A.~Suhr, S.~Xie, Y.~LeCun,
  Y.~Ma, et~al.
\newblock Fine-tuning large vision-language models as decision-making agents
  via reinforcement learning.
\newblock \emph{Advances in Neural Information Processing Systems},
  37:\penalty0 110935--110971, 2025.

\bibitem[Zhang et~al.(2024)Zhang, Madaan, Gao, Zheng, Mishra, Yang, Tandon, and
  Alon]{zhang2024context}
T.~Zhang, A.~Madaan, L.~Gao, S.~Zheng, S.~Mishra, Y.~Yang, N.~Tandon, and
  U.~Alon.
\newblock In-context principle learning from mistakes.
\newblock \emph{arXiv preprint arXiv:2402.05403}, 2024.

\bibitem[Zhou et~al.(2023)Zhou, Xu, Zhu, Zhou, Lo, Sridhar, Cheng, Ou, Bisk,
  Fried, et~al.]{zhou2023webarena}
S.~Zhou, F.~F. Xu, H.~Zhu, X.~Zhou, R.~Lo, A.~Sridhar, X.~Cheng, T.~Ou,
  Y.~Bisk, D.~Fried, et~al.
\newblock Webarena: A realistic web environment for building autonomous agents.
\newblock \emph{arXiv preprint arXiv:2307.13854}, 2023.

\end{thebibliography}

\appendix
\onecolumn
\resumetocwriting
\renewcommand{\baselinestretch}{0.75}\normalsize
\tableofcontents

\section{Environments \& Datasets}
We conduct experiments on three sets of environments: multi-armed bandits, contextual bandits and tic-tac-toe.
For the SFT experiments reported in Section \ref{sec:ablations}, we generate our own expert datasets. 
In this section, we provide additional details on our environments and datasets.

\subsection{Multi-arm Bandits: BanditBench}\label{appendix:mabs}
MABs \citep{slivkins2019introduction, lattimore2020bandit} are a classic problem setting in RL that isolates the \emph{exploration-exploitation} trade-off.
In contrast, commonly used RL environments \citep{bellemare2013arcade,tassa2018deepmind} often conflate exploration with other RL-specific aspects, such as delayed rewards \citep{arjona2019rudder}.
We rely on the MAB scenarios released in BanditBench \citep{nie2024evolve} and also used by \citep{krishnamurthy2024can}.
MABs come with a number of variable dimensions including the scenario type (textual description of the task), the type of reward distribution (Gaussian, Bernoulli) and its corresponding noise level (low/medium/high), the number of arms (i.e., actions), and the number of interaction steps per episode.
Consequently, MABs are a good testbed for LLM agents.

\begin{figure*}[h]
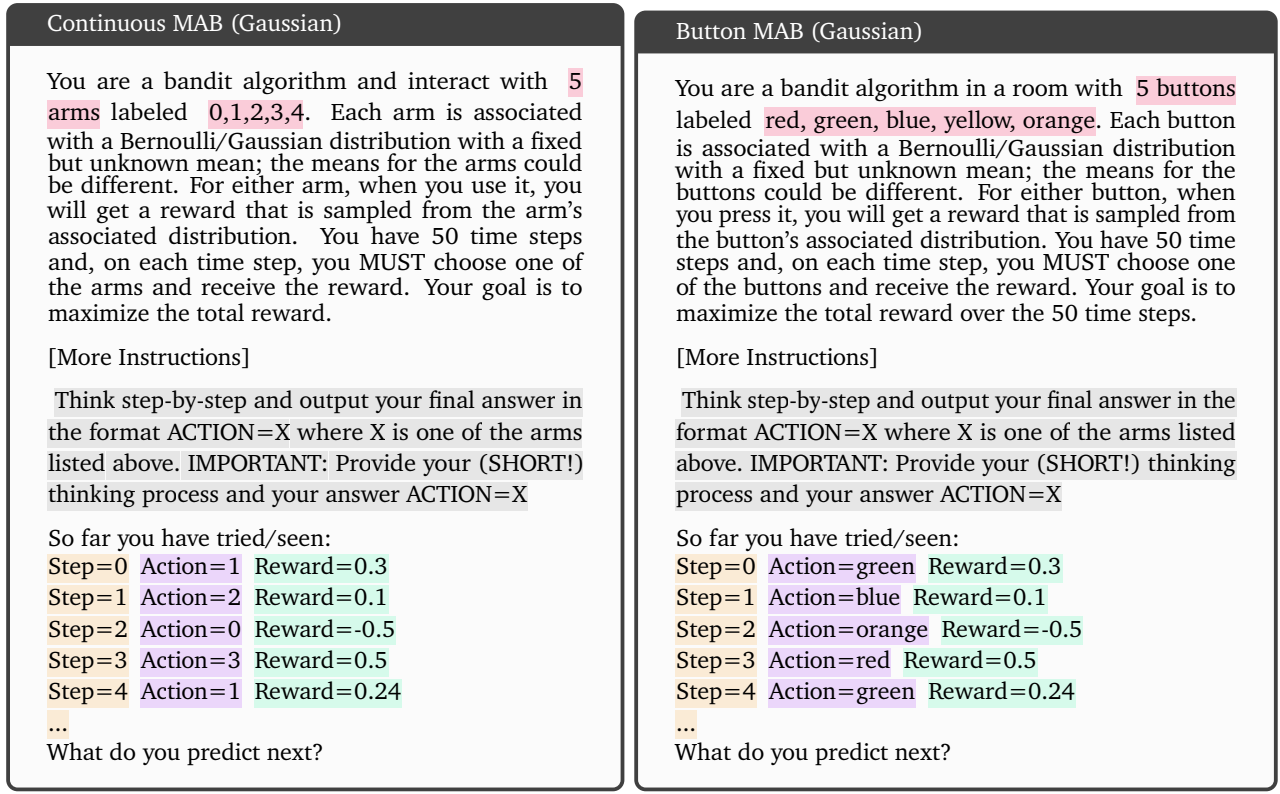

\noindent
\small
    \centering
\begin{minipage}[t]{0.49\textwidth}
  \footnotesize
    \begin{tcolorbox}[colback=gray!3!white, colframe=gray!50!black, title=Continuous MAB (Gaussian)]
        You are a bandit algorithm and interact with \sethlcolor{paleblush}~\hl{5 arms} labeled \sethlcolor{paleblush}~\hl{0,1,2,3,4}. Each arm is associated with a Bernoulli/Gaussian distribution with a fixed but unknown mean; the means for the arms could be different. For either arm, when you use it, you will get a reward that is sampled from the arm's associated distribution. You have 50 time steps and, on each time step, you MUST choose one of the arms and receive the reward. Your goal is to maximize the total reward.\newline
        
        [More Instructions] \newline 
    
        \sethlcolor{lightgray}~\hl{Think step-by-step and output your final answer in the format ACTION=X where X is one of the arms listed above. 
        IMPORTANT: Provide your (SHORT!) thinking process and your answer ACTION=X}\newline
        
        So far you have tried/seen: \newline
        \sethlcolor{antiquewhite}~\hl{Step=0} \sethlcolor{lavender}~\hl{Action=1} \sethlcolor{azure(web)(azuremist)}~\hl{Reward=0.3} \newline
        \sethlcolor{antiquewhite}~\hl{Step=1} \sethlcolor{lavender}~\hl{Action=2} \sethlcolor{azure(web)(azuremist)}~\hl{Reward=0.1} \newline
        \sethlcolor{antiquewhite}~\hl{Step=2} \sethlcolor{lavender}~\hl{Action=0} \sethlcolor{azure(web)(azuremist)}~\hl{Reward=-0.5} \newline
        \sethlcolor{antiquewhite}~\hl{Step=3} \sethlcolor{lavender}~\hl{Action=3} \sethlcolor{azure(web)(azuremist)}~\hl{Reward=0.5} \newline
        \sethlcolor{antiquewhite}~\hl{Step=4} \sethlcolor{lavender}~\hl{Action=1} \sethlcolor{azure(web)(azuremist)}~\hl{Reward=0.24} \newline
        \sethlcolor{antiquewhite}~\hl{...} \newline 
        What do you predict next? 
    \end{tcolorbox}
\end{minipage}%
~
\begin{minipage}[t]{0.51\textwidth}
  \footnotesize
    \begin{tcolorbox}[colback=gray!3!white, colframe=gray!50!black, title=Button MAB (Gaussian)]

        You are a bandit algorithm in a room with \sethlcolor{paleblush}~\hl{5 buttons} labeled \sethlcolor{paleblush}~\hl{red, green, blue, yellow, orange}. Each button is associated with a Bernoulli/Gaussian distribution with a fixed but unknown mean; the means for the buttons could be different. For either button, when you press it, you will get a reward that is sampled from the button's associated distribution. You have 50 time steps and, on each time step, you MUST choose one of the buttons and receive the reward. Your goal is to maximize the total reward over the 50 time steps.\newline
        
        [More Instructions] \newline 
    
        \sethlcolor{lightgray}~\hl{Think step-by-step and output your final answer in the format ACTION=X where X is one of the arms listed above. 
        IMPORTANT: Provide your (SHORT!) thinking process and your answer ACTION=X}\newline
        
        So far you have tried/seen:\newline
        \sethlcolor{antiquewhite}~\hl{Step=0} \sethlcolor{lavender}~\hl{Action=green} \sethlcolor{azure(web)(azuremist)}~\hl{Reward=0.3} \newline
        \sethlcolor{antiquewhite}~\hl{Step=1} \sethlcolor{lavender}~\hl{Action=blue} \sethlcolor{azure(web)(azuremist)}~\hl{Reward=0.1} \newline
        \sethlcolor{antiquewhite}~\hl{Step=2} \sethlcolor{lavender}~\hl{Action=orange} \sethlcolor{azure(web)(azuremist)}~\hl{Reward=-0.5} \newline
        \sethlcolor{antiquewhite}~\hl{Step=3} \sethlcolor{lavender}~\hl{Action=red} \sethlcolor{azure(web)(azuremist)}~\hl{Reward=0.5} \newline
        \sethlcolor{antiquewhite}~\hl{Step=4} \sethlcolor{lavender}~\hl{Action=green} \sethlcolor{azure(web)(azuremist)}~\hl{Reward=0.24} \newline
        \sethlcolor{antiquewhite}~\hl{...} \newline 
        What do you predict next? 
    \end{tcolorbox}
\end{minipage}%
\caption{Illustration of \emph{continuous} and \emph{button} Gaussian multi-armed bandits scenarios from BanditBench \citep{nie2024evolve} using our context representation and \textbf{with CoT} instructions.}
\label{fig:bandits-with-cot-double}
\end{figure*}

\begin{figure}[h]
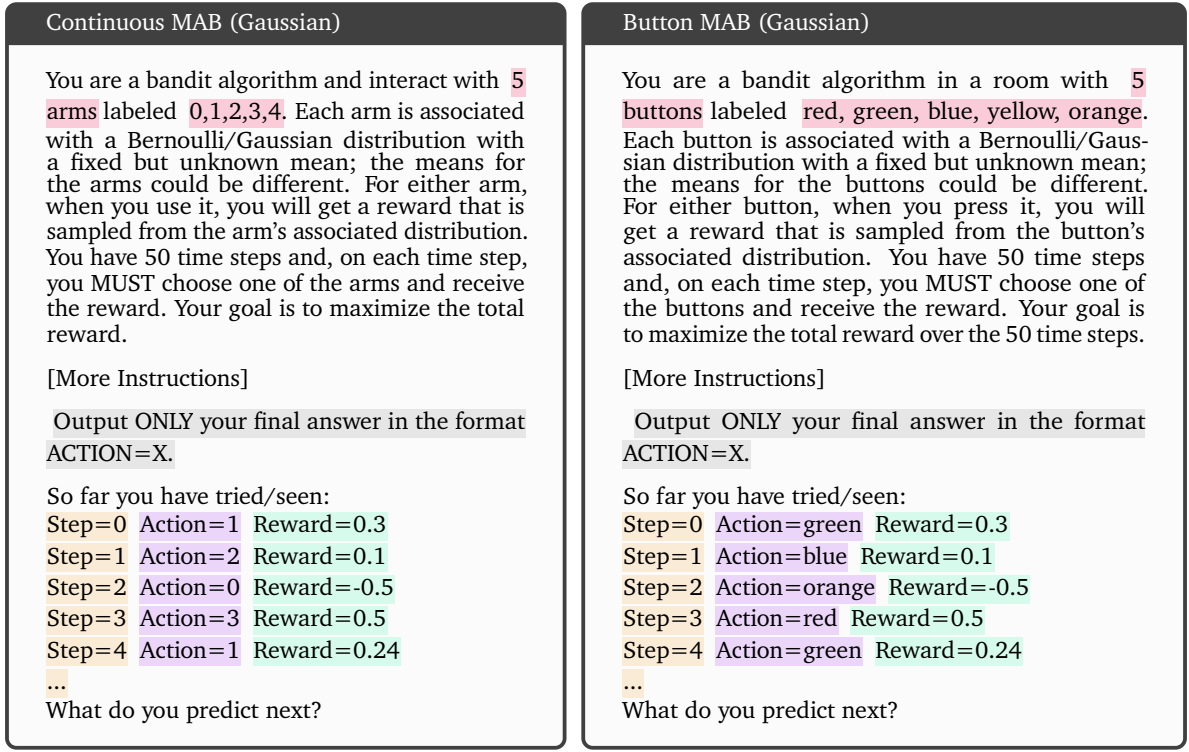

\noindent
\small
    \centering
\begin{minipage}[t]{0.445\textwidth}
  \footnotesize
    \begin{tcolorbox}[colback=gray!3!white, colframe=gray!50!black, title=Continuous MAB (Gaussian)]
        You are a bandit algorithm and interact with \sethlcolor{paleblush}~\hl{5 arms} labeled \sethlcolor{paleblush}~\hl{0,1,2,3,4}. Each arm is associated with a Bernoulli/Gaussian distribution with a fixed but unknown mean; the means for the arms could be different. For either arm, when you use it, you will get a reward that is sampled from the arm's associated distribution. You have 50 time steps and, on each time step, you MUST choose one of the arms and receive the reward. Your goal is to maximize the total reward.\newline
        
        [More Instructions] \newline 
    
        \sethlcolor{lightgray}~\hl{Output ONLY your final answer in the format ACTION=X.}\newline
        
        So far you have tried/seen: \newline
        \sethlcolor{antiquewhite}~\hl{Step=0} \sethlcolor{lavender}~\hl{Action=1} \sethlcolor{azure(web)(azuremist)}~\hl{Reward=0.3} \newline
        \sethlcolor{antiquewhite}~\hl{Step=1} \sethlcolor{lavender}~\hl{Action=2} \sethlcolor{azure(web)(azuremist)}~\hl{Reward=0.1} \newline
        \sethlcolor{antiquewhite}~\hl{Step=2} \sethlcolor{lavender}~\hl{Action=0} \sethlcolor{azure(web)(azuremist)}~\hl{Reward=-0.5} \newline
        \sethlcolor{antiquewhite}~\hl{Step=3} \sethlcolor{lavender}~\hl{Action=3} \sethlcolor{azure(web)(azuremist)}~\hl{Reward=0.5} \newline
        \sethlcolor{antiquewhite}~\hl{Step=4} \sethlcolor{lavender}~\hl{Action=1} \sethlcolor{azure(web)(azuremist)}~\hl{Reward=0.24} \newline
        \sethlcolor{antiquewhite}~\hl{...} \newline 
        What do you predict next? 
    \end{tcolorbox}
\end{minipage}%
~
\begin{minipage}[t]{0.48\textwidth}
  \footnotesize
    \begin{tcolorbox}[colback=gray!3!white, colframe=gray!50!black, title=Button MAB (Gaussian)]

        You are a bandit algorithm in a room with \sethlcolor{paleblush}~\hl{5 buttons} labeled \sethlcolor{paleblush}~\hl{red, green, blue, yellow, orange}. Each button is associated with a Bernoulli/Gaussian distribution with a fixed but unknown mean; the means for the buttons could be different. For either button, when you press it, you will get a reward that is sampled from the button's associated distribution. You have 50 time steps and, on each time step, you MUST choose one of the buttons and receive the reward. Your goal is to maximize the total reward over the 50 time steps.\newline
        
        [More Instructions] \newline 
        
        \sethlcolor{lightgray}~\hl{Output ONLY your final answer in the format ACTION=X.}\newline
        
        So far you have tried/seen:\newline
        \sethlcolor{antiquewhite}~\hl{Step=0} \sethlcolor{lavender}~\hl{Action=green} \sethlcolor{azure(web)(azuremist)}~\hl{Reward=0.3} \newline
        \sethlcolor{antiquewhite}~\hl{Step=1} \sethlcolor{lavender}~\hl{Action=blue} \sethlcolor{azure(web)(azuremist)}~\hl{Reward=0.1} \newline
        \sethlcolor{antiquewhite}~\hl{Step=2} \sethlcolor{lavender}~\hl{Action=orange} \sethlcolor{azure(web)(azuremist)}~\hl{Reward=-0.5} \newline
        \sethlcolor{antiquewhite}~\hl{Step=3} \sethlcolor{lavender}~\hl{Action=red} \sethlcolor{azure(web)(azuremist)}~\hl{Reward=0.5} \newline
        \sethlcolor{antiquewhite}~\hl{Step=4} \sethlcolor{lavender}~\hl{Action=green} \sethlcolor{azure(web)(azuremist)}~\hl{Reward=0.24} \newline
        \sethlcolor{antiquewhite}~\hl{...} \newline 
        What do you predict next? 
    \end{tcolorbox}
\end{minipage}%
\caption{Illustration of \emph{continuous} and \emph{button} Gaussian multi-armed bandits scenarios from BanditBench \citep{nie2024evolve} using our context representation \textbf{without CoT} instructions.}
\label{fig:banditbench-no-cot}
\end{figure}

\noindent We focus on the \emph{continuous} and \emph{button} variants released by \citet{nie2024evolve}.
We report results for MAB with $k \in \{5,10,20\}$ arms ($|\mathcal{A}| = k$) for three levels of stochasticity (low/medium/high).
In our experiments, for every arm the corresponding reward is sampled from a Gaussian distribution $r \sim \mathcal{N}(\mu, \sigma)$ where $\mu \sim \mathcal{U}(0, 1)$ and is a fixed scalar $\sigma \in \{0.1, 1, 3\}$ for the three levels of stochasticity, respectively.
For all MAB settings, we limit the horizon $T$ to 50 interaction steps.
Limiting the horizon is necessary to handle the increasing lengths and consequently RAM requirements for fine-tuning.
While we consider 50 interaction steps sufficient for 5 and 10 arms, it is insufficient for the 20 arms scenario. 
However, note that the general trends are well observable for the 20 arms scenario.
In Figure \ref{fig:bandits-with-cot-double}, we show the continuous and button Gaussian MABs with CoT instructions for the agent.
Similarly, in Figure \ref{fig:banditbench-no-cot} we show the same instances without CoT instructions. 

\noindent\subsubsection{Baselines.} 
We compare against two commonly used baselines for MABs: Upper-confidence Bound (UCB) \citep{auer2002using} and a random agent that selects actions uniformly at random (see Appendix \ref{appendix:exp-details} for details).
UCB is considered optimal and represents the upper-bound for agent performance, whereas the random baseline represents the lower bound. 
We provide implementations details for all baselines in Appendix \ref{appendix:exp-details}.

\noindent\subsubsection{SFT Datasets.}
In our main experiments, we focused on self-generated CoT rationales and action predictions produced by our fine-tuned agents, which do not require a pre-collected dataset.
In contrast, for our SFT experiments reported in Section \ref{sec:ablations}, we generated UCB expert datasets. 
In particular, we construct two dataset instances: a \emph{behavior cloning} dataset that only contains expert actions and a \emph{thought cloning} (TC) dataset that incorporates expert actions alongside a thought process for coming up with the respective action (i.e., w/ CoT).
For every arm and stochasticity level combination, we construct 32K rollouts from different MABs, which amounts to 1.6M transitions (state, action, reward) per dataset. 

\begin{figure}[h]
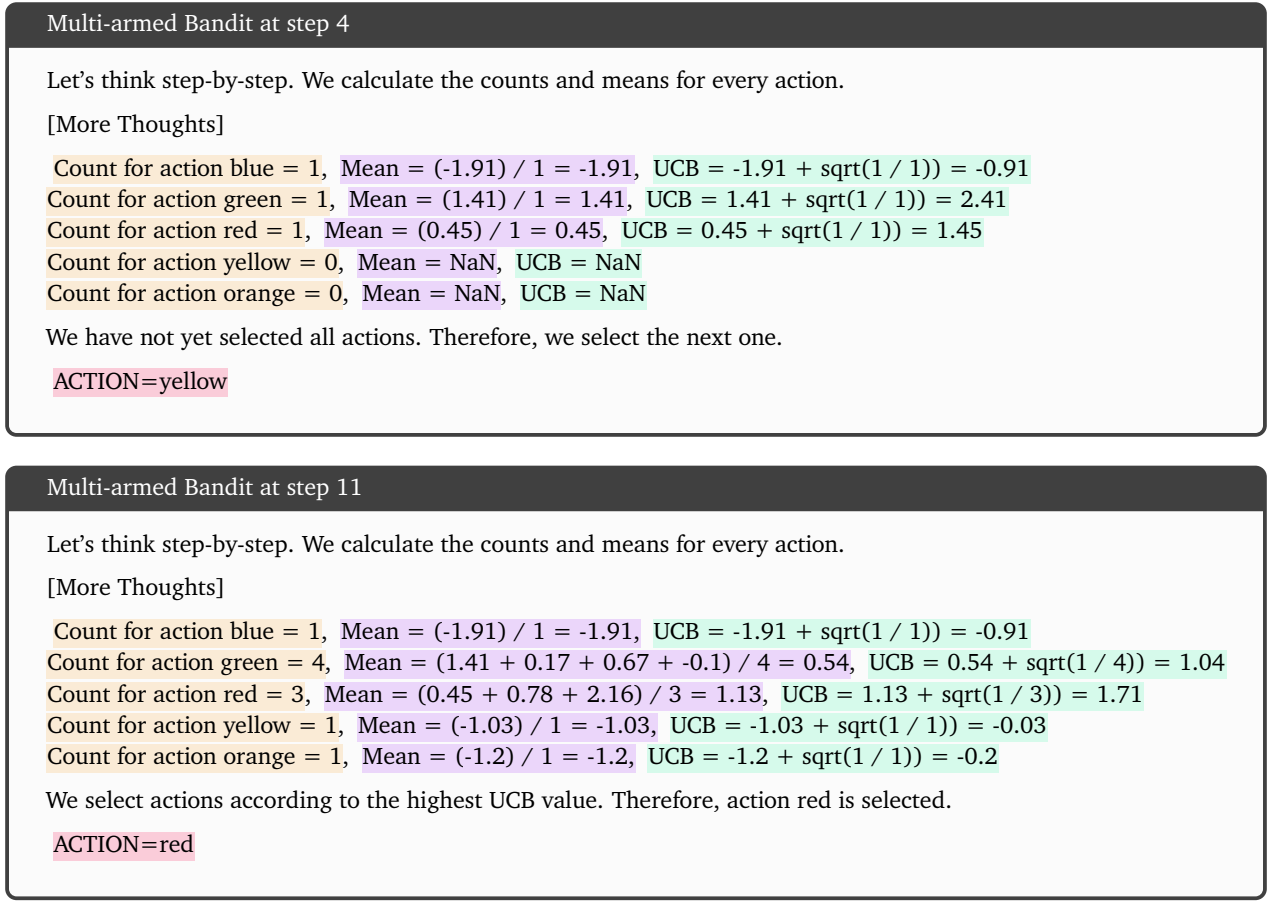

\noindent
\small
\centering
  \footnotesize
    \begin{tcolorbox}[colback=gray!3!white, colframe=gray!50!black, title=Multi-armed Bandit at step 4]
        Let's think step-by-step. We calculate the counts and means for every action. \newline

        [More Thoughts] \newline 
        
        \sethlcolor{antiquewhite}~\hl{Count for action blue = 1}, \sethlcolor{lavender}~\hl{Mean = (-1.91) / 1 = -1.91},  \sethlcolor{azure(web)(azuremist)}~\hl{UCB = -1.91 + sqrt(1 / 1)) = -0.91}  \newline 
        \sethlcolor{antiquewhite}~\hl{Count for action green = 1}, \sethlcolor{lavender}~\hl{Mean = (1.41) / 1 = 1.41},  \sethlcolor{azure(web)(azuremist)}~\hl{UCB = 1.41 + sqrt(1 / 1)) = 2.41}  \newline 
        \sethlcolor{antiquewhite}~\hl{Count for action red = 1}, \sethlcolor{lavender}~\hl{Mean = (0.45) / 1 = 0.45},  \sethlcolor{azure(web)(azuremist)}~\hl{UCB = 0.45 + sqrt(1 / 1)) = 1.45}  \newline 
        \sethlcolor{antiquewhite}~\hl{Count for action yellow = 0}, \sethlcolor{lavender}~\hl{Mean = NaN},  \sethlcolor{azure(web)(azuremist)}~\hl{UCB = NaN}  \newline 
        \sethlcolor{antiquewhite}~\hl{Count for action orange = 0}, \sethlcolor{lavender}~\hl{Mean = NaN},  \sethlcolor{azure(web)(azuremist)}~\hl{UCB = NaN}  \newline 

        We have not yet selected all actions. Therefore, we select the next one.\newline
        
        \sethlcolor{paleblush}~\hl{ACTION=yellow} \newline
    \end{tcolorbox}
~

\begin{minipage}[t]{1\textwidth}
  \footnotesize
    \begin{tcolorbox}[colback=gray!3!white, colframe=gray!50!black, title=Multi-armed Bandit at step 11 ]

        Let's think step-by-step. We calculate the counts and means for every action. \newline

        [More Thoughts] \newline 
        
        \sethlcolor{antiquewhite}~\hl{Count for action blue = 1}, \sethlcolor{lavender}~\hl{Mean = (-1.91) / 1 = -1.91,}   \sethlcolor{azure(web)(azuremist)}~\hl{UCB = -1.91 + sqrt(1 / 1)) = -0.91} \newline
        \sethlcolor{antiquewhite}~\hl{Count for action green = 4}, \sethlcolor{lavender}~\hl{Mean = (1.41 + 0.17 + 0.67 + -0.1) / 4 = 0.54},   \sethlcolor{azure(web)(azuremist)}~\hl{UCB = 0.54 + sqrt(1 / 4)) = 1.04} \newline
        \sethlcolor{antiquewhite}~\hl{Count for action red = 3}, \sethlcolor{lavender}~\hl{Mean = (0.45 + 0.78 + 2.16) / 3 = 1.13},   \sethlcolor{azure(web)(azuremist)}~\hl{UCB = 1.13 + sqrt(1 / 3)) = 1.71} \newline
        \sethlcolor{antiquewhite}~\hl{Count for action yellow = 1}, \sethlcolor{lavender}~\hl{Mean = (-1.03) / 1 = -1.03,}   \sethlcolor{azure(web)(azuremist)}~\hl{UCB = -1.03 + sqrt(1 / 1)) = -0.03} \newline
        \sethlcolor{antiquewhite}~\hl{Count for action orange = 1}, \sethlcolor{lavender}~\hl{Mean = (-1.2) / 1 = -1.2,}   \sethlcolor{azure(web)(azuremist)}~\hl{UCB = -1.2 + sqrt(1 / 1)) = -0.2} \newline
        
        We select actions according to the highest UCB value. Therefore, action red is selected.\newline
        
        \sethlcolor{paleblush}~\hl{ACTION=red} \newline

    \end{tcolorbox}
\end{minipage}%
\caption{Illustration of UCB rationales contained in our \textbf{SFT expert datasets} at two timesteps (4 and 11) in the same trajectory. Both examples show the Thought Cloning dataset instance containing both the produced CoT rationale along with the predicted action. The Behavior Cloning instances contain only the final action prediction (in red).}
\label{fig:sft-data-instances}
\end{figure}

\noindent To provide the thought process in the TC datasets, we reconstruct the computations of the UCB values conducted by the UCB expert in textual format.
The thought process ends with a final conclusion why a particular action was selected (i.e., highest UCB value or exploratory action).
Consequently, the action $z_t$ at step $t$ contains the thought process $z_{CoT}$ and the action to execute $a_t$ (see Section \ref{sec:method}).
We illustrate the actions contained in the dataset for a trajectory at steps 4 and 11 in Figure \ref{fig:sft-data-instances}.
The BC datasets do not contain the thought process.
Instead, they only contain the final predictions made by the model, for example \texttt{ACTION=yellow} as shown in Figure \ref{fig:sft-data-instances}.

\subsection{Contextual Bandits} \label{appendix:cbs}
\begin{figure*}[h]
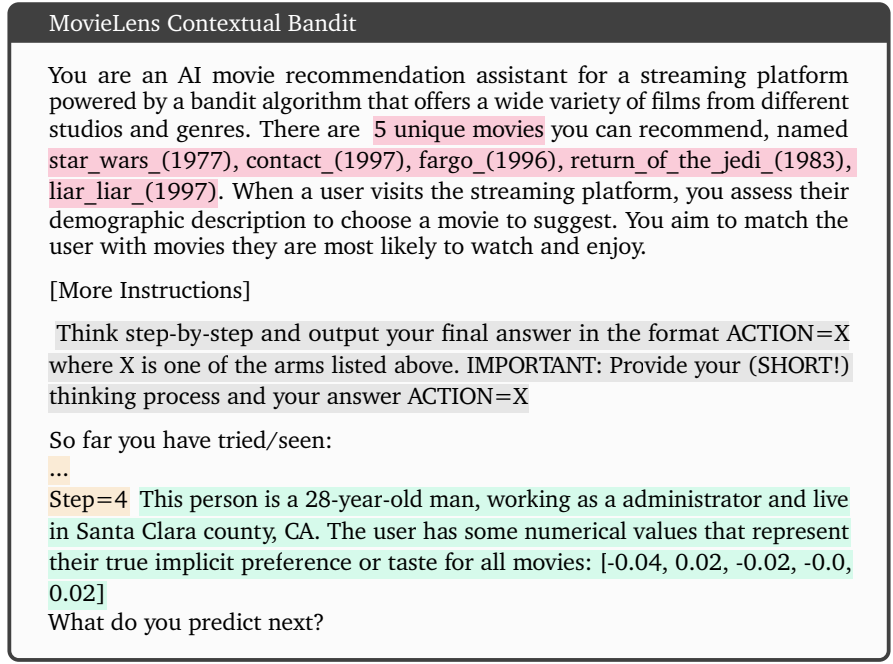

\noindent
\small
\centering
\begin{minipage}[t]{0.7\textwidth}
  \footnotesize
    \begin{tcolorbox}[colback=gray!3!white, colframe=gray!50!black, title=MovieLens Contextual Bandit]
        
        You are an AI movie recommendation assistant for a streaming platform powered by a bandit algorithm that offers a wide variety of films from different studios and genres. There are \sethlcolor{paleblush}~\hl{5 unique movies} you can recommend, named \sethlcolor{paleblush}~\hl{star\_wars\_(1977), contact\_(1997), fargo\_(1996), return\_of\_the\_jedi\_(1983), liar\_liar\_(1997)}.
        When a user visits the streaming platform, you assess their demographic description to choose a movie to suggest.
        You aim to match the user with movies they are most likely to watch and enjoy.\newline
        
        [More Instructions] \newline 
    
        \sethlcolor{lightgray}~\hl{Think step-by-step and output your final answer in the format ACTION=X where X is one of the arms listed above. 
        IMPORTANT: Provide your (SHORT!) thinking process and your answer ACTION=X}\newline
        
        So far you have tried/seen:\newline
        \sethlcolor{antiquewhite}~\hl{...} \newline 
        \sethlcolor{antiquewhite}~\hl{Step=4} \sethlcolor{azure(web)(azuremist)}~\hl{This person is a 28-year-old man, working as a administrator and live in Santa Clara county, CA. The user has some numerical values that represent their true implicit preference or taste for all movies: [-0.04, 0.02, -0.02, -0.0, 0.02]} \newline
        What do you predict next? 
    \end{tcolorbox}
\end{minipage}%
\caption{Illustration of \textbf{contextual MovieLens scenario} from BanditBench \citep{nie2024evolve} using our context representation and instructions.}
\label{fig:contextual-bandits-with-cot}
\end{figure*}

MABs do not emit states. 
In contrast, CBs emit state representations at every interaction step, making them contextual.
Consequently, CBs are interesting to test abilities of LLMs to make use of the given context when predicting the next action.
For our CB experiments, we leverage the MovieLens environment released by \citet{nie2024evolve}, a semisynthetic bandit task based on the MovieLens dataset \citep{harper2015movielens}.
In this setting, the agent operates as a movie recommendation engine given a contextual description of a user (10K users in total) and a list of $K$ possible movies.
The context representation provides a textual description of the user to recommend the movie to. 
This description includes the user's gender, age, profession, location, and a numeric description of the user's preferences for each of the possible movies. 
As for MABs, we report results for $K \in \{5, 10, 20\}$, limit the horizon to 50 interaction steps.
In Figure \ref{fig:contextual-bandits-with-cot}, we provide an example for a MovieLens CB with 5 actions with our context representation and CoT instructions.

\noindent\textbf{Baselines.}
Similar to MABs, we compare against LinUCB \citep{chu2011contextual} and an agent selecting actions uniformly at random.
We provide implementation details on our baselines in Appendix \ref{appendix:exp-details}.

\subsection{Tic-tac-toe} \label{appendix:ttt}
Finally, we use the text-based Tic-tac-toe environment released by \citet{ruoss2024lmact} (see Figure \ref{fig:tictactoe-input} for an example).
Unlike MABs and CBs, Tic-tac-toe is a stateful environment with proper state transitions (i.e., action predicted at step $t$ affects the state observed at step $t+1$).
The agent receives scalar rewards of 1, 0, and -1 for winning, drawing, and loosing against its opponent, respectively.
Episodes last until either of the players wins, draws or looses.
To enable easy extraction of actions from the generated rationales, we represent the action space as a discrete set of 9 actions, corresponding to the grid positions on the $3\times3$ grid used in Tic-tac-toe ($\mid \mathcal{A} \mid = 9$).
However, only at the start of an episode, all 9 actions are valid. 
Subsequently, only a subset is valid, because of taken board positions. 
We (optionally) provide the set of valid actions at a particular step in textual form in the context given to the agent.
\citet{ruoss2024lmact} demonstrated that frontier models struggle to achieve strong performance in this environment and barely beat a random opponent.
Consequently, we deem it a good target to investigate the efficacy of RLFT.

\noindent \textbf{Baselines.}
Following \citet{ruoss2024lmact}, we compare against a random agent by default. 
In addition, we also compare against (MCTS) \citep{coulom2006efficient},  and a noisy variant of MCTS that selects an action randomly with 50\% chance and according to MCTS otherwise.

\begin{figure*}[h]
\noindent
\small
    \centering
  \footnotesize
    \begin{tcolorbox}[colback=gray!3!white, colframe=gray!50!black, title=Tic-tac-toe]
    You are an agent playing tic-tac-toe. 
    You observe a board with 9 entries that looks like this:
    \newline
    \sethlcolor{paleblush}~\hl{000}\newline 
    \sethlcolor{paleblush}~\hl{100}\newline
    \sethlcolor{paleblush}~\hl{002}
    
    1 indicates that player 1 has placed a stone in that square.
    2 indicates that player 2 has placed a stone in that square.
    0 indicates that no stone has been placed in that square. You play as 1.
    
    There are 9 possible actions: \sethlcolor{paleblush}~\hl{0, 1, 2, 3, 4, 5, 6, 7, 8}. The actions correspond to the following board locations
    \newline
    \sethlcolor{paleblush}~\hl{012}\newline
    \sethlcolor{paleblush}~\hl{345}\newline
    \sethlcolor{paleblush}~\hl{678}\newline

    [More Instructions] \newline 
    
        \sethlcolor{lightgray}~\hl{Think step-by-step and output your final answer in the format ACTION=X where X is one of the arms listed above. 
        IMPORTANT: Provide your (SHORT!) thinking process and your answer ACTION=X}\newline
        
        So far you have tried/seen: \newline
        \sethlcolor{antiquewhite}~\hl{Step=0} \sethlcolor{azure(web)(azuremist)}~\hl{State=000000000}\sethlcolor{lavender}~\hl{Action=0} \sethlcolor{azure(web)(azuremist)}~\hl{Reward=0} \newline
        \sethlcolor{antiquewhite}~\hl{Step=1} \sethlcolor{azure(web)(azuremist)}~\hl{State=102000000}\sethlcolor{lavender}~\hl{Action=4} \sethlcolor{azure(web)(azuremist)}~\hl{Reward=0} \newline
        \sethlcolor{antiquewhite}~\hl{Step=2} \sethlcolor{azure(web)(azuremist)}~\hl{State=102010002}\sethlcolor{lavender}~\hl{Action=5} \sethlcolor{azure(web)(azuremist)}~\hl{Reward=0} \newline
        What do you predict next? 
    \end{tcolorbox}
\caption{Illustration of the text-based \textbf{ Tic-tac-toe} environment.}
\label{fig:tictactoe-input}
\end{figure*}

\section{Experimental \& Implementation Details} \label{appendix:exp-details}
\subsection{Training \& Evaluation}
In our experiments, we fine-tune Gemma2 models in three model sizes (2B/9B/27B).
For all experiments, we use the instruction-tuned versions of Gemma2 and leverage the respective instruction pre-and-postfixes.  
For bandits, we fine-tune all models for a total of 30K updates and evaluate after every 10K steps.
with an accumulated batch size of 128.
Similarly, we fine-tune for 12K updates and evaluate every 4K updates on Tic-tac-toe.
We report the mean and 95\% confidence intervals over three seeds, as suggested by \citet{agarwal2021deep}.

\noindent \textbf{General.} We train all agents with an accumulated batch size of 128. 
We use a learning rate of $1e^{-4}$, 100 linear warm-up steps followed by a cosine decay to $1e^{-6}$. 
To allow for memory-efficient fine-tuning of 2B and 9B models, we train using the AdaFactor optimizer \citep{shazeer2018adafactor}.
We experiment with LoRA \citep{hu2022lora} for fine-tuning the 9B and 27B models but found it insufficient for improving the agent's decision-making abilities in our setting.
However, LoRA considerably reduces the amount of memory required for RLFT and has been shows to work well for supervised fine-tuning of decision-making agents \citep{schmied2023learning}. 
Therefore, we deem it a promising candidate for RLFT in decision-making scenarios. 
Furthermore, we employ gradient clipping of 1.0. 
We list all hyperparameters in Table \ref{tab:hparams}.
\begin{table}[h]
    \centering
    \resizebox{1\textwidth}{!}{
    \begin{tabular}{llr}
    \toprule
    \textbf{Name} & \textbf{Value} & \textbf{Description} \\
    \midrule
    \multicolumn{3}{l}{\textbf{Training}} \\
    \midrule
    \texttt{training\_steps} & $30$K or $12$K & Number of training steps. \\
    \texttt{eval\_freq} & 10K or 4K & Evaluation frequency (in updates).\\
    \texttt{batch\_size} & $128$ & Accumulated batch size. \\
    \texttt{lr\_scheduler} & Linear + cosine &  Learning rate scheduler \\
    \texttt{warmup steps} & 100 &  Warmup steps. \\
    \texttt{lr} & $1e^4$ to $1e^6$ & Maximum learning rate. \\
    \texttt{optimizer} & AdaFactor & Optimizer. \\
    \midrule
    \multicolumn{3}{l}{\textbf{Sequence Length \& Generation Budget}} \\
    \midrule
    \texttt{context\_length} & 1792 & Input context length. \\
    \texttt{num\_tokens} & 256 & Generation budget. \\
    \midrule
    \multicolumn{3}{l}{\textbf{RLFT}} \\
    \midrule
    \texttt{rollout\_steps} & 800 or 2048 & Rollout steps in-between updates. \\
    \texttt{update\_epochs} & 1 or 2 & Update epochs over rollout-buffer. \\
    \texttt{reward\_penalty} & -5 & Reward penalty for invalid actions. \\
    \texttt{loss} & PPO clipping objective + KL constraint & Objective function. \\
    \texttt{baseline} & MC-baseline or state-value head & Baseline. \\
    \texttt{envs} & 16 & Number of parallel envs. \\
    \texttt{$\epsilon$} & 0.2 & Clipping value. \\
    \texttt{$\beta$} & 0.05 & KL coefficient. \\
    \texttt{reward\_norm} & True & Whether reward normalization is used. \\
    \texttt{train\_temp} & 1.0 & Sampling temp during rollouts. \\
    \texttt{eval\_temp} & 0.0 & Sampling temp during evaluation. \\
    \texttt{top\_p} & 1.0 & Sampling top-p. \\
    \midrule
    \multicolumn{3}{l}{\textbf{Hardware}} \\
    \midrule
    \texttt{accelerator} & 8 $\times$ H100 & Hardware accelerator.\\
    \bottomrule
    \end{tabular}
    }
    \caption{Default \textbf{hyperparameters} used in our experiments.}
    \label{tab:hparams}
\end{table}

\noindent\textbf{Context Lengths \& Generation Budget.}
For all model sizes and tasks, we use a context length of 1792 for the input context.
By default, we set the generation budget to 256 tokens, except for the knowing-doing gap analyses reported in Section \ref{sec:failure-modes}, which require a larger budget of 2048 tokens.
Consequently, the effective sequence length for fine-tuning is 2048.

\noindent\textbf{Hardware Setup.}
We train all models on a server equipped with 8 $\times$ H100 GPUs.

\subsection{RLFT}
For our RLFT experiments on bandits, we employ the context representation, action factorization, reward shaping terms, and training objectives described in Section \ref{sec:method}.
To extract the target action $a_t$ from $z_t$, we make use of a stack of regex expressions against the target pattern (i.e., \texttt{ACTION=X}) and consider the last match in the generated tokens as $a_t$. 
In addition to being fairly robust, we found that this approach allows for more flexibility during the RLFT process and led to better outcomes than a more structured approach. 
Furthermore, across model sizes, we found it essential to introduce a reward shaping term to penalize rationales that contain no valid actions. 
By default, we use a reward penalty of -5 for invalid actions.
Empirically, we found that this reward shaping term is sufficient for the models to produce valid actions early on in the training.  

\noindent We fine-tune using the clipping objective introduced by \citet{schulman2017proximal} with an additional KL constraint to the reference policy $\pi_{ref}$.
We set $\beta=0.05$ and $\epsilon=0.2$ for all experiments. 
We make use of the approximated (per-token) KL divergence instead of computing the full KL. 
While we found that computing the full KL slightly improves performance, it slows down training considerably. 
In contrast to \citet{ahmadian2024back} and \cite{ramesh2024group}, we do not rely on producing multiple rollouts, because it is impractical for the mutli-step nature of decision-making tasks.
While generating multiple actions at a particular timestep is possible for simulated environments, it requires environment resets.
Therefore, we rely on standard MC-baselines to estimate $A_{adv}$. 

\noindent For bandit experiments, we maintain a pool of 512 stochastic MABs.
For every rollout, we let the agent interact with a subset of 16 bandits for a single episode (50 timesteps). 
Consequently, every rollout contains 800 transitions.
Similarly, for Tic-tac-toe, we maintain 16 parallel environments and collect 2048 rollout steps. 
We conduct 1 and 2 update epochs over the rollout buffer for bandits and Tic-tac-toe, respectively.

\subsection{SFT}
For our SFT experiments on MABs, we fine-tune on either on the expert action or expert rationales produced by UCB. 
We employ standard SFT training using a cross-entropy objective on the target tokens. 

\subsection{Exploration Mechanisms}\label{appendix:exp-mechanisms}
In Section \ref{fig:exp-mechanisms}, we compare a variety of classic exploration mechanisms and LLM-specific approaches and study their effects on agent performance on Gaussian MABs with 10 arms. 
Here, we provide a description for each mechanism. 

\noindent\textbf{Try-all.}
The try-all strategy is inspired by UCB, which incorporates an initial phase for trying all untried actions.
This is because the UCB values for all untried actions are $\infty$. 
Therefore, we incorporate the same exploration phase when performing ICL and RLFT at the beginning of every episode. 
To enable fine-tuning on exploration actions, we provide an action rationale template to the model (e.g., \texttt{Action X has not been tried yet, let's explore it. ACTION=X}).
While simple, we find that this try-all strategy is effective for lowering regret across all model sizes (see Figure \ref{fig:exp-mechanisms}).
This suggests that the model is able to select appropriate actions if given sufficient information, but struggles to explore. 

\noindent\textbf{$\epsilon$-greedy.}
$\epsilon$-greedy is classic exploration mechanism and commonly used in RL algorithms \citep{mnih2015human,hessel2018rainbow}.
For our experiments, we use $\epsilon=0.1$ both during training and evaluation.
We explored other values for $\epsilon$ but did not observe performance improvements.
As for the try-all strategy, we provide an action rationale template to enable fine-tuning on exploration actions.

\noindent\textbf{Context Randomization.}
Context Randomization is an LLM-specific mechanism designed to introduce randomness in the action predictions by modifying the context representation.
At every interaction step, we construct a mapping from the original action labels to a shuffled list of the same action labels. 
Subsequently, we remap action in the context history according to the constructed mapping.
Finally, the predicted action is mapped back to the original action label space and executed environment. 
Besides introducing randomness, context randomization acts as a control mechanism to ensure that the observed biases do not only stem from biases towards particular action-tokens (e.g., blue occurs more often than magenta in the pre-training dataset).

\noindent\textbf{Context Summary.}
Similar to \citet{krishnamurthy2024can} and \citet{nie2024evolve}, we evaluate the effects of providing a context summary to the agent.
After the context history, we provide the model with a summary of that history that contains the number of times every action has been selected so far, along with their respective mean rewards. 

\noindent\textbf{Self-Correction.}
Inspired by \citet{kumar2024training} and \citet{wang2025critique}, we employ self-correction to the model's predicted actions.
First, we let the model generate its is initial rationale and corresponding action prediction.
Then we append the generated rationale along with a self-correction message (similar to \citet{kumar2024training}) to the input context, and repeat the action generation.
Finally, we extract the action from the final response and execute it in the environment. 
For RLFT, we only fine-tune on the final response, but retain the initial response along with the self-correction message in the context.

\noindent\textbf{Self-Consistency.}
Instead of generating a single answer, self-consistency \citep{wang2022self} relies on generating multiple responses.
Subsequently, self-consistency employs a majority voting mechanism to determine the final response.
For our experiments in Figure \ref{fig:exp-mechanisms}, we report results for self-consistency with 16 generated responses. 
Instead of majority voting, we experimented with sampling from the respective response distribution or random mechanisms.

\noindent\textbf{Exploration Bonus.}
Finally, we evaluate a reward shaping mechanism in the form of an exploration bonus. 
In particular, we give an exploration bonus of +1 if the agents selects an action not yet tried within the respective episode. 
While simple, we find that the exploration bonus effectively narrows the gap to the UCB expert.
This highlights the importance of reward shaping for fine-tuning LLMs in decision-making scenarios.

\begin{figure*}[h]
\noindent
\small
    \centering
\subfigure[Example: History]{
\begin{minipage}[t]{0.28\textwidth}
  \footnotesize
    \begin{tcolorbox}[colback=gray!3!white, colframe=gray!50!black, title=Button MAB,]
        [Instructions]\newline
        
        \noindent\sethlcolor{antiquewhite}~\hl{S=0} \sethlcolor{lavender}~\hl{A=blue} \sethlcolor{azure(web)(azuremist)}~\hl{R=1} \newline
        \sethlcolor{antiquewhite}~\hl{S=1} \sethlcolor{lavender}~\hl{A=blue} \sethlcolor{azure(web)(azuremist)}~\hl{R=1} \newline
        \sethlcolor{antiquewhite}~\hl{S=2} \sethlcolor{lavender}~\hl{A=blue} \sethlcolor{azure(web)(azuremist)}~\hl{R=1} \newline
        \sethlcolor{antiquewhite}~\hl{S=3} \sethlcolor{lavender}~\hl{A=green} \sethlcolor{azure(web)(azuremist)}~\hl{R=0} \newline
        \sethlcolor{antiquewhite}~\hl{S=4} \sethlcolor{lavender}~\hl{A=red} \sethlcolor{azure(web)(azuremist)}~\hl{R=2} \newline
        What do you predict next?\newline
    \end{tcolorbox}
\end{minipage}%
}
~
\subfigure[Example: Probability Mass]{
    \includegraphics[width=0.28\linewidth]{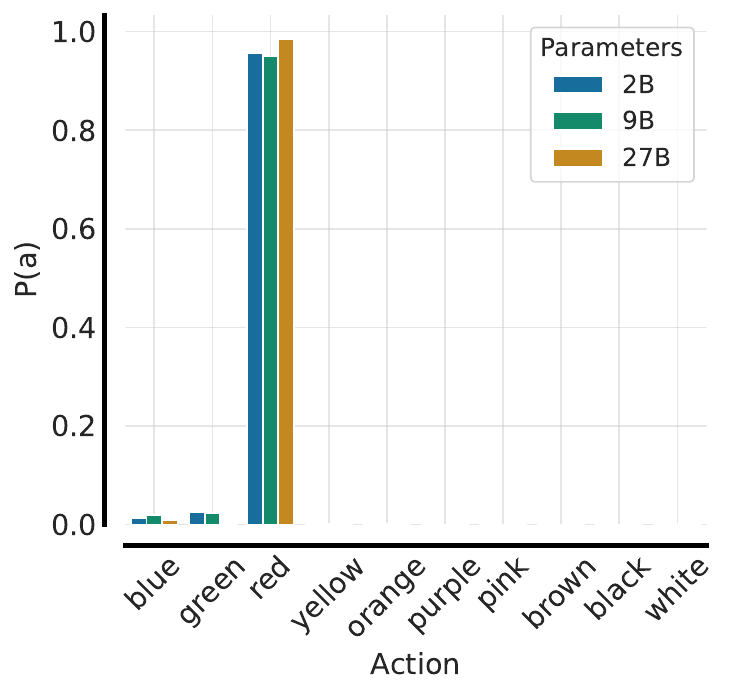} 
}
\caption{Illustration of \textbf{action probabilities} leading to greediness behavior. Models exhibit overly high action probabilities in the presence of rewards, potentially resulting in repeatedly selecting sub-optimal actions}
\label{fig:failure-greedy-pmass}
\end{figure*}

\section{Additional Results} \label{appendix:results}

\subsection{Failure Modes} \label{appendix:failures-modes}

\subsubsection{Greediness} \label{appendix:failures-modes-greediness}
Greediness is characterized by the LLM overly favoring the best performing action among a small set of actions seen so far. 
We define action coverage $C_t$ at step $t$ as the fraction of available actions that have been selected at least once, $C_t=\frac{\{ a \in \mathcal{A} :\, N_t(a) > 0 \}}{\mid \mathcal{A} \mid}$ with $N_t(a)$ representing the number of times action $a$ has been selected until $t$.

\noindent\textbf{Action probabilities.} The suboptimal action coverage reported in Section \ref{sec:failure-modes} is caused by the model overly favoring high-reward actions (i.e., overly high action probabilities). 
In Figure \ref{appendix:failures-modes-greediness}, we provide an illustration of the action probabilities for a given input history. 
Across model sizes, Gemma2 exhibits overly high action probabilities in the presence of reward, which results in repeatedly selecting a potentially suboptimal action.
\begin{figure*}[h]
\noindent
\small
\centering
\subfigure[Action Coverage: 10 arms]{
    \includegraphics[width=0.35\linewidth]{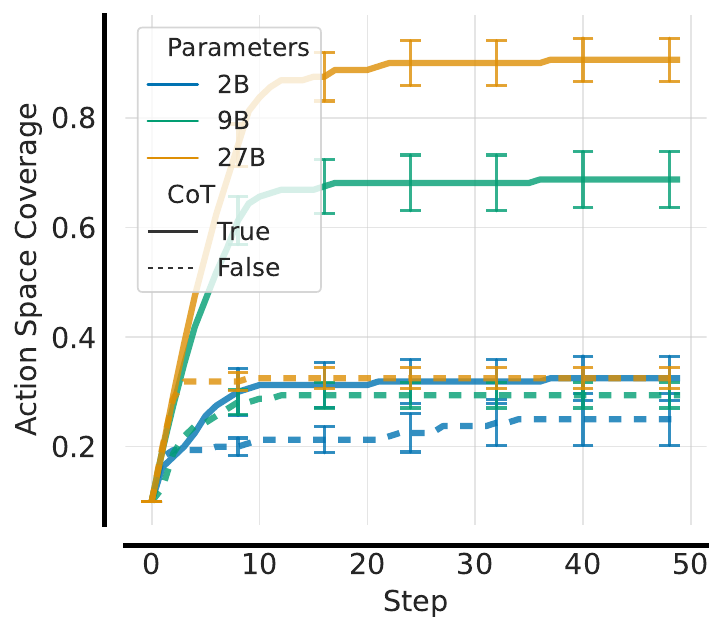}
}
~
\subfigure[Action Coverage: 20 arms]{
    \includegraphics[width=0.35\linewidth]{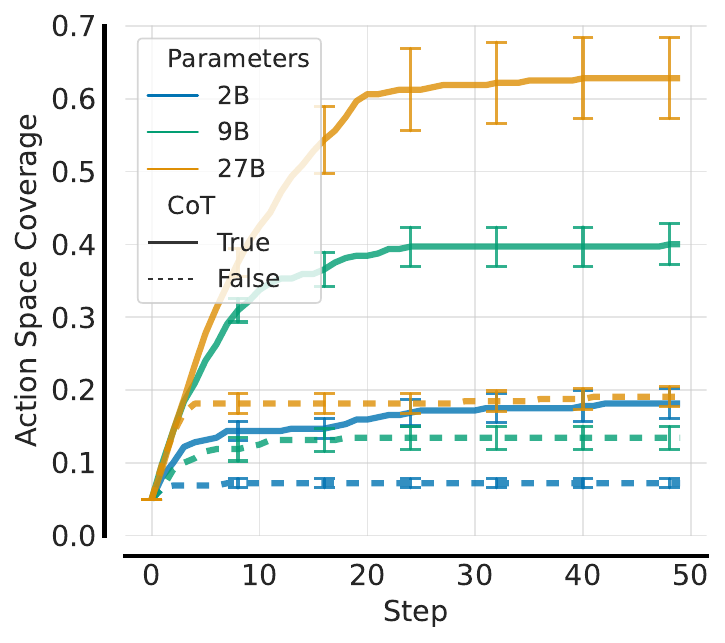}
}
\caption{Illustration of greediness for the \textbf{numbers} scenario.}
\label{fig:failure-greedy-numbers}
\end{figure*}

\noindent\textbf{Greediness on Continuous MABs.}
We repeat the analyses conducted in Section \ref{sec:failure-modes} using \emph{numbers} instead of \emph{buttons} as the possible actions. 
Indeed, we find that the same trends hold. 
Without CoT the performance remains low. 
For Gemma2 27B, we observe an increase in the action coverage to almost 90\% for the 10 arms scenario, and to 60\% for the 20 arms scenario.  

\begin{figure}[h]
\noindent
\small
\centering
\includegraphics[width=0.45\linewidth]{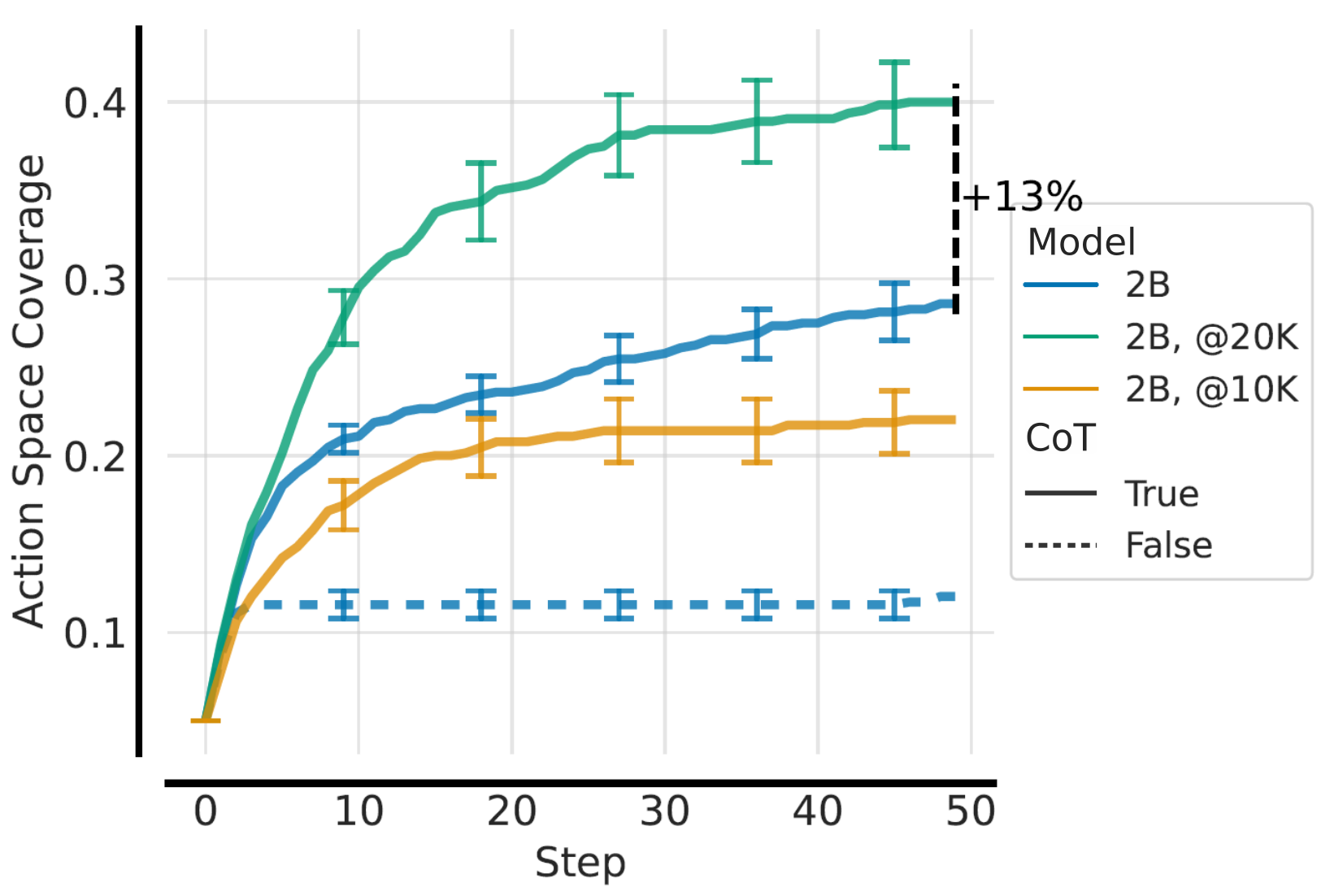}
\caption{Effect of RLFT on \textbf{greediness} for 2B (20 arms, medium noise).}
\label{fig:failure-greedy-rlft-20arms}
\end{figure}

\noindent\textbf{Post RLFT.}
In line with Figure \ref{fig:failure-greedy-rlft}, we present the post RLFT action coverage on the 20 arms scenario in Figure  \ref{fig:failure-greedy-rlft-20arms}.
Similar to the effects on the 10 arms scenario, we observe that RLFT improves the action coverage by 13\%. 

\subsubsection{Frequency Bias}\label{appendix:failures-modes-freq}
Frequency bias is characterized by repeatedly selecting the most frequently occurring actions in the context, even when the dominant action gives low reward.
To measure frequency bias, we first construct a variety of interaction histories (occurred during environment interaction) containing between 2 and 10 transitions. 
This interaction history is collected using a random policy.
Given an initial interaction history, we repeat the last action in the history, which we also refer to as target action, between 0 and 100 times. 
Finally, we report the entropy all actions, $H(\theta) = -\sum_{a \in A} \pi_{\theta}(a \mid \tau) \log \pi_{\theta}(a \mid \tau)$ .
To achieve this, we conduct a separate forward pass for every possible action in the action space and report the respective log probabilities.
We repeat the same procedure for different interaction histories and target actions (see Figure \ref{fig:failure-copycat}a and c).
For the 10 arms scenario, every interaction history therefore results in 1000 (10 arms * 100 repetitions of the target action) forward passes.
We repeat this procedure for the 5 target actions reported in Figure  \ref{fig:failure-copycat} using 5 interaction histories per action, accumulating to a total of 25K model forward passes (1000 * 5 * 5) per figure.

\noindent To quantify frequency bias, we categorize the resulting actions as \emph{frequent} action, \emph{greedy}, or \emph{other} if they are neither frequent nor greedy.
Subsequently, we compute the frequent $F_f$, greedy $F_g$ and other $F_o$ fractions as reported in Figure \ref{fig:failure-copycat}: 
\begin{equation}
F_f = \frac{N_T(a_f)}{N}; \quad
F_g = \frac{N_T(a_g)}{N}; \quad
F_o = \frac{\sum_{a \in A \setminus \{a_f, a_g\}} N_T(a)}{N}, \quad \text{with } N = \sum_{a \in A} N_T(a).
\end{equation}

\noindent Note that there can be an overlap between greedy and frequent actions. 
In these (rare) cases, the greedy action category is dominant, i.e., we categorize the action as greedy even if it would also be the frequent action. 
This implies that the actions classified as frequent in Figure \ref{fig:failure-copycat}, are always suboptimal/bad compared to the respective greedy action. 
Consequently, a high $F_f$ indicates that the model prefers the most frequent action even when observing a better action in the context.

\begin{figure*}[h]
\noindent
\small
\centering
\subfigure[Action Strategies]{
    \includegraphics[width=0.45\linewidth]{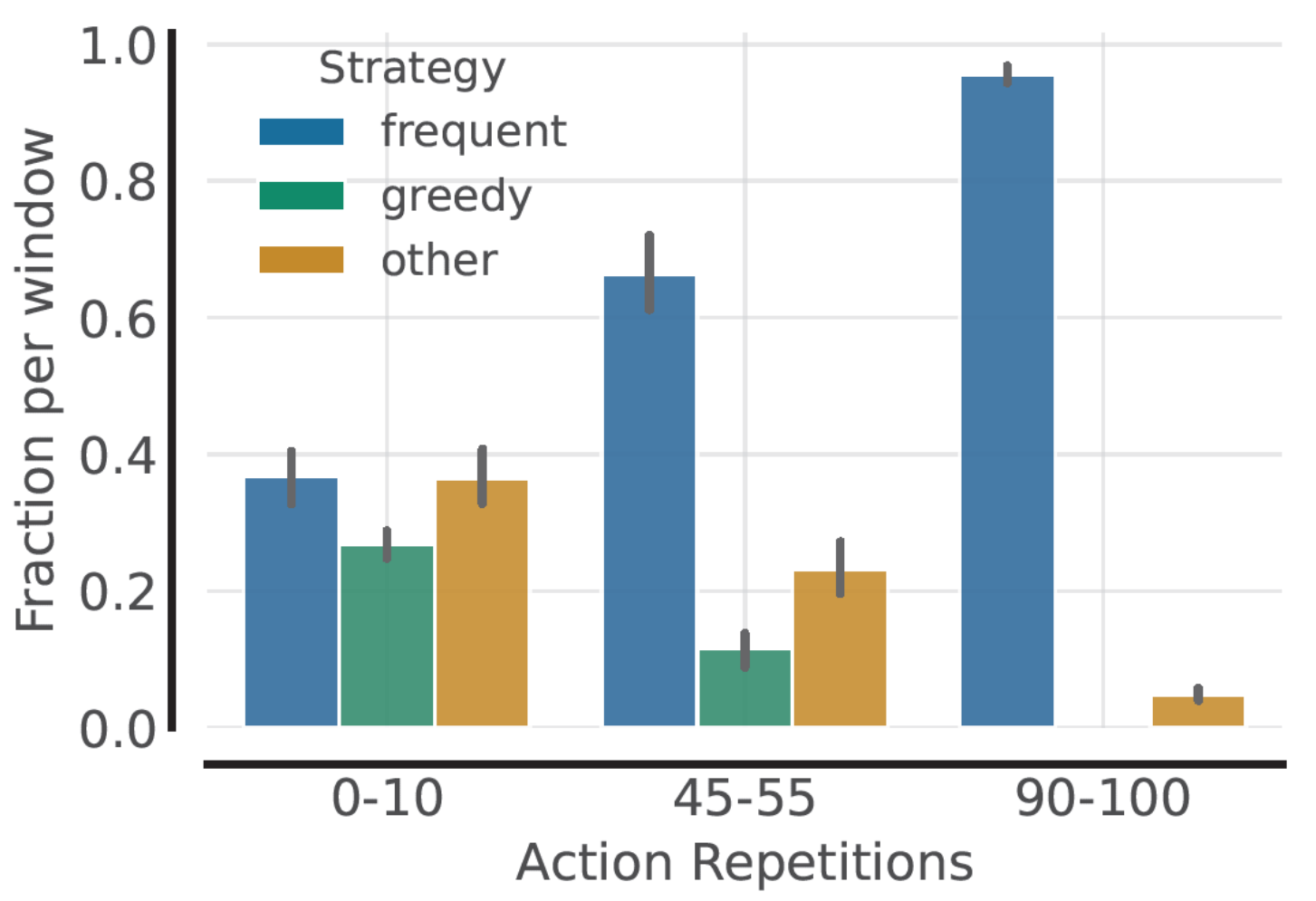}
}
~
\subfigure[Corrleations]{
    \includegraphics[width=0.33\linewidth]{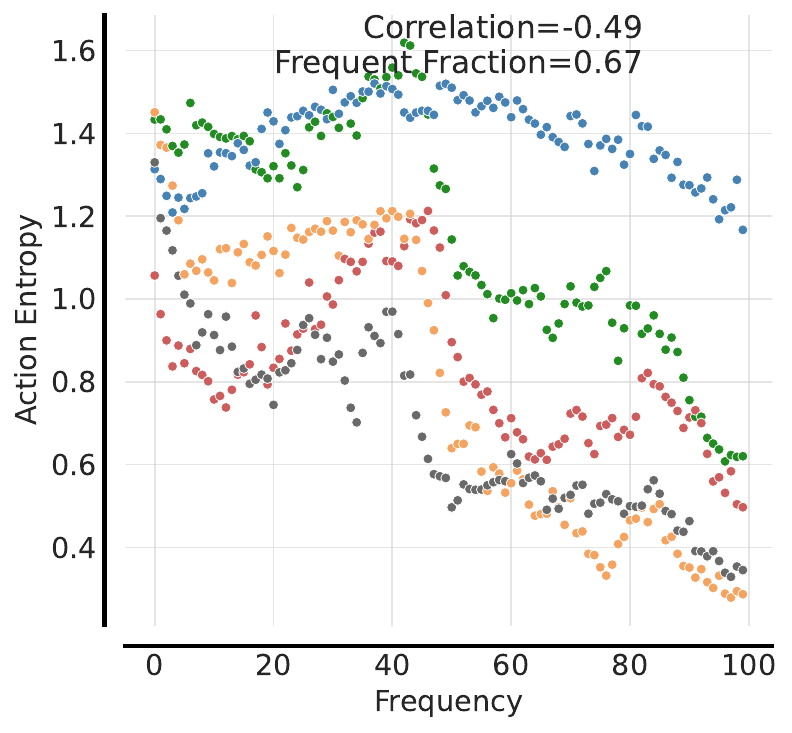}
}
\caption{Effect of RLFT on \textbf{frequency bias} for 2B (10 arms, medium noise).}
\label{fig:failure-copycat-postrlft-both}
\end{figure*}

\noindent\textbf{Post RLFT.}
In Section \ref{sec:multi-armed-bandits}, we observed that RLFT counteracts frequency bias. 
In addition to frequency buckets reported in Figure \ref{fig:failure-copycat-postrlft-both}a, we provide the plot frequency against action entropy post RLFT in Figure \ref{fig:failure-copycat-postrlft-both}b.
Compared to Figure \ref{fig:failure-copycat}a, we observe that after RLFT the models maintains a higher action entropy for longer.
Only at high repetition frequencies the action entropy decreases severely. 
Consequently, RLFT counteracts frequency bias, but does not completely alleviate it.  

\subsubsection{Knowing-Doing Gap}\label{appendix:failures-modes-kd}
The \emph{knowing-doing gap} has been observed by \citet{paglieri2024balrog} and \citet{ruoss2024lmact}. 
It states that models possess knowledge about a task or consequences of their behavior (i.e., they \emph{know} what to do), but cannot materialize this knowledge when acting (i.e., they are incapable of \emph{doing}).
We illustrate the knowing-doing gap empirically in Figure \ref{fig:failure-knowing-doing}.
To this end, we first task Gemma2 27B to produce the UCB algorithm and to compute the relevant quantities required to act according to UCB (“knowing”). 
This involves counting how often every action was selected, computing the mean rewards for every action, and computing the final UCB values.
After producing the quantities, the model is tasked to act according to them (i.e., “doing”). 
In Figure \ref{fig:ucb-agent-knowing-doing}, we present and example of the respective instructions given to the model along with a response produced by Gemma2 27B. 

\noindent To evaluate performance empirically, we let Gemma2 27B interact with the environment (64 parallel instances) for 50 timesteps. 
We extend the token generation budget to 2048 tokens per step, to accommodate the additional required computations. 
Every produced action $z$ contains both the CoT rationale $z_{CoT}$ and the final selected action $a$.
We first extract the computed UCB values from the produced rationale $z_{CoT}$. To achieve this, we task Gemma2 27B to enclose the computed values by \texttt{<ucb\_values>} and \texttt{</ucb\_values>} blocks.
Then we extract the selected action $a$ and execute it in the environment.
For this experiment, we use Gemma2 27B, because we found that 2B and 9B struggled with computing the relevant UCB quantities and with enclosing them appropriately under the desired blocks.

\noindent\textbf{Quantifying “Knowing”.} 
To quantify “knowing”, we compare the UCB values computed by the model and extracted from $z_{CoT}$ against the real UCB values.
To this end, we recompute the real UCB values for every action at every time-step given the observed history. 
We consider the rationale as correct if the arm with the highest UCB values match.
We opt for this choice rather than checking for exact equality, because we observed that the model struggles with exact calculations for complex operations.
This is expected, because the necessary computations involve logarithm and square roots of floating point values.
While tool use (e.g., calculator) could mitigate this issue, we observed that Gemma2 27B gets the quantities approximately right, resulting in valid rationales. 
Thus, the fraction of correct rationales is $F_c= \frac{1}{N} \sum_{i=1}^{N}g(z_{CoT}^i)$ given a classifier $g$.

\noindent\textbf{Quantifying “Doing”.} 
To quantify “doing”, we categorize the generated actions as \emph{optimal} action if the model selects the action with the highest UCB value, as \emph{greedy} if it selects the action with the highest UCB value among the set of actions tried so far, and as \emph{other} if the action is neither optimal nor greedy.
It is possible that the greedy action is the optimal action. 
However, in this case the action is considered optimal instead of greedy. 
Subsequently, we compute the percentages of greedy/optimal/other actions (e.g., $F_g \times 100$). 
We find that the model clearly \emph{knows} how to solve the task, with 89\% of all rationales being correct (see Figure \ref{fig:failure-knowing-doing}). 

\subsection{Multi-armed Bandits}\label{appendix:results-mabs}
\begin{figure}[htbp]
\noindent
\small
    \centering
\includegraphics[width=0.9\linewidth]{figures/mabs/legend_with_9B.png} 
\vspace{-0.25cm}

\subfigure[High Noise ($\sigma=3$)]{
    \includegraphics[width=0.95\linewidth]{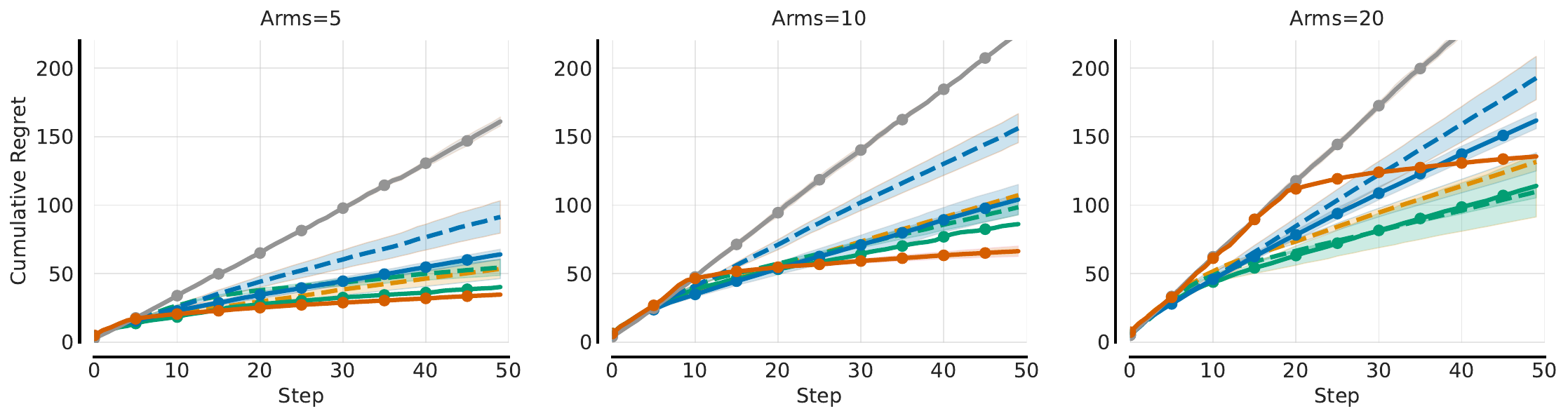} 
}
\subfigure[Low Noise ($\sigma=0.1$)]{
    \includegraphics[width=0.95\linewidth]{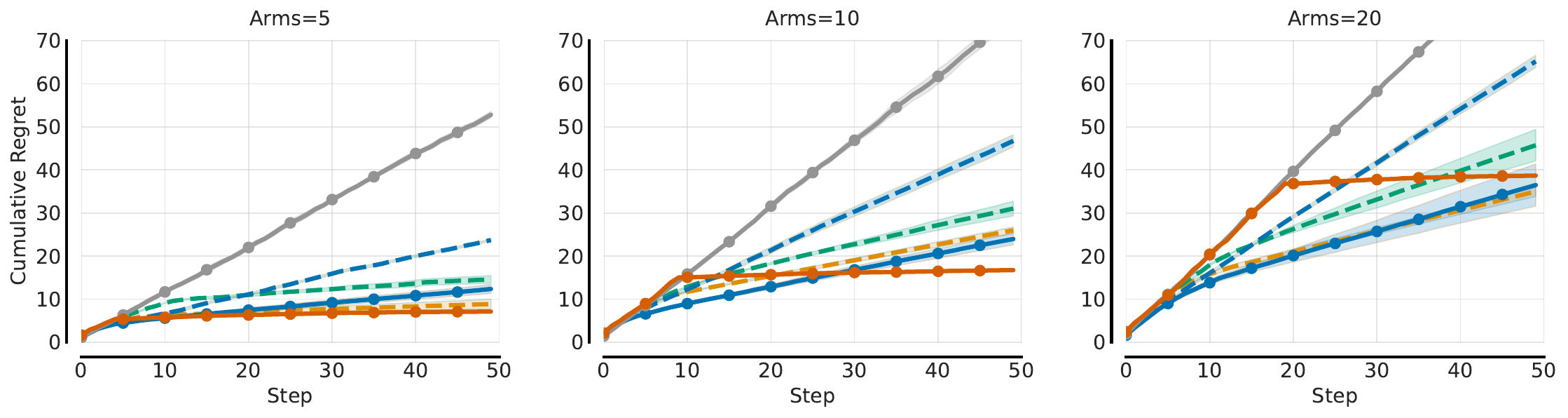} 
}
\caption{Main Comparison on \textbf{Gaussian MABs} button scenario in the \textbf{(a)} high $\sigma=3$ and \textbf{(a)} low $\sigma=0.1$ noise settings. We compare cumulative regrets (lower is better) of classic baselines against ICL and RLFT performances for Gemma2 2/9/27B for 5, 10, and 20 arms. }
\label{fig:mab-gaussian-karms-low-mednoise}
\end{figure}

In Figure \ref{fig:mab-gaussian-karms}, we report the cumulative regrets across model sizes and arms for a medium noise ($\sigma=1.0$) scenario.
In addition, we repeat the same experiment in the low-noise ($\sigma=0.1$) and the high-noise $\sigma=3.0$ setting in Figure \ref{fig:mab-gaussian-karms-low-mednoise}. 
For both noise levels, we observe similar trends as for the medium noise setting. 
In particular, we observe that LLMs clearly outperform the random baseline and RLFT lowers the cumulative regret for Gemma2 2B across all arm scenarios. 

\subsection{Contextual Bandits}\label{appendix:results-cbs}
\begin{figure}[htbp]
\noindent
\small
    \centering
\includegraphics[width=0.75\linewidth]{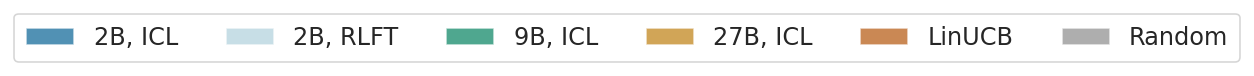} 

\includegraphics[width=0.85\linewidth]{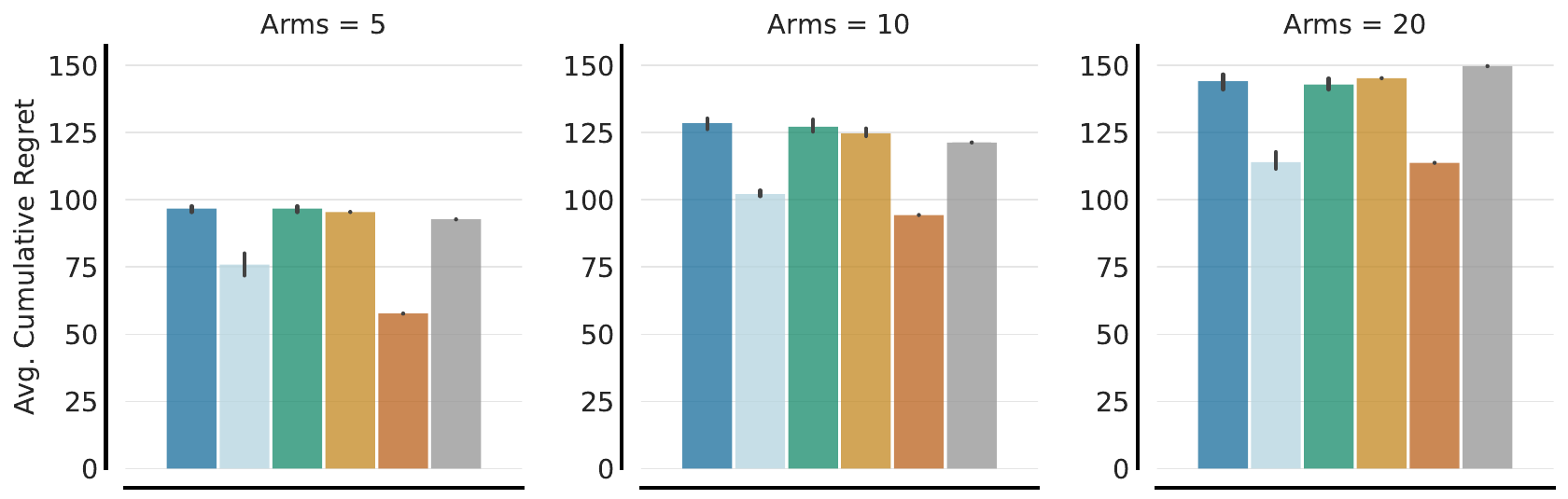} 

\caption{Main Comparison on Gaussian \textbf{MovieLens CBs} for \textbf{(a)} 5, \textbf{(b)} 10, and \textbf{(c)} 20 arms. We compare classic baselines against ICL and RLFT performances for Gemma2 2/9/27B. }
\label{fig:cb-gaussian-karms}
\end{figure}

We repeat the same fine-tuning experiment for the contextual MovieLens bandits described in Section \ref{appendix:cbs}. 
In Figure \ref{fig:cb-gaussian-karms}, we report the cumulative regrets attained by Gemma2B across different model sizes and for 5, 10 and 20 arms. 
Furthermore, we compare against a LinearUCB and a Random baseline. 
Overall, we observe similar performance improvements for RLFT on CBs as on MABs. 
While the ICL performances barely attain the same performance as a Random agent, RLFT fine-tuned Gemma2 2B perform similar to UCB.

\section{Ablations} \label{appendix:ablations}
Finally, we provide additional details on the ablations conducted in this work. 

\noindent \subsection{Tic-tac-toe: Effect of Legal Actions in State}
By default, we provided the legal actions available at the current turn within the input context to the agent.
We found this design choice to be essential for effective fine-tuning compared to training without legal actions (see Figure \ref{fig:tictactoe}b).
Without legal actions in the context the average return drops from 0.75 (w/ legal actions) to 0.45. 
This suggests that the LLM fails at identifying the appropriate actions among the set of all possible actions when not given legal actions at the current state.
In contrast, when provided with sufficient information (i.e., legal actions), the LLM is able to select actions appropriately (similar to Section \ref{sec:exps-exp-interventions}).
Providing the legal actions in the agent's context alleviates the need to explore/identify invalid actions.
Consequently, this shortcoming further highlights the need for principled exploration strategies for LLMs in decision-making scenarios. 
\begin{figure}[h]
\noindent
\small
\centering
\includegraphics[width=0.31\linewidth]{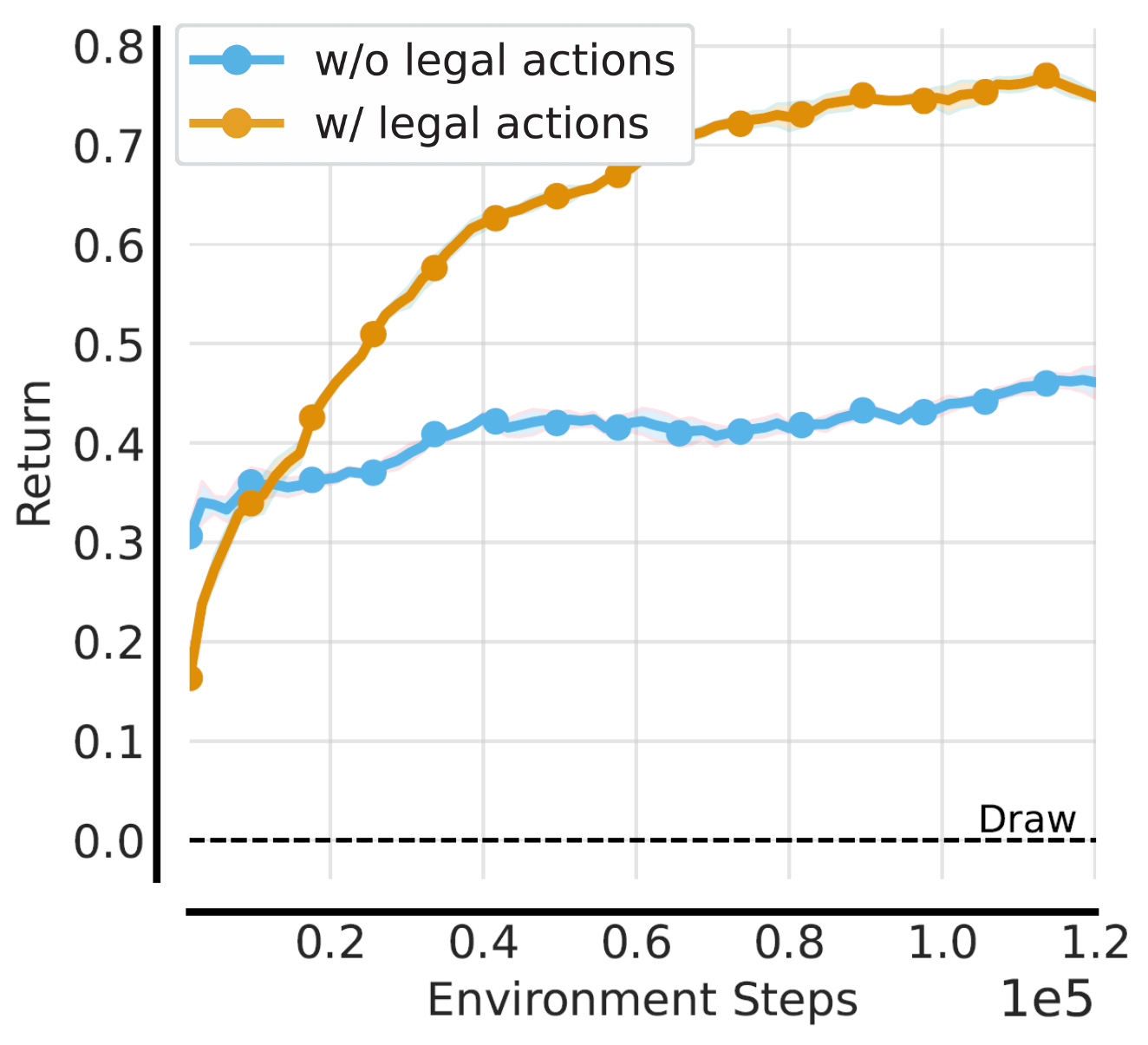} 
\caption{Effect of removing \textbf{legal actions} from the context in Tic-tac-toe.}
\label{fig:tictactoe-legal-actions}
\end{figure}

\noindent \subsection{Removing Chain-of-Thought (CoT)}
For our results without CoT reported in Figure \ref{fig:tictactoe}b, we remove the CoT instructions given to our agents. 
Instead, we instruct the agents to not perform any reasoning steps and to only produce the action to execute in the environment $a$. 
In addition, we limit the token generation budget $G$ to 16 to avoid that the model ignores the instructions and makes use of the additional tokens.
Furthermore, this considerably speeds up training due faster rollout times and shorter context lengths. 

\noindent \subsection{Expert Behavior Cloning vs. Thought Cloning}
A prevalent approach in sequence models for decision-making is behavior cloning (BC) \citep{pomerleau1988alvinn,reed2022generalist,brohan2022rt,brohan2023rt}, which relies on expert datasets. 
Consequently, we construct two UCB expert dataset comprising 32K rollouts either w/o CoT (\emph{behavior cloning}) or w/ CoT (\emph{thought cloning}), as described in Appendix \ref{appendix:mabs}.
Notably, both SFT variants successfully mimic the expert achieving comparable regret to the UCB expert (see Figure \ref{fig:tictactoe}b, SFT).
This result underscores the efficacy of training on expert data in decision-making scenarios when available, echoing recent findings in reasoning tasks \citep{muennighoff2025s1}. 
While BC and TC attain similar performance levels on the simplistic MABs, we anticipate that TC is advantageous in more complex decision-making scenarios as found by \citet{hu2023thought}.  

\noindent \subsection{“Thinking” Time}
We investigate the effect of giving the agent more time to “think” in Figure \ref{fig:tictactoe}c.
To achieve this, we vary the maximal number of tokens that the agent can generate per action $G \in \{16, 64, 256, 512\}$. 
By default, we set $G$ to 256. 
Indeed, we observe that the performance improves consistently with more thinking tokens.
Decreasing $G$ to 16 or 64 results in poor performance, because the agent is unable to rationalize its decisions within the restricted generation budget. 
This is similar to the performance without CoT, but in contrast, the agent is instructed to produce the reasoning process. 
Over the course of RLFT, the agents learn to produce short rationales $z_{CoT}$, including the action $a$ due to our reward shaping mechanism (see Section \ref{sec:method}).
However, the produced short rationales are unhelpful to improving agent performance.

\noindent In contrast, doubling $G$ from 256 to 512 results in a considerable performance increase to the level of Gemma2 9B with RLFT (see Figure \ref{fig:mab-gaussian-karms}).
We observe an increase in the average sequence length over the course of the RLFT process.
This suggests that the agent learns to effectively leverage the additional “thinking time” and reflects recent observations in mathematical reasoning \citep{guo2025deepseek}. 
However, the increased performance comes with additional training cost due to the multistep nature of decision-making scenarios.
In fact, we observed that rollout generation can make up the majority of the training time required by the RLFT process. 
This is because the agent has to produce more tokens at every environment interaction step.
For example, for our default horizon of 50 timesteps and a generation budget of 500, the agent produces 25K tokens (at maximum). 

\begin{figure}[h]
\noindent
\small
    \centering
\resizebox{1\textwidth}{!}{
\begin{minipage}[t]{1.\textwidth}
  \footnotesize
    \begin{tcolorbox}[colback=gray!3!white, colframe=gray!50!black, title=Instructions for Gemma2 27B as UCB agent]
        [Instructions]\newline 
        
        \sethlcolor{paleblush}~\hl{Your task is to act according to the Upper-Confidence-Bound (UCB) algorithm. First, write down the UCB algorithm. Then compute the relevant UCB values for every button (approximate if necessary). Finally, select your action according to the computed quantities.}\newline 
        
        [More Instructions]\newline 
    
        So far you have tried/seen: \newline
        \sethlcolor{antiquewhite}~\hl{Step=0} \sethlcolor{lavender}~\hl{Action=blue} \sethlcolor{azure(web)(azuremist)}~\hl{Reward=1.06} \newline
        \sethlcolor{antiquewhite}~\hl{Step=1} \sethlcolor{lavender}~\hl{Action=blue} \sethlcolor{azure(web)(azuremist)}~\hl{Reward=1.82} \newline
        \sethlcolor{antiquewhite}~\hl{Step=2} \sethlcolor{lavender}~\hl{Action=green} \sethlcolor{azure(web)(azuremist)}~\hl{Reward=1.0} \newline
        \sethlcolor{antiquewhite}~\hl{Step=3} \sethlcolor{lavender}~\hl{Action=green} \sethlcolor{azure(web)(azuremist)}~\hl{Reward=-0.26} \newline
        \sethlcolor{antiquewhite}~\hl{Step=4} \sethlcolor{lavender}~\hl{Action=blue} \sethlcolor{azure(web)(azuremist)}~\hl{Reward=-0.58} \newline
        \sethlcolor{antiquewhite}~\hl{Step=5} \sethlcolor{lavender}~\hl{Action=blue} \sethlcolor{azure(web)(azuremist)}~\hl{Reward=-0.34} \newline
        \sethlcolor{antiquewhite}~\hl{Step=6} \sethlcolor{lavender}~\hl{Action=green} \sethlcolor{azure(web)(azuremist)}~\hl{Reward=1.19} \newline
        \sethlcolor{antiquewhite}~\hl{Step=7} \sethlcolor{lavender}~\hl{Action=green} \sethlcolor{azure(web)(azuremist)}~\hl{Reward=2.21} \newline
        \sethlcolor{antiquewhite}~\hl{Step=8} \sethlcolor{lavender}~\hl{Action=green} \sethlcolor{azure(web)(azuremist)}~\hl{Reward=0.07} \newline
        \sethlcolor{antiquewhite}~\hl{Step=9} \sethlcolor{lavender}~\hl{Action=green} \sethlcolor{azure(web)(azuremist)}~\hl{Reward=1.45} \newline
        \sethlcolor{antiquewhite}~\hl{Step=10} \sethlcolor{lavender}~\hl{Action=green} \sethlcolor{azure(web)(azuremist)}~\hl{Reward=-0.11} \newline
        What do you predict next? 
    \end{tcolorbox}
\end{minipage}%
}

\resizebox{1\textwidth}{!}{
\begin{minipage}[t]{1\textwidth}
  \footnotesize
    \begin{tcolorbox}[colback=gray!3!white, colframe=gray!50!black, title=Generated response by Gemma2 27B]
        Here's how I'd approach this as a UCB bandit algorithm:

        \textbf{UCB Algorithm (Short Version)}
        
        \begin{itemize}
            \item \textbf{Initialization:} For each button, set a count (number of times pressed) to 0 and an estimated reward average to 0.
            \item \textbf{Iteration:}
            \begin{itemize}
                \item For each button, calculate the UCB value: $UCB = \text{average\_reward} + \sqrt{\frac{2 \cdot \ln(\text{total\_steps})}{\text{button\_count}}}$
                \item Choose the button with the highest UCB value.
                \item Update the button's count and average reward based on the observed reward.
            \end{itemize}
        \end{itemize}
        
        \textbf{UCB Calculations (Approximate)}
        
        Let's denote the buttons by their colors. Here's a breakdown after 10 steps:
        
        \begin{itemize}
            \item \textbf{blue:}
            \begin{itemize}
                \item Count = 4
                \item Average Reward = (1.06 + 1.82 - 0.58 - 0.34) / 4 = 0.49
                \item UCB $\approx 0.49 + \sqrt{\frac{2 \cdot \ln(11)}{4}} \approx$ \sethlcolor{paleblush}~\hl{1.17}
            \end{itemize}
            \item \textbf{green:}
            \begin{itemize}
                \item Count = 7
                \item Average Reward = (1.0 + -0.26 + 1.19 + 2.21 + 0.07 + 1.45 - 0.11) / 7 $\approx$ \sethlcolor{paleblush}~\hl{0.92}
                \item UCB $\approx 0.92 + \sqrt{\frac{2 \cdot \ln(11)}{7}} \approx$ \sethlcolor{paleblush}~\hl{1.47}
            \end{itemize}
            \item \textbf{Other buttons:} Count = 0, Average Reward = 0, UCB = $\sqrt{\frac{2 \cdot \ln(11)}{0}} = \infty$ (We'll explore these later)
        \end{itemize}
        
        \textbf{Action Selection:} The button with the highest UCB value is \textbf{green}.

        \textbf{Therefore, ACTION=green}
        
        \textbf{Explanation:} The UCB algorithm balances exploration (trying buttons we haven't seen much) and exploitation (choosing buttons that seem to have high rewards).
        \begin{itemize}
            \item Green has a higher average reward than blue, but blue has been explored less.
            \item The UCB formula takes this into account, giving a higher value to green because its average reward is more reliable due to more observations.
            \item \sethlcolor{paleblush}~\hl{The other buttons have infinite UCB values because we haven't tried them yet}, so they are highly prioritized for exploration. However, green's current UCB is high enough to warrant another try.
        \end{itemize}
    \end{tcolorbox}
\end{minipage}%
}
\caption{Illustration of the \textbf{knowing-doing gap}. \textbf{(a)} Instructions for the agent, which is prompted to act like a UCB algorithm. \textbf{(b)} The response generated by Gemma2 27B with greedy decoding (temperature=0). The LLM “knows” the UCB algorithm and computes the relevant quantities approximately correctly, but acts erroneously by selecting the next action greedily.}
\label{fig:ucb-agent-knowing-doing}
\end{figure}

\end{document}